\documentclass[final,5p,times,twocolumn]{elsarticle}

\usepackage{amssymb}
\usepackage{amsmath}
\usepackage{graphicx}
\usepackage{subfig}  
\usepackage{multirow} 
\usepackage{booktabs} 
\usepackage[percent]{overpic}  
\usepackage{xcolor}      
\usepackage{tabularx} 
\usepackage{hyperref}

\makeatletter
\def\ps@pprintTitle{
 \let\@oddhead\@empty
 \let\@evenhead\@empty
 \let\@oddfoot\@empty
 \let\@evenfoot\@empty
}
\makeatother

\begin{document}

\begin{frontmatter}

\title{RePaintGS: Reference-Guided Gaussian Splatting\\for Realistic and View-Consistent 3D Scene Inpainting}

\author[inst1,inst2]{Ji Hyun Seo} 
\ead{jihyun.seo@wrl.onl}
\author[inst1,inst3]{Byounhyun Yoo\corref{cor}} 
\ead{yoo@byoo.org}
\author[inst2]{Gerard Jounghyun Kim\corref{cor}} 
\ead{gjkim@korea.ac.kr}
\cortext[cor]{Co-corresponding authors.}

\affiliation[inst1]{organization={Intelligence and Interaction Research Center, Korea Institute of Science and Technology},
            addressline={5 Hwarang-ro, Seongbuk-gu}, 
            city={Seoul},
            postcode={02792}, 
            country={Republic of Korea}}

\affiliation[inst2]{organization={Department of Computer Science and Engineering, Korea University},
            addressline={145 Anam-ro, Seongbuk-gu}, 
            city={Seoul},
            postcode={02841}, 
            country={Republic of Korea}}

\affiliation[inst3]{organization={AI-Robotics, KIST School, Korea National University of Science and Technology},
            addressline={5 Hwarangro14-gil, Seongbuk-gu}, 
            city={Seoul},
            postcode={02792}, 
            country={Republic of Korea}}

\begin{abstract}
Radiance field methods, such as Neural Radiance Field or 3D Gaussian Splatting, have emerged as seminal 3D representations for synthesizing realistic novel views. For practical applications, there is ongoing research on flexible scene editing techniques, among which object removal is a representative task. However, removing objects exposes occluded regions, often leading to unnatural appearances. Thus, studies have employed image inpainting techniques to replace such regions with plausible content—a task referred to as 3D scene inpainting. However, image inpainting methods produce one of many plausible completions for each view, leading to inconsistencies between viewpoints. A widely adopted approach leverages perceptual cues to blend inpainted views smoothly.  However, it is prone to detail loss and can fail when there are perceptual inconsistencies across views.
In this paper, we propose a novel 3D scene inpainting method that reliably produces realistic and perceptually consistent results even for complex scenes by leveraging a reference view.
Given the inpainted reference view, we estimate the inpainting similarity of the other views to adjust their contribution in constructing an accurate geometry tailored to the reference. This geometry is then used to warp the reference inpainting to other views as pseudo-ground truth, guiding the optimization to match the reference appearance. Comparative evaluation studies have shown that our approach improves both the geometric fidelity and appearance consistency of inpainted scenes. For more details, please visit our \href{https://repaintgs.github.io/}{project page}.
\end{abstract}


\end{frontmatter}

\section{Introduction}
\label{sec1}
\begin{figure*}[!ht]
\center
\includegraphics[width=\textwidth]{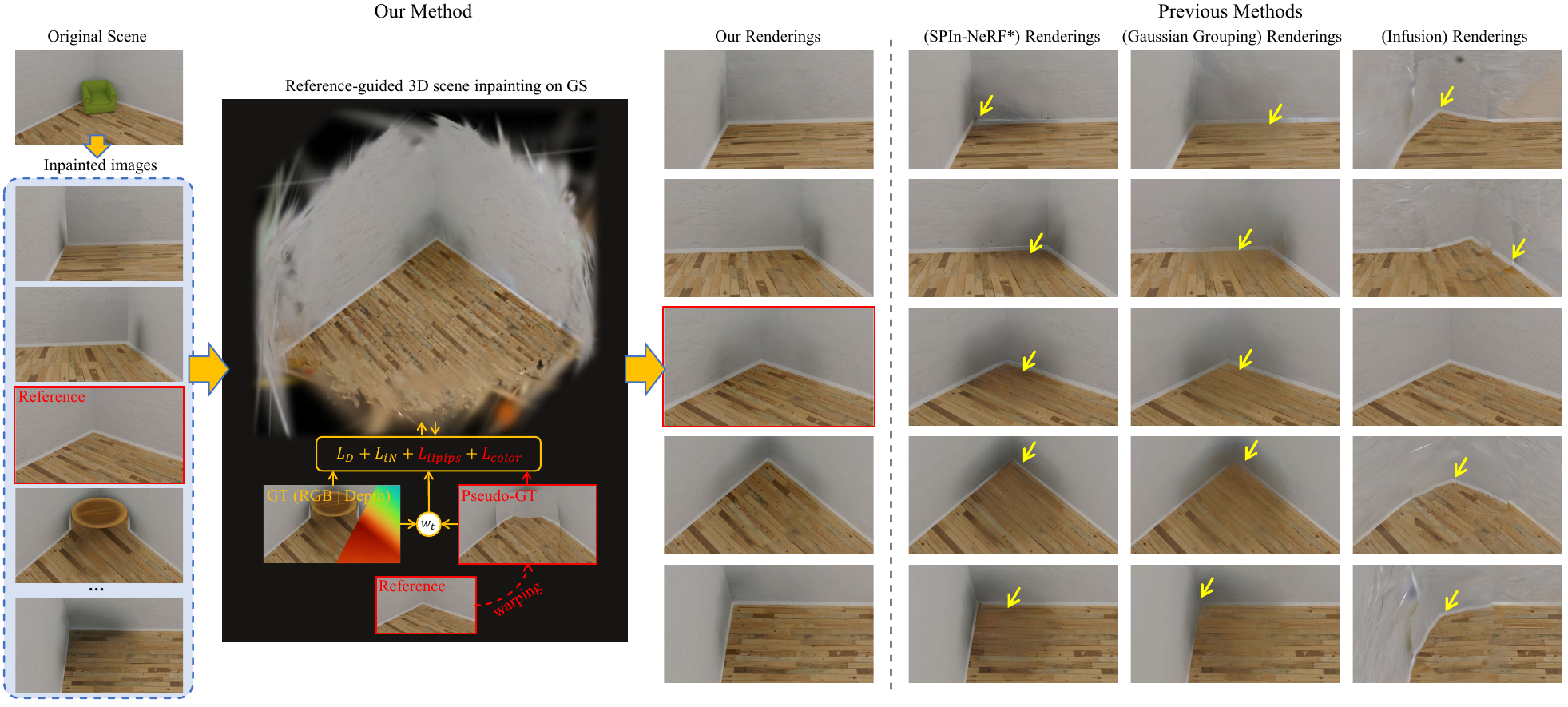}
\caption{Our method provides high-fidelity and view-consistent 3D scene inpainting using 3DGS, even in complex scenes with severe inconsistencies across inpainted views. Given a user-selected reference view among inpaintings, our method guides 3DGS to preserve the reference appearance while adaptively leveraging other views based on their reliability. Arrows indicate areas where the previous method underperforms. (*SPIn-NeRF results are obtained under our re-implemented setup with conditions closely matching ours; Refer to Section~\ref{sec:experiment_implement} for details.)}
\label{fig:overview_summary}
\end{figure*}

Radiance field methods, such as Neural Radiance Field (NeRF)~\cite{Mildenhall2020NeRF, Barron2022MipNeRF360, Mueller2022Instant} and 3D Gaussian Splatting (3DGS)~\cite{Kerbl20233DGS, Guedon2024SuGaR, Cheng2024Gaussianpro}, have emerged as leading and indispensable techniques for 3D representation due to their ability to synthesize highly realistic novel views. These technologies are increasingly replacing traditional 3D representation methods in applications such as Digital Twins, Virtual/Augmented Reality (VR/AR), and remote collaboration. To enable more versatile usage, demand for editable radiance fields~\cite{Chen2024GaussianEditor} has been increasing, prompting active research efforts to support flexible scene modifications. Among various editing tools, such as relighting~\cite{Rudnev2022NeRF, Liang2024GS-IR, Gao2025Relightable} and style transfer~\cite{Liu2023StyleRF, Liu2024StyleGaussian}, object removal is considered one of the essential features for complete scene editing, allowing users to eliminate unnecessary elements and thus support more flexible scene rearrangement. However, removing object exposes previously occluded regions, often leading to an unnatural or incomplete appearance~\cite{Yang2021Learning}. Reconstructing these regions is particularly challenging when they are not visible from any viewpoint. 
To address this, techniques that generate plausible 3D regions that blend naturally with the surrounding scene are needed. Nevertheless, current 3D-aware generative models~\cite{Anciukevicius2023Renderdiffusion,Xiang20233Daware, ChanMonteiro2020piGAN} are typically optimized for limited types of training data and simplified scene structures, making it difficult to generalize to complex and diverse real-world environments.

Accordingly, studies applied image inpainting techniques~\cite{Suvorov2022LaMa, Rombach2022SD, Podell2023SDXL} to the radiance field method to reconstruct hidden regions of the 3D scene, a task referred to as 3D scene inpainting. However, because image inpainting methods are applied independently per view and do not consider the 3D consistency in structure and texture, they produce inconsistencies that can lead to distortions and hallucinations in radiance fields. Therefore, 3D scene inpainting research focuses on attempts to blend image inpainting results into a plausible and coherent 3D scene. A notable work, SPIn-NeRF~\cite{Mirzaei2023Spin}, leveraged the LPIPS metric~\cite{Zhang2018LPIPS} to measure perceptual similarity rather than pixel-level accuracy for the inpainted region. While subsequent research~\cite{Yin2023Ornerf, Huang2023Pointn, Wang2023Inpaintnerf360, Ye2024GaussianGrouping} also adopted the LPIPS metric to blend inconsistent inpainting results more seamlessly, it remains unsatisfactory due to the loss of fine-grained details and its vulnerability to discrepancies beyond the perceptual level.
While image inpainting techniques are being used to reconstruct occluded regions seamlessly, it is notable that inpainting does not aim to find the `correct' solution but instead provides one of many plausible outcomes. 
The larger the areas required for inpainting, the more diverse the possible outcomes. Consequently, expecting the inpainting output to maintain perceptual consistency leads to inherent instability. 
Even without considering the context level, image inpainting on scenes with geometry or texture that are complex or irregular can hardly achieve perceptual consistency across views.
Despite this, very few studies acknowledge this issue, with most merely stating that inpainting is performed on `selected' views perspective.
Mirzaei et al.~\cite{Mirzaei2023Reference} propose a depth-based view transformation approach that warps a user-provided single inpainting view to others to mitigate this issue. This input, called the reference image, improves view consistency by filling occluded regions in other views with a single source. However, the method heavily depends on the accuracy of depth estimation derived from the reference image. While errors may not be apparent around the reference view, they can become significant when the viewpoint changes. 

In this paper, we propose a 3D scene inpainting method that robustly produces realistic and consistent results across multiple views, despite inconsistent image inpainting results per view—common in complex or wide view scenes(see Fig.~\ref{fig:overview_summary}). To achieve this, we introduce inpainting confidence evaluation and a reference-guided 3DGS from the single inpainted image. The reference image, selected by the user from the inpainted views, serves as the intended target for 3D reconstruction. By warping a reference image across multiple viewpoints, we enable the seamless reconstruction of occluded regions while preserving 3D consistency. Furthermore, instead of relying solely on the estimated depth from the reference image, we incorporate inpainted views based on their similarity to the reference. We measure content-level similarity between the reference image and the inpainted views after aligning them via warping, enabling adaptive confidence weighting to improve 3D geometry. By leveraging inpainting confidence evaluation and the reference-guided approach, our method addresses a challenge that previous studies have struggled to overcome. 
Our approach leverages a single-reference inpainted image to infer hidden areas, ensuring that inpainted regions remain realistic and coherent across multiple viewpoints. This method significantly reduces inconsistencies, providing a robust solution for complex 3D scene editing where occluded information is unavailable.

Our contributing points are as follows:
\begin{itemize}
    \item Introduce a reference-guided 3D scene inpainting method that leverages confidence-based multi-view fusion and reference image warping to achieve realistic and perceptually consistent reconstruction in complex scenes.
    \item Estimate the confidence of multi-view inpainting results from the reference image to guide the construction of accurate, reference-aligned geometry.
    \item Demonstrate superior 3D scene inpainting performance on complex backgrounds compared to conventional inpainting methods, particularly in wide-view and irregular cases.
\end{itemize}

\begin{figure*}[!t]
\center
\includegraphics[width=\textwidth]{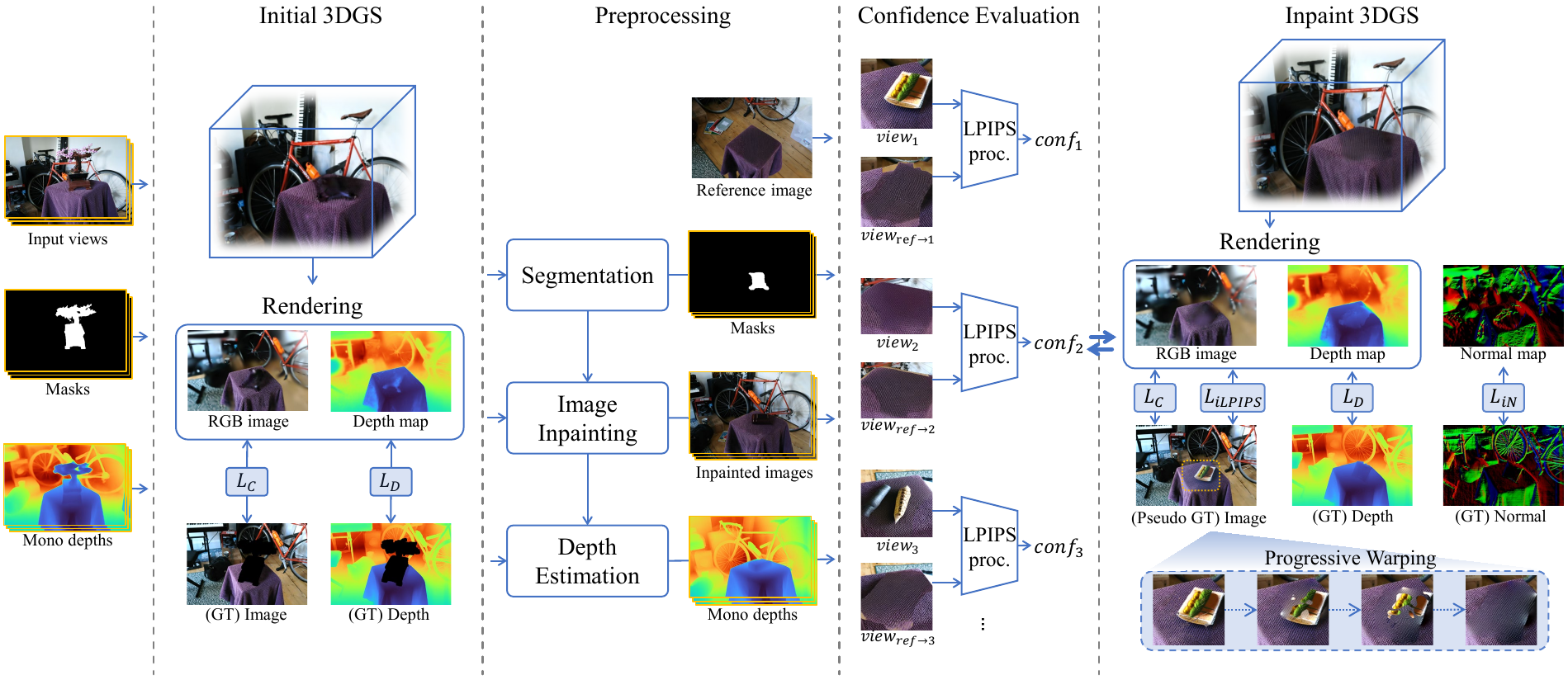}
\caption{Overview of our proposed method. First, the method performs an Initial 3DGS to reconstruct the background 3D scene while excluding the target object. This allows rendering the background behind the object, reducing unnecessary inpainting. Next, segmentation and inpainting are applied to the incomplete regions, which are primarily heavily occluded areas. Among the inpainted images, the user defines one as the reference image to guide the 3D reconstruction. Based on the inpainted results and the reference, the inpainting confidence evaluation computes the reliability of each view with respect to the reference image warped to the same viewpoint. Finally, the Inpaint-3DGS reconstructs the 3D inpainting by guiding the inpainting regions closer to the reference. This is done by adjusting each view's weight based on their confidence and warping the reference view to serve as a pseudo-ground truth.}
\label{fig:overview}
\end{figure*}

\section{Related Works}

\begin{figure}[!htb]
    \centering
\includegraphics[width=\columnwidth]{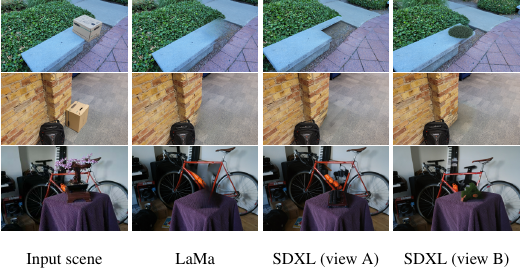}

    \caption{Limitation in image inpainting. For complex scenes, image inpainting often fails to fill in the background seamlessly or produces semantically different outputs across views. The results shown here are obtained using LaMa~\cite{Suvorov2022LaMa} and SDXL~\cite{Podell2023SDXL}, two widely used inpainting models.}
    \label{fig:inpainting_challenges}
\end{figure}

\begin{figure}[!htb]
 \center

     \subfloat[Inpainted image w/ Mask]{\label{fig:lpips_limit_a}
     \includegraphics[width=0.45\columnwidth]{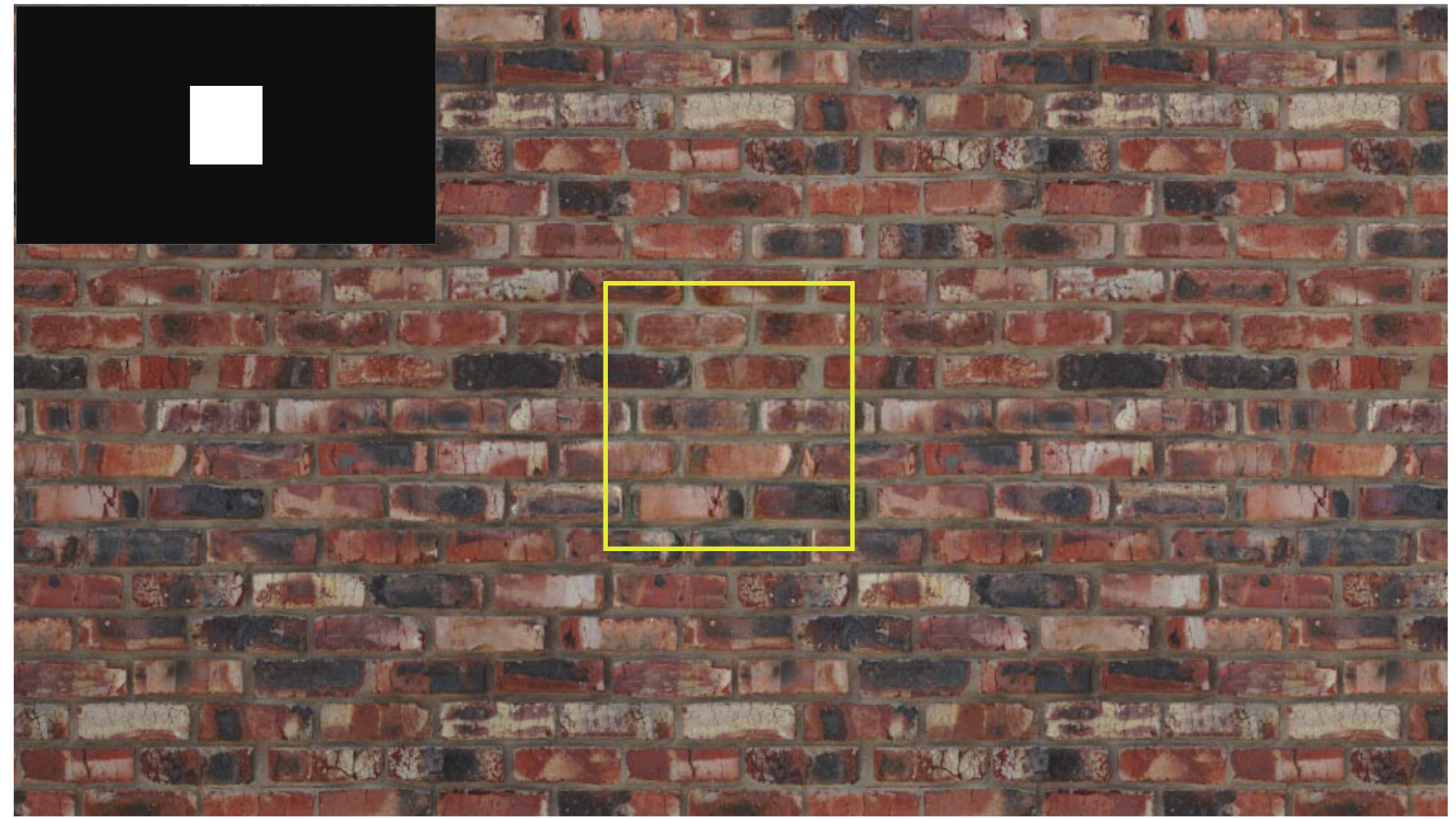}}
    \hfill         
     \subfloat[LPIPS-optimized result]
     {\label{fig:lpips_limit_b}
     \includegraphics[width=0.45\columnwidth]
     {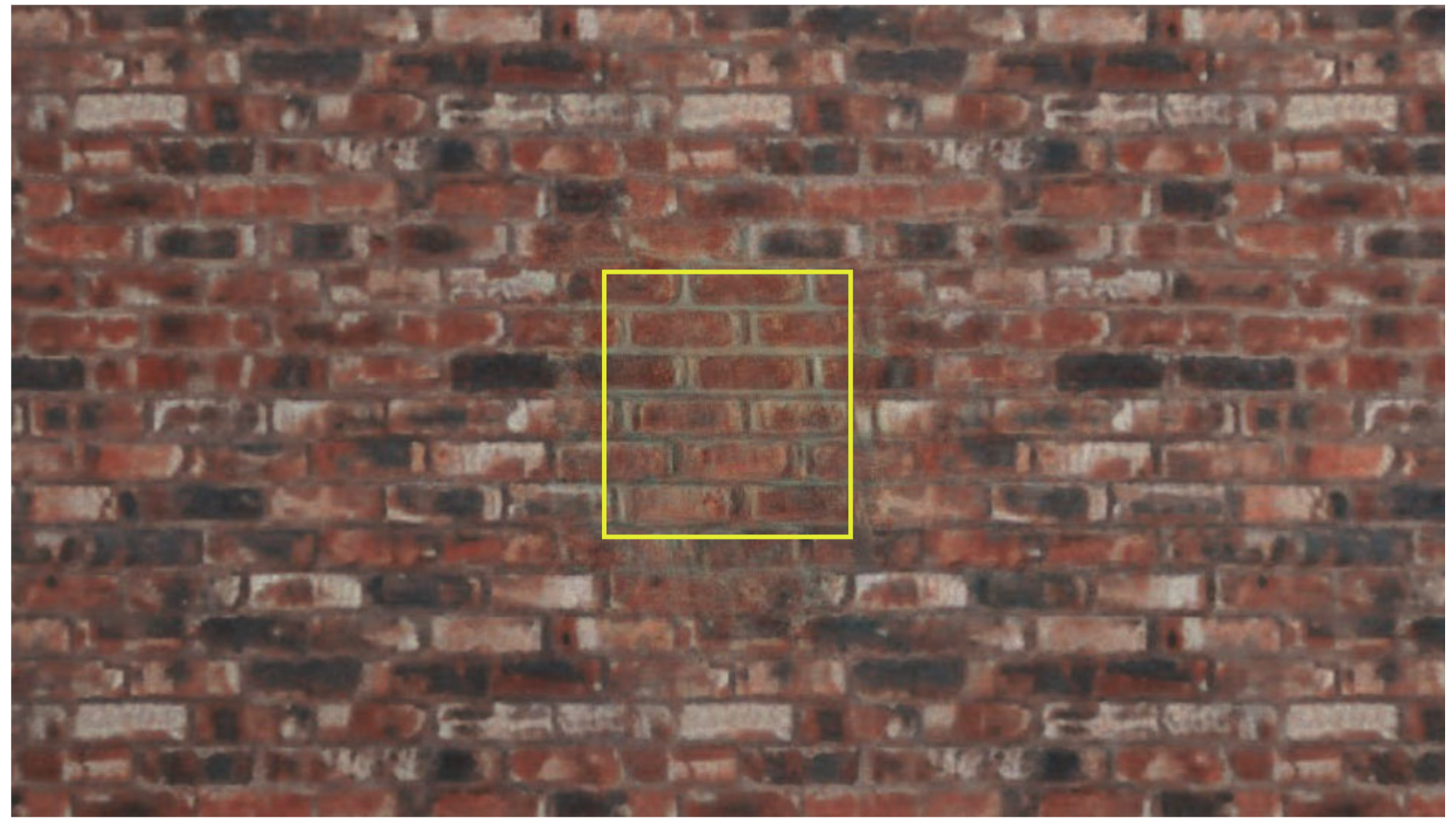}
     }

    \subfloat[Inpainted image w/ Mask]{
        \label{fig:lpips_limit_c}
        \includegraphics[width=0.45\columnwidth]{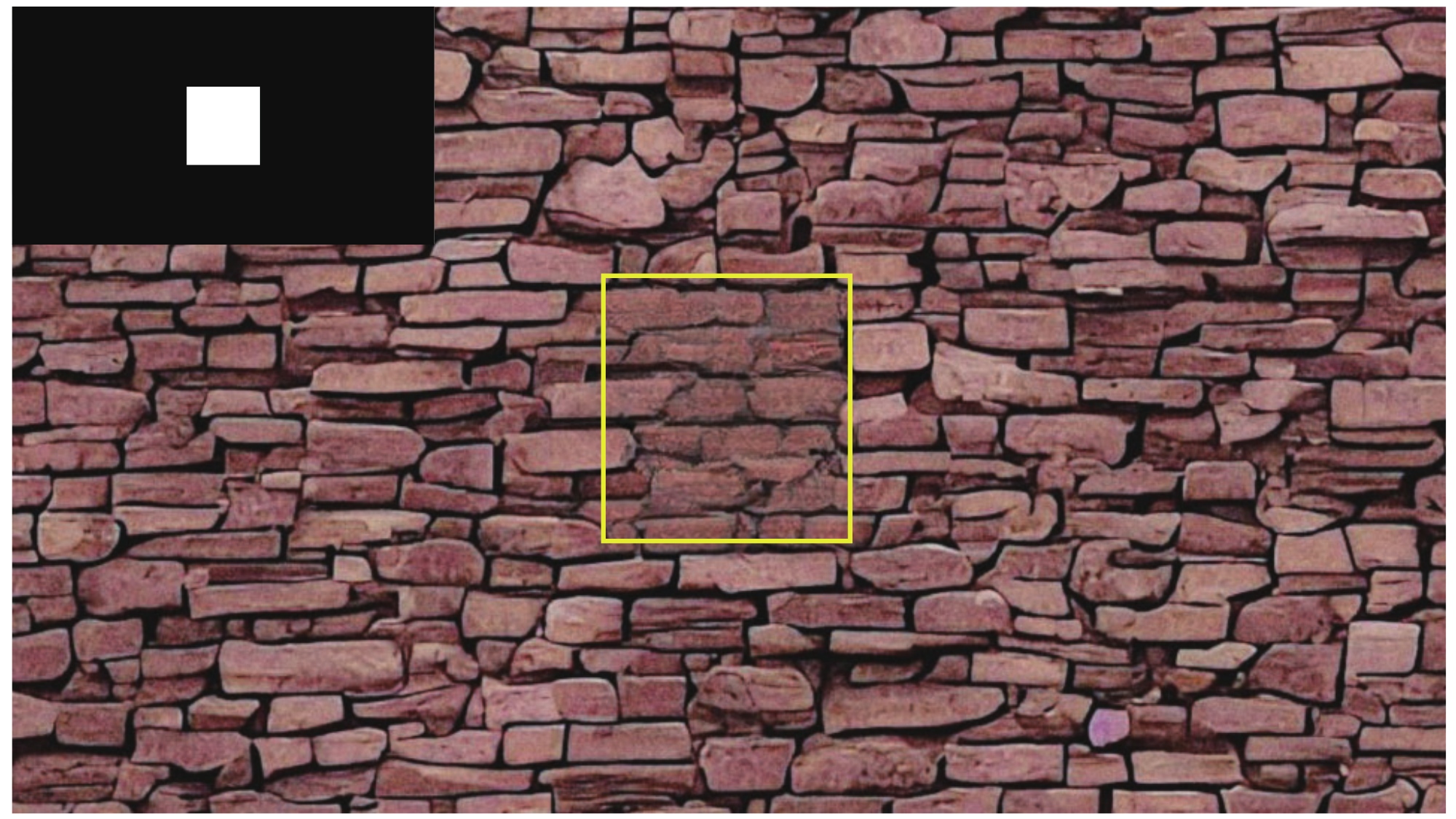}
    }
    \hfill
    \subfloat[LPIPS-optimized result]{
        \label{fig:lpips_limit_d}
        \includegraphics[width=0.45\columnwidth]{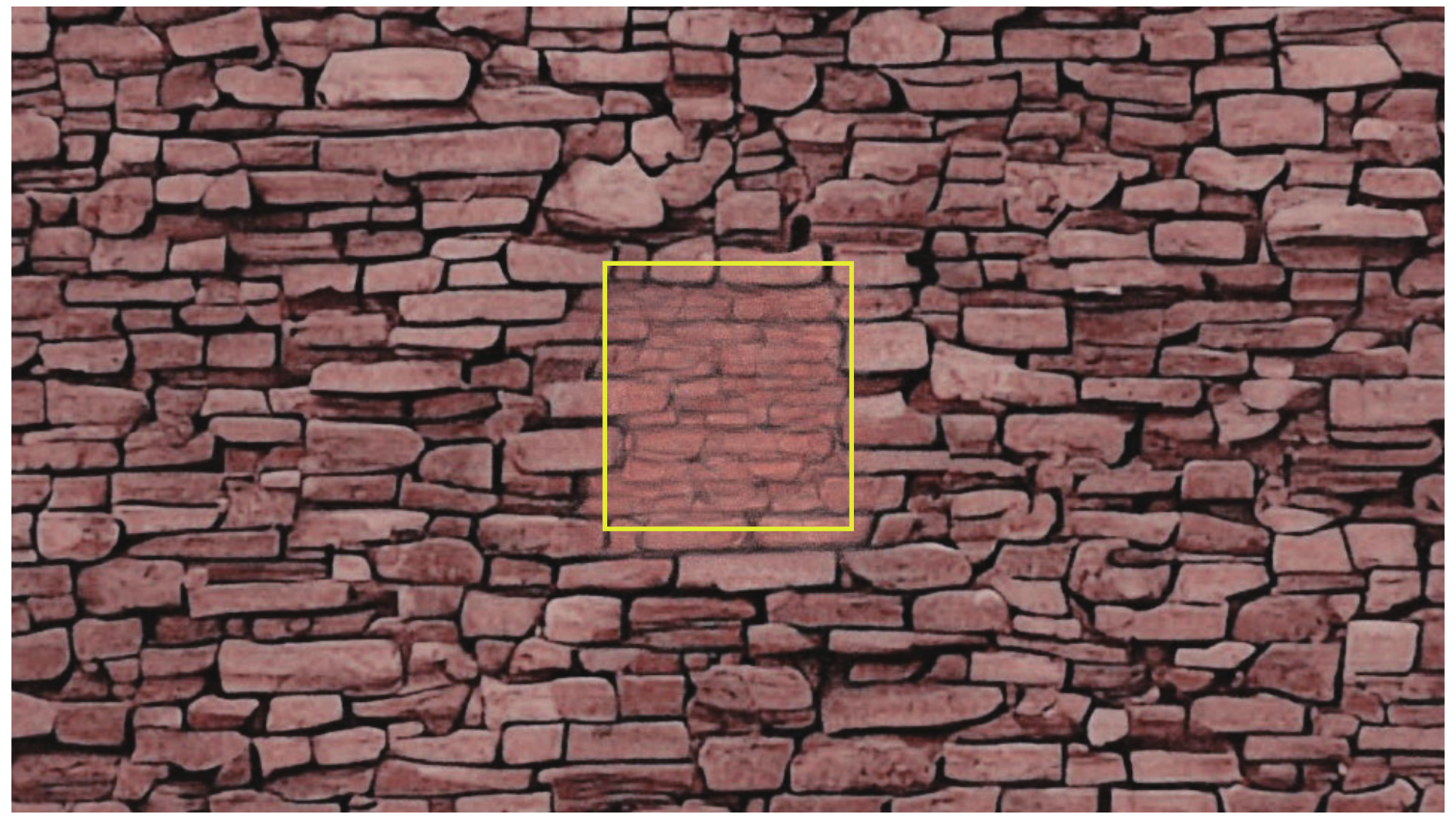}
    }

      \caption{Limitation in LPIPS metric. LPIPS performs well on regular patterns when optimizing multi-view inpainting results, as it captures shared visual structures across views beyond pixel-wise differences. However, it performs poorly on irregular ones, where achieving consistent visual structures across views is challenging. (a) and (c) show samples of inpainted input views with regular and irregular patterns, respectively, while (b) and (d) present the corresponding results.}
      \label{fig:lpips_limit}
\end{figure}

The 3D scene inpainting method for radiance fields leverages image inpainting techniques to synthesize a new spatial layout on the 3D scene. While 3D scene inpainting may involve modifying styles or inserting new objects, this paper specifically addresses the recovery of backgrounds exposed by object removal.
Before delving into this, it is important to note that standard radiance field methods assume a static scene and consistent input across views, enabling stable 3D reconstruction based on view poses and image data. 
However, the image inpainting technique is performed individually on images without considering 3D information, resulting in differences across views. Studies have proposed diverse methods to address the issue of inconsistent inpainting results. We categorize these approaches into the following four parts: view selection, 3D segmentation, 3D optimization, and geometric transformation.

\textbf{View Selection.} View selection refers to selecting appropriate views for reconstructing the inpainted regions. Filtering out inpainting results that have failed or significantly differ from other views can help reduce inconsistencies. Though recent image inpainting techniques~\cite{Suvorov2022LaMa, Podell2023SDXL} show high-quality and stable performance across various scenes, they may fail in complex scenes or under unfamiliar view directions (see Fig.~\ref{fig:inpainting_challenges}). Moreover, inconsistencies across multi-view can range from minor details to content level. 
Despite this, most studies do not handle how to select appropriate training samples in the presence of inconsistent inpaintings. 
Some merely mention that the experiments were conducted on `user-chosen' views without further explanation~\cite{Shen2024NeRFIn}.
Weder et al.~\cite{Weder2023Removing} attempts to adjust contribution per view by estimating the inpainting confidence as a loss attenuation term, similar to aleatoric uncertainty. However, this confidence fails to select reliable views when the majority of inpainted views lack mutual consistency, making it particularly vulnerable to diverse inpainting completions caused by large masked regions.

\textbf{3D segmentation.} 3D segmentation involves separating the target object from the background in the 3D scene rather than segmenting it directly in 2D images for inpainting.
While early studies in object removal in radiance fields applied image inpainting to the entire regions containing target objects per view, Huang et al.~\cite{Huang2023Pointn} pointed out that applying inpainting even to parts occluded in one view but visible in another is unnecessary. To address this, they propose segmenting the target object during the initial 3DGS training, which enables extracting the reconstructed background behind the object and thus reduces unnecessary inpainting. A related idea is seen in Gaussian Grouping~\cite{Ye2024GaussianGrouping} and Infusion~\cite{Liu2024Infusion}. In the case of Gaussian Grouping, it jointly reconstructs and segments components of the scene and then supports object removal for each instance. 3D segmentation within the radiance field helps reduce unnecessary inpainting, effectively narrowing down the regions that 3D scene inpainting needs to address. However, further solutions are needed to handle inconsistent inpainting in heavily occluded regions.

\textbf{3D optimization.} Regardless of the proposed preprocessing steps, 3D scene inpainting still requires an optimization process to integrate inconsistent inpainting results across views into a coherent and unified space. 
Differences in inpainting across views can confuse the geometry estimation process in the radiance field, leading to artifacts such as floating or hallucinations. To mitigate this, NeRF-In~\cite{Shen2024NeRFIn} proposes utilizing depth inpainting to assist geometry construction in radiance fields. They optimize the depth loss to enforce the geometry with the inpainted depth. Since implementing a 3D scene relying solely on image inpainting was unstable, depth inpainting has been widely adopted in subsequent related studies to stabilize the 3D geometry to some extent~\cite{Yin2023Ornerf, Weder2023Removing, Mirzaei2023Spin, Huang2023Pointn}.
For depth inpainting, various estimation methods have been applied and can be substituted. 
One commonly used model is LaMa~\cite{Suvorov2022LaMa}, which was used for RGB image inpainting. Mirazaei et al.~\cite{Mirzaei2023Spin} found that the same LaMa used for RGB inpainting also provides sufficiently high-quality results for depths. 
These are also demonstrated in experiments by Fischer et al.~\cite{Fischer2023Inpainting}, which compare different deep neural networks for depth inpainting.
However, while LaMa approximates depth in unknown regions based on the surrounding context, the resulting geometry is often inaccurate.  Pointing out these limitations, Liu et al.~\cite{Liu2024Infusion} propose using an image-conditioned depth completion model that directly restores the depth map from the RGB image. This method estimates depth in a way that is more semantically consistent with the inpainted content by leveraging surrounding depth cues and RGB context.
However, while depth inpainting assists 3D reconstruction under inconsistent image inpainting, the problem of reconstructing fine geometric details and coherently integrating their appearance still remains.

Not to mention the semantically different inpaintings, even slight differences across views, such as minor distortions or shifts, can be fatal for optimizing radiance fields-based methods when comparing pixel-to-pixel differences.
Therefore, Mirzaei et al.~\cite{Mirzaei2023Spin} propose using perceptual similarity for the inpainted region. The LPIPS metric~\cite{Zhang2018LPIPS} measures the content similarity between images in a way that aligns with human visual perception, allowing for distortions, noise, lighting changes, shifts, and color variations within a perceptible range. Comparing the similarities between contexts can smoothly gather the subtle differences between inpaintings as a coherent form.
Due to its ability to integrate inconsistent inpainting inputs into a consistent perceptual context, the use of LPIPS loss has been widely adopted in subsequent studies~\cite{Yin2023Ornerf, Wang2023Inpaintnerf360, Huang2023Pointn, Wang2024Innerf360, Ye2024GaussianGrouping}.
The LPIPS metric is effective when the inpainted train set is in perceptual similar relationships, as referred in Fig.~\ref{fig:lpips_limit}(b). However, in practice, the inpainting results are not deterministic due to the nature of the generative models. Achieving consistent inpainting outcomes across multiple views is challenging, as image inpainting can yield varying outputs not only in details but also in semantic content. For instance, the SDXL results in Fig.~\ref{fig:inpainting_challenges} show that, despite visually convincing complex scenes, the outputs can differ semantically even between closely related viewpoints.
As shown in Fig.~\ref{fig:lpips_limit}(d), LPIPS struggles with complex textures where consistent outputs are difficult to achieve.

\textbf{Geometric transformation.} To avoid the inherent limitations of inpainting, an alternative approach has been proposed that estimates depth from one or a few inpainted views and restores the scene through geometric transformation. One representative approach is by Mirzaei et al.~\cite{Mirzaei2023Reference}, who propose using a single inpainted image as a reference and transforming it to other views based on its estimated depth. This approach reduces the amount of inpainting required for other views by limiting it to view-specific occluded areas, thereby minimizing inconsistencies across views.
However, the depth of the reference image is estimated using monocular depth estimation, which may be less accurate. These errors can become more pronounced as the viewpoint moves away from the reference view.
Infusion~\cite{Liu2024Infusion} also proposes utilizing user-selected reference inpainting to improve fidelity. Instead of relying on cross-validation across multiple inpaintings to refine the geometry, their approach establishes a plausible initial geometry by unprojecting the inpainted depth into a 3D Gaussian point cloud. This serves as a robust initialization for subsequent GS fine-tuning. 
However, this approach also relies on estimated depth from one or a few reference views, and inaccuracies in these estimates can lead to distorted forms from other viewpoints. In other words, successful geometric transformation of the reference image into a 3D scene or other views requires validating the estimated depth of the selected views.

\section{Methods}
Our proposed method consists of three key processes: initial 3DGS, inpainting confidence evaluation, and inpaint-3DGS. The pipeline of our proposed approach is illustrated in Fig.~\ref{fig:overview}. The first key process of our method, initial 3DGS (section~\ref{sec:Method_3DGS_initial}), reconstructs the scene without the target object to reveal partially occluded backgrounds. We then apply state-of-the-art (SOTA) image inpainting and monocular depth estimation models to the rendered background images. 
Out of the inpainted results, the user selects a reference image which will provide a consistent visual guideline for how the target object should appear. The remaining inpainted views are compared with this reference to compute their content-level similarity. This similarity serves as an indicator of how reliably each inpainting can reconstruct the scene in accordance with the reference, a process we refer to as inpainting confidence evaluation(section~\ref{sec:Method_3_selection}). 
Based on the computed confidence scores, inpaint-3DGS optimizes the geometry of the inpainted region while reconstructing its appearance using warping techniques (section~\ref{sec:Method_3DGS_inp}). Ultimately, our method integrates inconsistent inpaintings into a coherent 3D reconstruction that faithfully preserves high-frequency visual details from the reference image.

\subsection{Initial 3DGS: Removal 3DGS}
\label{sec:Method_3DGS_initial}
Before attempting to inpaint the occluded region behind the target object, initial 3DGS is conducted on the original images with the segmented masks to reconstruct the background 3D scene. 

Initial 3DGS builds upon vanilla 3DGS~\cite{Kerbl20233DGS}, adopting its overall pipeline while introducing modifications specific to our task. The color $C$ of a pixel is computed by blending a depth-ordered set $\mathcal{N}$ of points overlapping the pixel, considering each point's color $c_i$ and its' opacity $\alpha_i$, and transmittance from pixel to that point.
\begin{equation}
C=\sum_{i\in\mathcal{N}}c_i\alpha_i\prod_{j=1}^{i-1}(1-\alpha_j)
\end{equation}
From this, we obtain the initial 3DGS by excluding the masked target region during reconstruction, thereby producing a 3D background scene without the target object.

For the color loss of the original region, the rendered image $\hat{I}$ is compared with input image $I$ at corresponding view using $\text{L1}$ loss with a D-SSIM term, excluding the masked region $M$:
\begin{equation}\label{eq:loss_ocolor}
\begin{aligned} 
\mathcal{L}_{oC} = (1 - \lambda_{\text{ssim}}) \frac{1}{|M^c|} \sum_{p \in M^c} \left| I(p) - \hat{I}(p) \right| \\
+ \lambda_{\text{ssim}} (1-\text{SSIM}(I|_{M^c}, \hat{I}|_{M^c}))
\end{aligned}
\end{equation}
Here, $I(p)$ denotes a pixel value at position $p$; $M^c$ denotes the complement of the masked region; $|M^c|$ is the number of pixels in that set; $I|_{M^c}$ is the restrictions of $I$ to the unmasked region $M^c$; and $\lambda_{ssim} \in [0,1]$ is a weighting coefficient.
Completely occluded regions across all views appear as holes, whereas partially occluded areas visible in other views can be recovered. 

However, even if a region is partially visible, reconstructing it may be difficult when the visible views are insufficient. In general, neural rendering struggles with limited view diversity, biased viewpoints, or visually simple indoor scenes. Such conditions often suffer from inaccurate geometry due to limited constraints or view diversity. Accordingly, to provide regularized optimization under sparse constraints, many studies have proposed using depth priors—such as sparse points from Structure-from-Motion (SfM)~\cite{Deng2022DSNeRF}, monocular depth estimation~\cite{Chung2024DepthRegul, turkulainen2024dnsplatter, Zhu2025FSGS}, or depth completion~\cite{Roessle2022DDPNeRF}—to guide the optimization better and boost reconstruction accuracy.
To this end, we utilize monocular depth estimates to guide geometry reconstruction during the initial 3DGS stage, enabling better generalization in diverse situations. Monocular depth provides context-aware dense estimates from a single image, whereas SfM yields only sparse depths at matched feature points, and depth completion struggles in large missing regions.
However, considering the inherent scale ambiguity and object-level noise in monocular depth estimation, we align the estimated depth maps to the COLMAP~\cite{Schoenberger2016Sfm} sparse point cloud, following DN-Splatter~\cite{turkulainen2024dnsplatter}.
By applying the closed-form linear regression solution, optimal scale $a$ and shift $b$ parameters for each image are found.
\begin{equation}
\hat{a}, \hat{b} = \underset{a,b}{\arg \min} \sum_{p \in P_\textbf{spc}} \left| (a D_{\textrm{mono}}(p) + b) - D_{\textrm{spc}}(p) \right|^2
\label{eq:align}
\end{equation}
Here, $P_{spc}$ is the set of pixels corresponding to the sparse point cloud, with $D_{mono}$ and $D_{spc}$ denoting the monocular estimated depth and the sparse point cloud depth, respectively.
During the optimization, per-pixel z-depth $D$ was estimated similar to colors, but using $i^{\text{th}}$ Gaussian's z-depth $d_i$ in view space instead of its' color. 
\begin{equation}
D=\sum_{i\in\mathcal{N}}d_i\alpha_i\prod_{j=1}^{i-1}(1-\alpha_j)
\end{equation}
With the rendered depth $\hat{D}$ of the specific view, corresponding estimated depth $D_{mono}$ is compared. The depth loss of the original region is formulated as follows:
\begin{equation}
\mathcal{L}_{oD} = \frac{1}{|M^c|} \sum_{p \in M^c} \left| D_{mono}(p) - \hat{D}(p) \right| 
\label{eq:depth_loss_bg}
\end{equation}
Here, the estimated depth serves as auxiliary guidance, improving structural accuracy.

Initial 3DGS reduces the unnecessary inpainting in regions that are occluded but visible from other points of view. Rendered background images are used to generate preprocessing data for 3D inpainted scene reconstruction. Using SOTA models such as Segment Anything Model 2~\cite{Ravi2024SAM2}, Stable Diffusion XL~\cite{Podell2023SDXL}, and Depth Anything v2~\cite{Yang2024Depthanythingv2}, we sequentially generate a reduced mask by segmenting unseen regions, then perform image inpainting, and estimate depth. The estimated depth, currently obtained from monocular estimation, is then aligned to the sparse point cloud from SfM.

\subsection{Inpainting Confidence Evaluation}
\label{sec:Method_3_selection}
\begin{figure}[htb]
\centering
\includegraphics[width=\columnwidth]{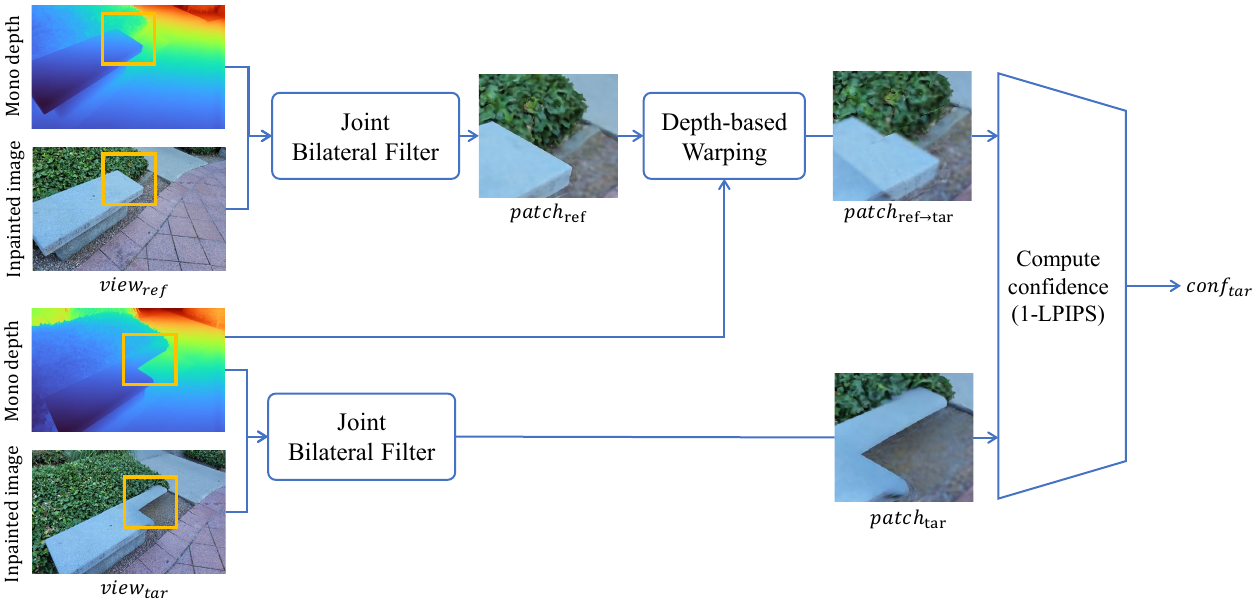}
\caption{Confidence evaluation of inpainted region. Each region's confidence is assessed by warping a reference view to each target view using their estimated depth maps and comparing their content-level similarity. To reduce the influence of fine texture differences, a strong bilateral filter is applied beforehand to suppress details on smooth surfaces.}
\label{fig:warp_sim}
\end{figure}

Considering that inpainting results can vary significantly from low-level details to high-level semantics,  we evaluate the reliability of each view’s inpainting based on a reference image. Among the inpainted views generated in the previous stage, where background images were rendered from the initial 3DGS and then inpainted, a reference image is selected by the user. This reference reflects the target that the user intends to build. Unlike previous methods~\cite{Mirzaei2023Reference, Liu2024Infusion} that rely solely on the accurate depth estimation from a few reference views, our approach utilized other inpainted views—not indiscriminately, but by weighting their influence according to their similarity to the reference. Specifically, we estimate the reliability of each inpainted view by warping the reference image and measuring content-level similarity, enabling adaptive weighting in the reconstruction process. This process can be seen in Fig.~\ref{fig:warp_sim}.

\subsubsection{Warping}
\label{subsec:Method_3_warping}
Image warping is the process of manipulating an image by distorting its shape. It is also used to geometrically transform an image based on correspondences such as depth, simulating how the image would appear from a different viewpoint.
Through this process, different inpainting results are geometrically aligned to a common view, thereby facilitating a structural similarity comparison. 
However, since some regions may be occluded or outside the field of view in certain views, we assume that most regions requiring inpainting are visible in the reference view.
The reference image is warped to the inpainted region of the target view using its depth.
Based on the depth map $D_{j}$ from view $j$, the corresponding point $p$ in view $i$ can be computed as follows:
\begin{equation}
    p_{j \rightarrow i} \approx KR_iR_j^{-1}D_j(p_j)K^{-1}p_j
\end{equation}
where $K$ denotes the camera intrinsic parameters, and $R_i$, $R_j$ represent the extrinsic matrices of views $i$ and $j$, respectively. To handle occlusions that can lead to many-to-one matching, the pixel closest to the target camera is selected as the matching point using Z-buffering. Unmatched pixels after warping are filled via bilinear interpolation using nearby valid samples, enabling smooth propagation of image content.  

\subsubsection{Confidence evaluation}
\label{subsec:Method_3_evaluation}

Once the inpainted images are aligned to the same view, their similarity can be assessed.
As multi-view inpainted images are used to construct geometry that supports stable image warping, our focus is on geometric consistency with the reference image rather than subtle pixel-level differences.
For this reason, we use the perceptual similarity metric LPIPS~\cite{Zhang2018LPIPS}, which measures structural similarity between images. It provides a way to quantitatively assess visual resemblance in a manner closer to human perception and shows robustness to distortions or color differences, unlike traditional metrics (e.g., L2, PSNR, SSIM). We extract a patch region $P$ from an enlarged bounding region that fully contains the inpainted area in the two views being compared.
However, perceptual similarity also responds to variations in surface texture, which is not desirable for our purpose. To suppress fine textures on smooth surfaces, we apply strong bilateral filtering~\cite{Tomasi1998BilateralFilter} to the inpainted images, guided by the depth and normal maps derived from estimated depth. We denote the bilateral-filtered image as $I^{BF}$. With detailed texture flattened, the perceptual similarity is evaluated between the inpainting of the target view patch $I_{j}|_{P}$ and the warped reference patch $I_{ref \rightarrow j}|_{P}$, both restricted to the patch region.
\begin{equation}
\text{conf}_j = 1 - LPIPS(I_{ref \rightarrow j}^{BF}|_{P}, I_{j}^{BF}|_{P})
\end{equation}
The estimated confidence reflects the quality of both RGB and depth inpaintings of the view. Since RGB and depth inpaintings are evaluated as a pair, imprecise depth estimation or alignment that causes substantial distortion during warping can lead to low confidence despite the similarity in RGB inpainting. To minimize such cases, each view's confidence is periodically updated during inpaint-3DGS by refining the alignment between the estimated depth and the geometry progressively reconstructed during training.
Based on the confidence, each view is weighted and incorporated into training inpaint-3DGS.
Incorporating inpaintings from highly relevant views helps mitigate the limitations of single-view dependency by providing complementary information.

\subsection{Inpaint-3DGS}
\label{sec:Method_3DGS_inp}

As the final step, inpaint-3DGS is performed to reconstruct previously hidden regions based on the inpainted data.  

For the region outside the mask, the optimization largely follows the standard strategy of Initial 3DGS, minimizing the photometric loss based on pixel-wise color differences (Eq. (\ref{eq:loss_ocolor})) and applying depth regularization on estimated depth (Eq. (\ref{eq:depth_loss_bg})) to further improve the overall geometric quality. However, slight modifications are made to the loss formulation to handle the inpainting regions.

To reconstruct the inpainted scene, we incorporate losses designed to align the inpainted region with a user-selected reference image, promoting cross-view consistency.
We aim to build perceptually coherent geometry across multiple reliable inpainted views and then transfer the appearance of the reference image via warping. 
Thus, we introduce losses focusing on two complementary objectives: constructing the geometry and aligning the appearance with a reference image.
Please note that the loss is evaluated independently for each view, randomly sampled at each iteration during optimization. For notational simplicity, we omit explicit indices for the view in the formulation. The loss on the inpainted region is modulated by the inpainting weight $w_{inp}=\sigma(\alpha(\text{conf}-\beta))$, where $\sigma(\cdot)$ is the sigmoid function, and $\alpha, \beta \in R$ control the scale and offset of the input, respectively. 

The depth loss is formulated as follows, taking into account both the inpainted region $M$ and the original region $M^c$:
\begin{equation}\label{eq:loss_idepth2}
\begin{aligned} 
\mathcal{L}_{D}
&= \frac{1}{|D|} \sum_{p \in M^c} \left| D_{mono}(p) - \hat{D}(p) \right| \\
&\quad + \frac{w_{inp}}{|D|} \sum_{p \in M} \left| D_{mono}(p) - \hat{D}(p) \right| 
\end{aligned}
\end{equation}

The depth loss helps suppress floating artifacts caused by inconsistent inpainting input, promoting stable and consistent geometry reconstruction. However, the estimated depth is a rough guide for placing Gaussians near the ground-truth surface rather than indicating their precise positions. 
Commonly, neural rendering methods suffer from artifacts and ambiguities due to the lack of 3D cues and surface constraints. To mitigate this, some approaches~\cite{turkulainen2024dnsplatter} jointly supervise both depth and normal to produce visually and geometrically plausible 3D reconstructions suitable for mesh conversion. Normal regularization encourages Gaussians to better conform to the underlying scene geometry.
Accordingly, we employ a normal loss to enhance geometric consistency, computed on a random patch region $P$ whose center is randomly sampled from the inpainting region. 
\begin{equation}
\mathcal{L}_{iN} = w_{inp}(1 - CosSim(N_{mono}|_{P}, \hat{N}|_{P}))
\end{equation}
It is computed via cosine similarity between $N_{mono}|_{P}$ and $\hat{N}|_{P}$, the normals derived from estimated and rendered depth, respectively. 

For appearance supervision, we overlay the warped reference image onto the current view and use it as a pseudo-ground truth to compute LPIPS loss and color loss. 

However, when the constructed geometry is yet inaccurate, leveraging warped outputs as pseudo-ground truth can further amplify distortion. 
To prevent this issue, we selectively warp regions based on the gap between the rendered depths, inspired by the approach of GeCoNeRF~\cite{Kwak2023GeCoNeRF}. GeCoNeRF warps input images to unseen views to perform regularization from sparse views. In this process, regions where the reprojected depth deviates from the corresponding view’s depth beyond a certain threshold are treated as occluded and excluded from the comparison. Following this idea, we exclude pixels from warping when the gap between the two aligned depths exceeds a threshold, treating them as geometrically inconsistent regions. As the training progresses and the geometry improves, the warped area progressively expands (see Fig.~\ref{fig:overview}).
Formally, we define the set of geometrically consistent pixels $P'$ as:
\begin{equation*}
P' = \left\{  p \in P \cap M \mid |\hat{D}_{\text{ref}}(p) - \hat{D}_{\text{tar} \to \text{ref}}(p)| \leq \tau \right\}
\end{equation*}
where $\hat{D}_{\text{tar} \to \text{ref}}$ is the rendered depth from the target view reprojected to the reference view, and $\tau$ is the threshold for consistency.
Using the set $P'$, the warped reference image $I_{\text{ref} \to \text{tar}}$ is blended into the target view $I_{\text{tar}}$. The resulting image $I_\text{warp}$ is defined as follows:
Within $P'$, $I_{\text{warp}}$ is obtained by solving the Poisson equation with the Laplacian of $I_{\text{ref} \to \text{tar}}$ as the guidance field, with Dirichlet boundary conditions imposed such that $I_{\text{warp}} = I_{\text{tar}}$ on the boundary of $P'$. Outside $P'$, $I_{\text{warp}}$ is set to $I_{\text{tar}}$. This formulation ensures that the inserted content blends smoothly into the surrounding region of the target image, preserving second-order structural consistency and eliminating visible seams~\cite{Perez2003Poisson}.

\begin{figure}[htb]
\centering
\includegraphics[width=0.45\textwidth]{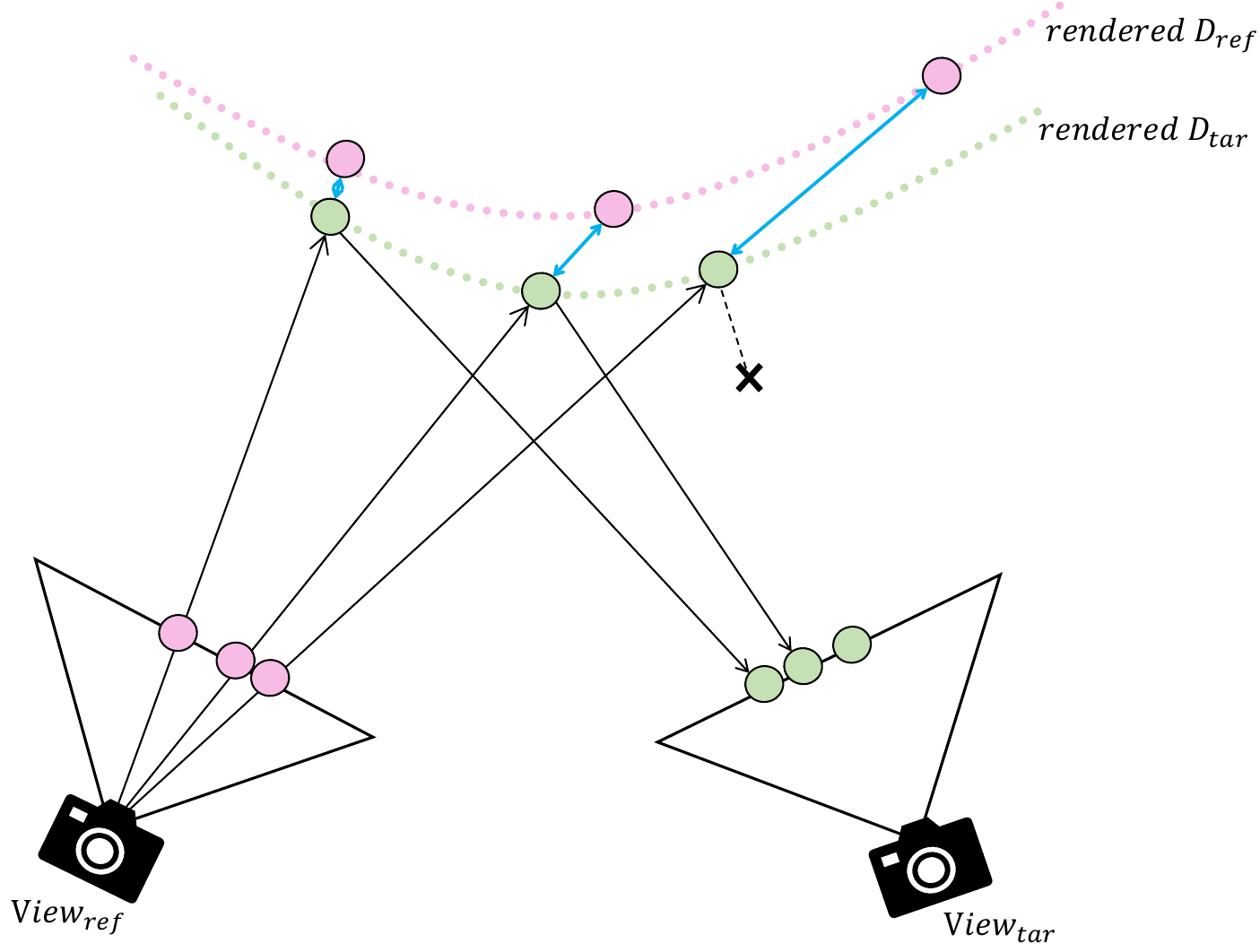}
\caption{Geometry-Consistent Warping. When warping a reference view to a target view, only regions with rendered depth differences below a threshold are deemed geometrically consistent and transferred. During optimization, warping is progressively applied, leaving out inconsistent regions.}
\label{fig:warp_consistent}
\end{figure}

Based on the warped image, LPIPS and color loss are utilized to optimize the image's overall appearance and fine details.
\begin{equation}\label{eq:loss_lpips}
\mathcal{L}_{ilpips} = w_\text{inp} \mathrm{LPIPS}(I_\text{warp}|_{P}, \hat{I}|_{P})
\end{equation}

\begin{equation}\label{eq:loss_icolor}
\begin{aligned} 
\mathcal{L}_{C} 
&= \frac{1 - \lambda_{\text{ssim}}}{|M^c \cup P'|} \sum_{p \in M^c \cup P'} \left| I_{warp}(p) - \hat{I}(p) \right| \\
&+ \lambda_{\text{ssim}} (1-\text{SSIM}(I_{warp}|_{M^c \cup P'}, \hat{I}|_{M^c \cup P'}))
\end{aligned}
\end{equation}
It should be noted that LPIPS loss, $\mathcal{L}_{ilpips}$, is computed over the entire inpainting patch region $P$ to help build a rough initial shape from the non-uniform inpainting. In contrast, the color loss is computed only over the warped regions $P'$ together with the original background region $M^c$ to refine the fine details of the warped reference information that are deemed geometrically consistent. 

Overall, the total loss for optimization is defined as follows: 
\begin{equation}\label{eq:total_loss}
\mathcal{L}_{\text{total}} = \mathcal{L}_{\text{C}} + \lambda_\text{D}\mathcal{L}_{\text{D}} + \lambda_{\text{iN}}\mathcal{L}_{\text{iN}} + 
\lambda_{\text{iLPIPS}}\mathcal{L}_{\text{iLPIPS}}
\end{equation}
Such an approach refines the intended geometry of the inpainted regions using depth and normal estimates weighted by reliability, while simultaneously guiding the appearance toward the reference through comparisons with progressively warped images, resulting in a realistic and perceptually consistent 3D scene.

\section{Experiments}
In this section, we evaluate our method on forward-facing and wide-view datasets. Our method is compared to existing 3D scene inpainting methods, followed by an ablation test.

\begin{table*}[!ht]
\center
\caption{Comparison of baseline method components related to inpainted region construction, and our experimental settings for fair comparison. Specifically, SPIn-NeRF (reimpl.) was adapted to 3DGS and evaluated using the shared mask and inpainted images generated via 3D segmentation in 3DGS.}
\begin{tabularx}{\textwidth}{llllll}
\toprule
\multicolumn{2}{c}{Category} & \multicolumn{1}{c}{SPIn-NeRF (reimpl.)} & \multicolumn{1}{c}{Gaussian Grouping} & \multicolumn{1}{c}{Infusion} & \multicolumn{1}{c}{Ours} \\
\midrule
\multirow{5}[+30]{*}{\begin{tabular}[c]{@{}l@{}}Inpainting-Related\\Method Details\end{tabular}} & Rendering Base & \begin{tabular}[c]{@{}l@{}} 3DGS \\ (Original: NeRF) \end{tabular} & 3DGS   & 3DGS  & 3DGS \\
\cmidrule(lr){2-6}
& 3D segmentation       & \begin{tabular}[c]{@{}l@{}} Yes \\(Original: None) \end{tabular}   & Yes  & Yes & Yes   \\
\cmidrule(lr){2-6}
& View usage            & \begin{tabular}[c]{@{}l@{}} All \\(weight: uniform) \end{tabular} &   \begin{tabular}[c]{@{}l@{}} All \\(weight: uniform) \end{tabular} & \begin{tabular}[c]{@{}l@{}} Reference views \\ only \end{tabular}& \begin{tabular}[c]{@{}l@{}} All + Reference \\ (weight: confidence) \end{tabular}\\
\cmidrule(lr){2-6}
& \begin{tabular}[c]{@{}l@{}} Reference \\Integration \end{tabular}& N/A    & N/A  & \begin{tabular}[c]{@{}l@{}} Estimated \\ depth-based \\ reprojection \end{tabular} &   \begin{tabular}[c]{@{}l@{}} Optimized\\depth-based\\warping\end{tabular}\\
\cmidrule(lr){2-6}
& Optimization          & \begin{tabular}[c]{@{}l@{}}- LPIPS loss\\ - Depth loss\end{tabular} & - LPIPS loss & - Color loss & \begin{tabular}[c]{@{}l@{}}- Color loss\\ - LPIPS loss\\ - Depth loss\\ - Normal loss\end{tabular} \\
\midrule
\multicolumn{1}{c}{\multirow{3}[+7]{*}{\begin{tabular}[c]{@{}l@{}} Experimental \\ Setting \end{tabular}}} & Mask & \multicolumn{4}{c}{SAM2 on our initial 3DGS render} \\
\cmidrule(lr){2-6}
\multicolumn{1}{c}{}& Inpainted image  & \multicolumn{4}{c}{SDXL inpainting on our initial 3DGS render}  \\
\cmidrule(lr){2-6}
\multicolumn{1}{c}{} & Estimated depth       & Depth anything v2 & N/A  & \begin{tabular}[c]{@{}l@{}} Diffusion-based \\ completion \\ (Infusion)\end{tabular}  & Depth anything v2 \\
\bottomrule
\end{tabularx}
\label{tab:baseline_comparison}
\end{table*}

\subsection{Implementation details}
\label{sec:experiment_implement}
Our code for initial 3DGS and inpaint-3DGS was built upon the vanilla 3DGS. For data preparation, three tasks are required in our method: 2D segmentation, image inpainting, and depth estimation.
For 2D segmentation, we utilized Segment Anything Model 2 (SAM2)~\cite{Ravi2024SAM2} to generate segmentation masks of the target object. SAM2 provides video segmentation capability, which aligns well with multi-view image sets typically used in neural rendering, as they exhibit overlapping regions resembling sequential frames. With a point prompt annotated on a single image, SAM2 can propagate the mask throughout the sequence. However, due to occasional loss of detail in complex geometries, we expand the masks to ensure better coverage.
For image inpainting, we used Stable Diffusion XL (SDXL)~\cite{Podell2023SDXL}, a latent diffusion model for text-to-image synthesis that shows robust performance due to its large U-Net backbone architecture compared to Stable Diffusion~\cite{Rombach2021StableDiffusion}.
Lastly, we used Depth Anything v2 (Metric Depth version)~\cite{Yang2024Depthanythingv2} for monocular depth estimation. In our method, all monocular depth estimates are first aligned to either COLMAP's sparse points or the rendered depth from 3DGS before use.
However, the models can be substituted with other alternatives. 
Our experiment was implemented on an NVIDIA A100 discrete GPU with 40GB of memory.

\subsection{Experiment Setups}
\label{sec:experiment_setup}

\textbf{Datasets} For experimental evaluation, we used a total of 10 sample scenes from three datasets: five from SPIn-NeRF~\cite{Mirzaei2023Spin}, two from MipNeRF360~\cite{Barron2022MipNeRF360}, and three from our custom dataset. For qualitative evaluation, we utilized the SPIn-NeRF dataset, which provides real-world images containing the removal target along with clean background images as ground truth. Among the scenes, we selected five with complex structures or textures that relatively showed larger inpainting discrepancies between views. Each scene comprises 60 training views containing the target object and 40 test views depicting the background after object removal. 
However, to demonstrate the effectiveness of our reference-guided 3D scene inpainting method, we use one image from the test set as the reference. Ultimately, training was performed on 61 views, with the remaining 39 views were used for evaluation.
However, the SPIn-NeRF dataset is limited to forward-facing scenes and thus insufficient to demonstrate robust performance across diverse viewpoints. Therefore, we additionally employed two 360$^{\circ}$ view real-world scenes from MipNeRF360 to visually showcase our performance across diverse viewpoints. Each scene contains 292 and 185 images, respectively. We split the views using 3/4 for training and the rest for testing.

Furthermore, to qualitatively assess the robustness of our method across a wide range of viewpoints and in complex scenes, we created a custom synthetic dataset of three scenes using Blender~\cite{blenderPackage} and assets from BlenderKit~\cite{blenderkit}. These scenes were designed to contain complex structures or textures, making it challenging to achieve consistent inpainting across views. Each scene contains a number of training and test views as follows: Scene 1 (75 / 33), Scene 2 (32 / 53), and Scene 3 (107 / 53).

\textbf{Baseline} 
Among related studies, we identified and selected the following methods as baselines based on their relevance to our approach. Firstly, SPIn-NeRF~\cite{Mirzaei2023Spin} as it is one of the most prominent approach for object removal in radiance fields. It optimizes the inpainted region using perceptual similarity, emphasizing the overall structure of the image rather than fine-grained inconsistencies, which is also adopted in our work. 

However, SPIn-NeRF is based on NeRF, and a notable performance gap exists between NeRF and 3DGS. Therefore, to compare under similar conditions, we re-implemented SPIn-NeRF on 3DGS by closely following the procedure described in the original paper. 

When comparing the LPIPS scores on the SPIn-NeRF sample datasets using the LaMa inpainting technique~\cite{Suvorov2022LaMa}, our 3DGS-based reimplementation achieved 0.394, outperforming the original implementation's 0.487. This lower perceptual error supports the validity of our comparison.

Additionally, Gaussian Grouping~\cite{Ye2024GaussianGrouping}, which minimizes the inpainting region by performing 3D segmentation in advance, was included as a baseline for comparison. The integration method is similar to SPIn-NeRF but does not utilize depth inpainting.
Finally, we included Infusion~\cite{Liu2024Infusion} in our comparison, which achieves multi-view consistency by reprojecting estimated depth from reference images onto the segmented 3D Gaussian scene. The Table~\ref{tab:baseline_comparison} summarizes key components related to inpainted region construction and experimental setup.

\textbf{Metrics} 
To evaluate the quality of 3D scene inpainting, we referred to the evaluation criteria used in prior studies on object removal in radiance fields~\cite{Weder2023Removing, Yin2023Ornerf, Mirzaei2023Spin, Mirzaei2023Reference,Liu2024Infusion}.
We mainly employed LPIPS~\cite{Zhang2018LPIPS}, FID~\cite{Heusel2017FID}, and SSIM~\cite{Wang2004SSIM} to assess perceptual similarity, distributional similarity, and structural consistency by comparing the rendered image with the ground truth image in the test set.

\subsection{3D Scene Inpainting Evaluation}

\begin{figure*}[htbp]
\centering
\begingroup
\setlength{\tabcolsep}{2pt}
\begin{tabular}{cccc @{\hspace{2mm}} cccc}
    \rotatebox{90}{\scriptsize SPIn-NeRF (reimpl.)} &
    \includegraphics[width=0.15\textwidth]{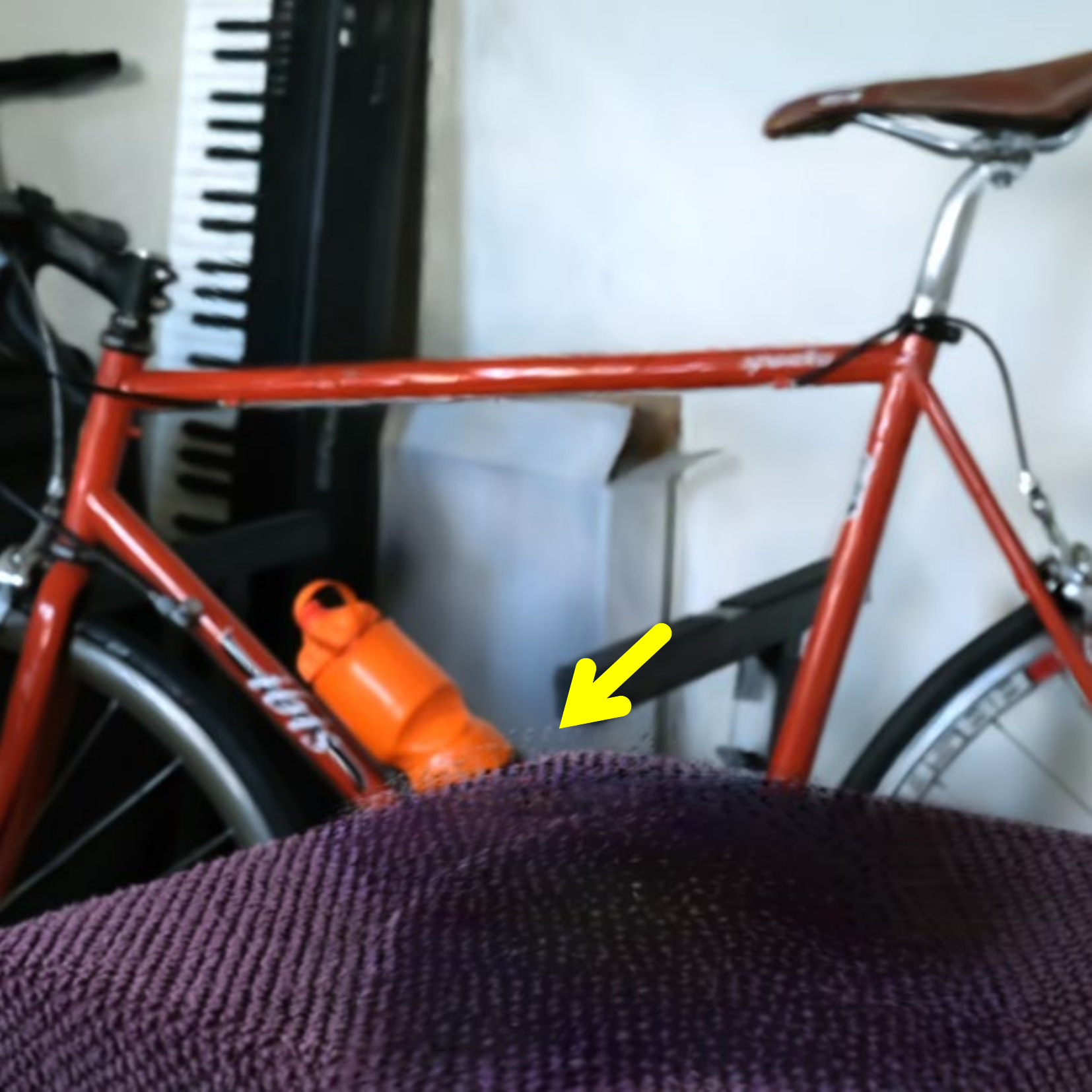} &
    \includegraphics[width=0.15\textwidth]{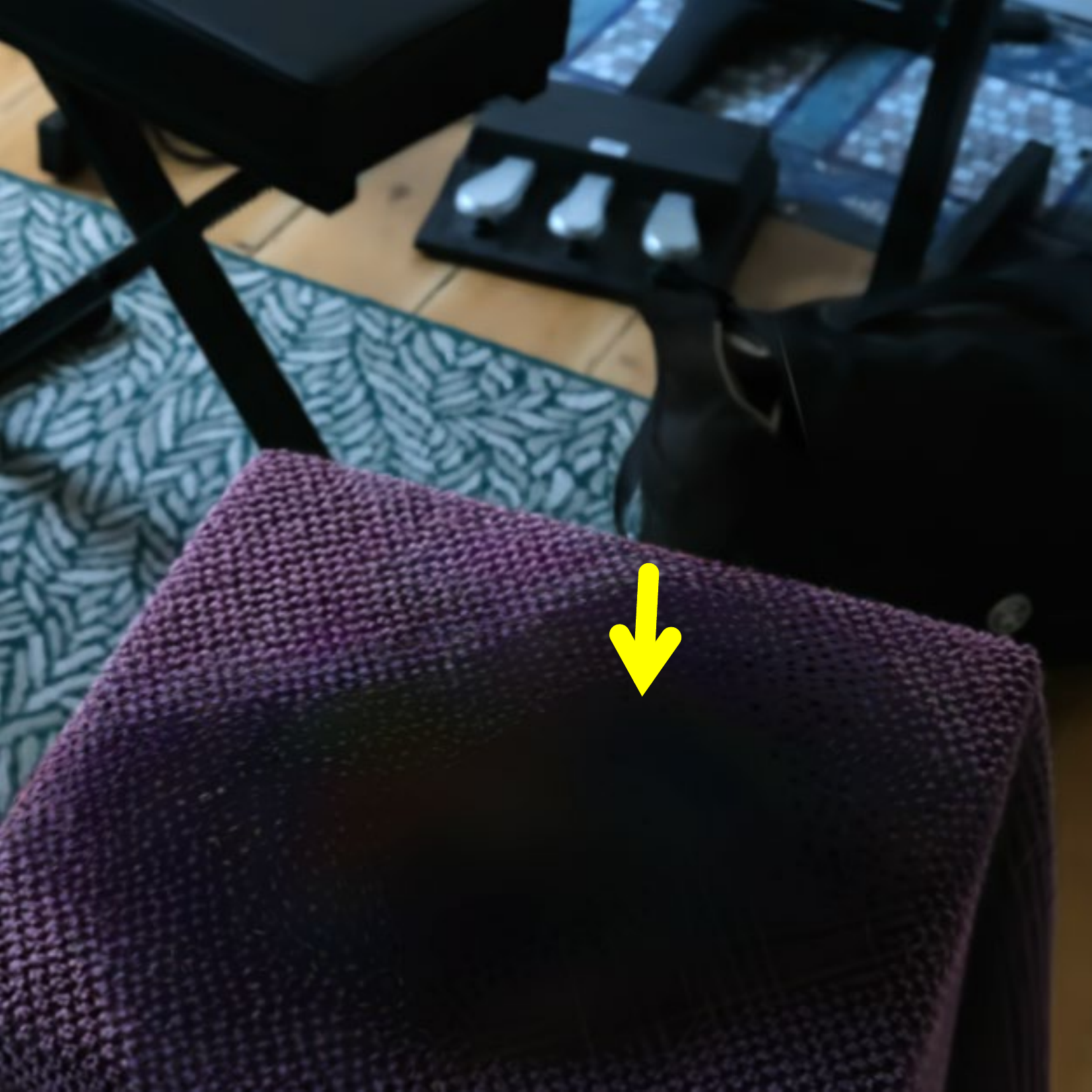} &
    \includegraphics[width=0.15\textwidth]{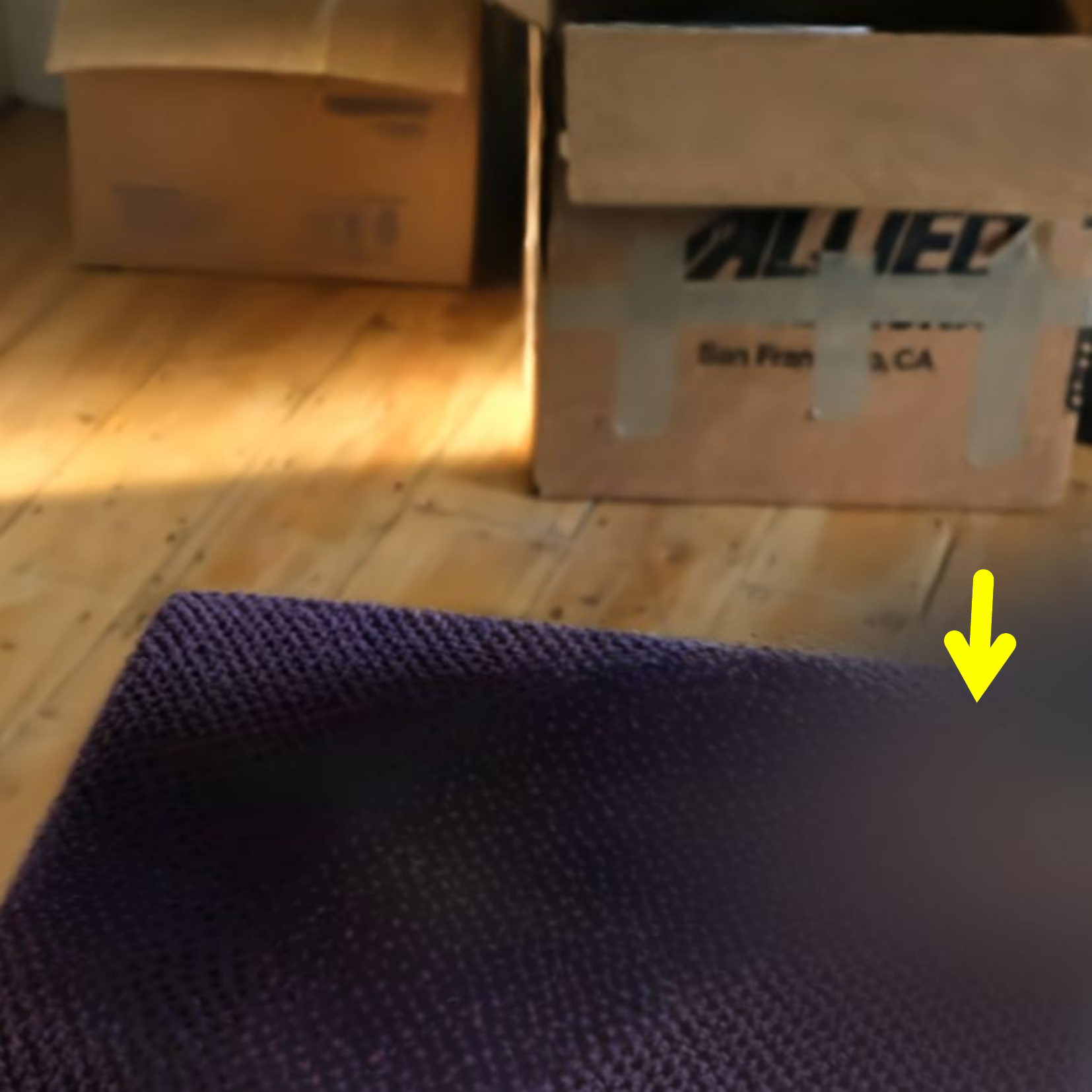} &
    \rotatebox{90}{\scriptsize Gaussian Grouping} &
    \includegraphics[width=0.15\textwidth]{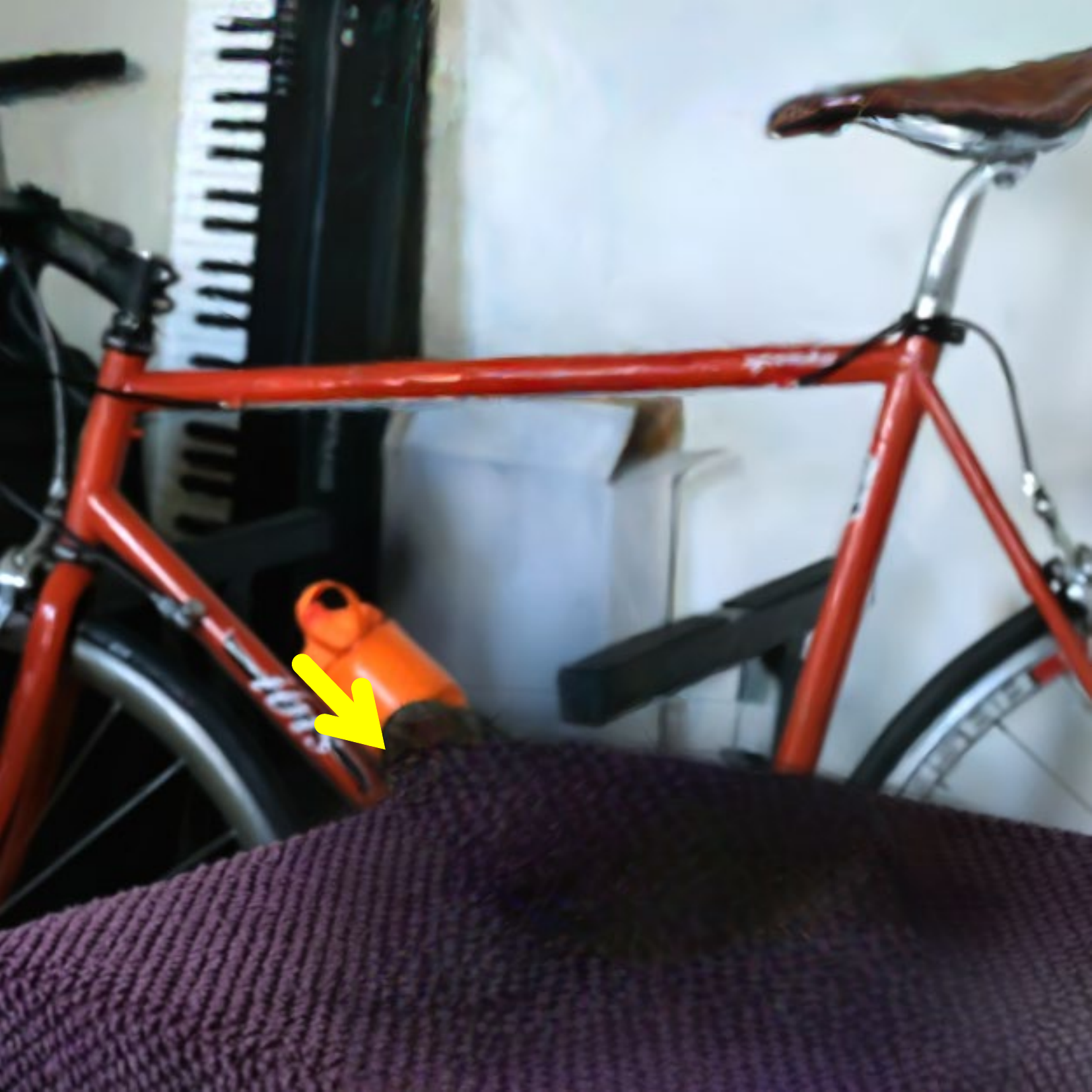} &
    \includegraphics[width=0.15\textwidth]{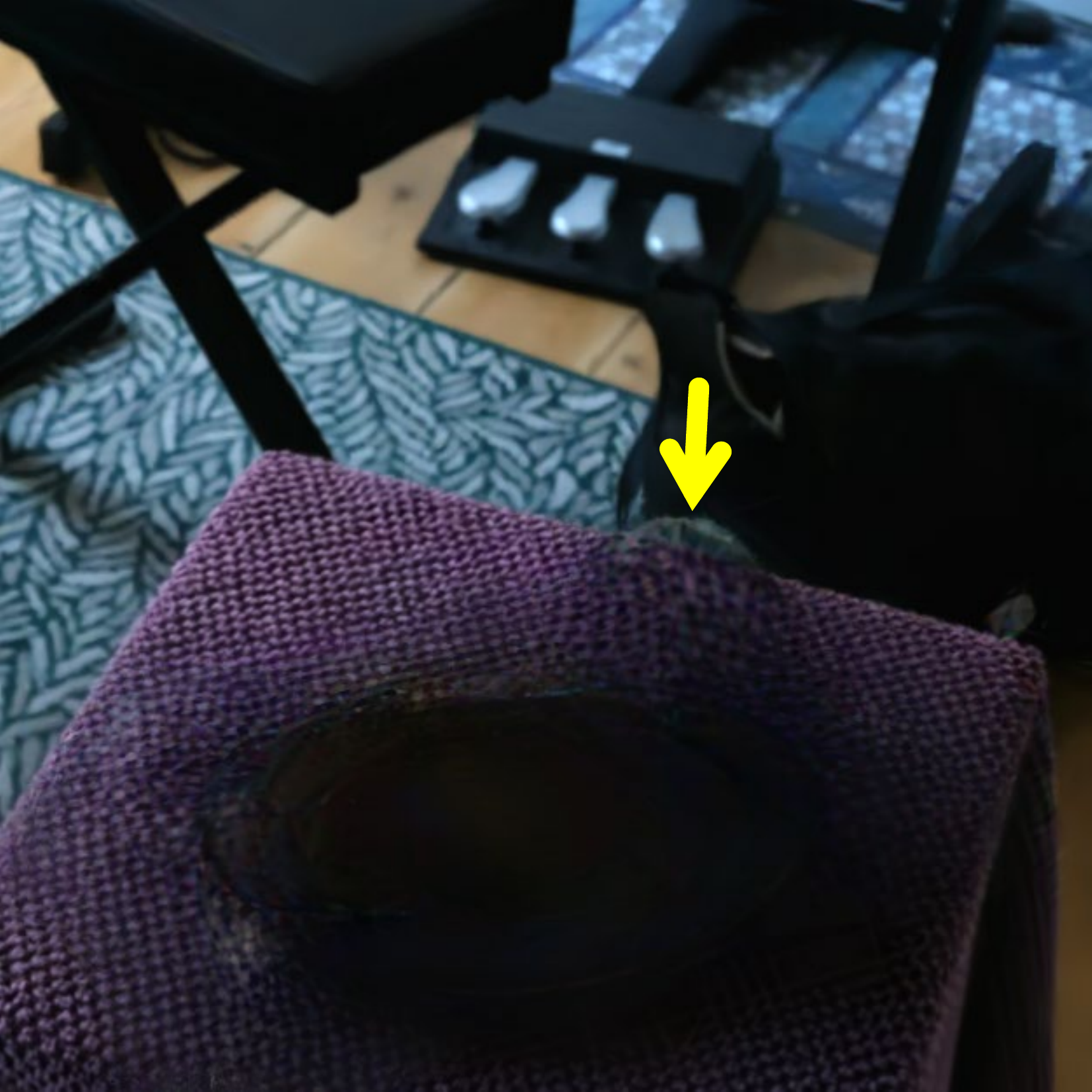} &
    \includegraphics[width=0.15\textwidth]{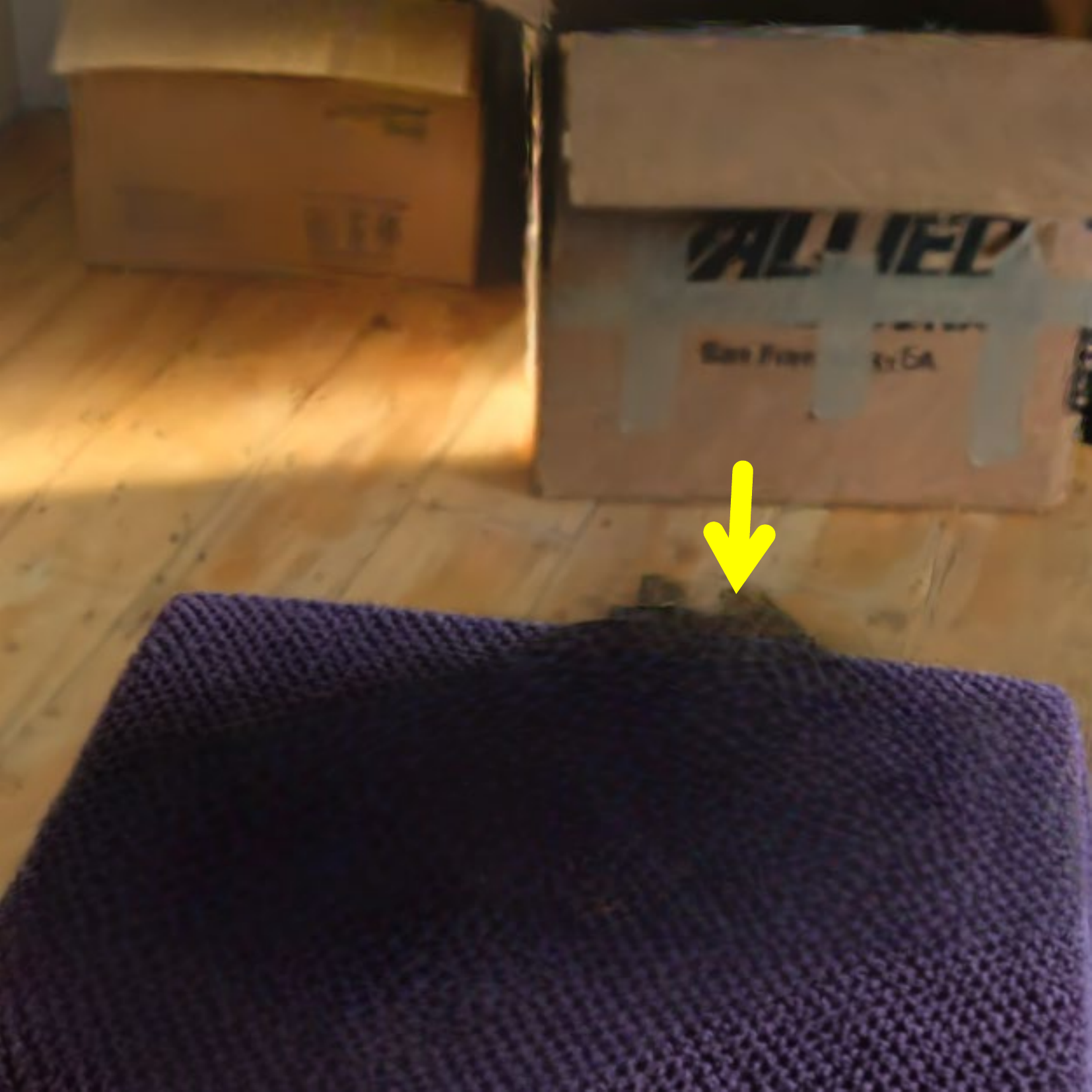} \\

    \rotatebox{90}{\scriptsize Infusion} &
    \includegraphics[width=0.15\textwidth]{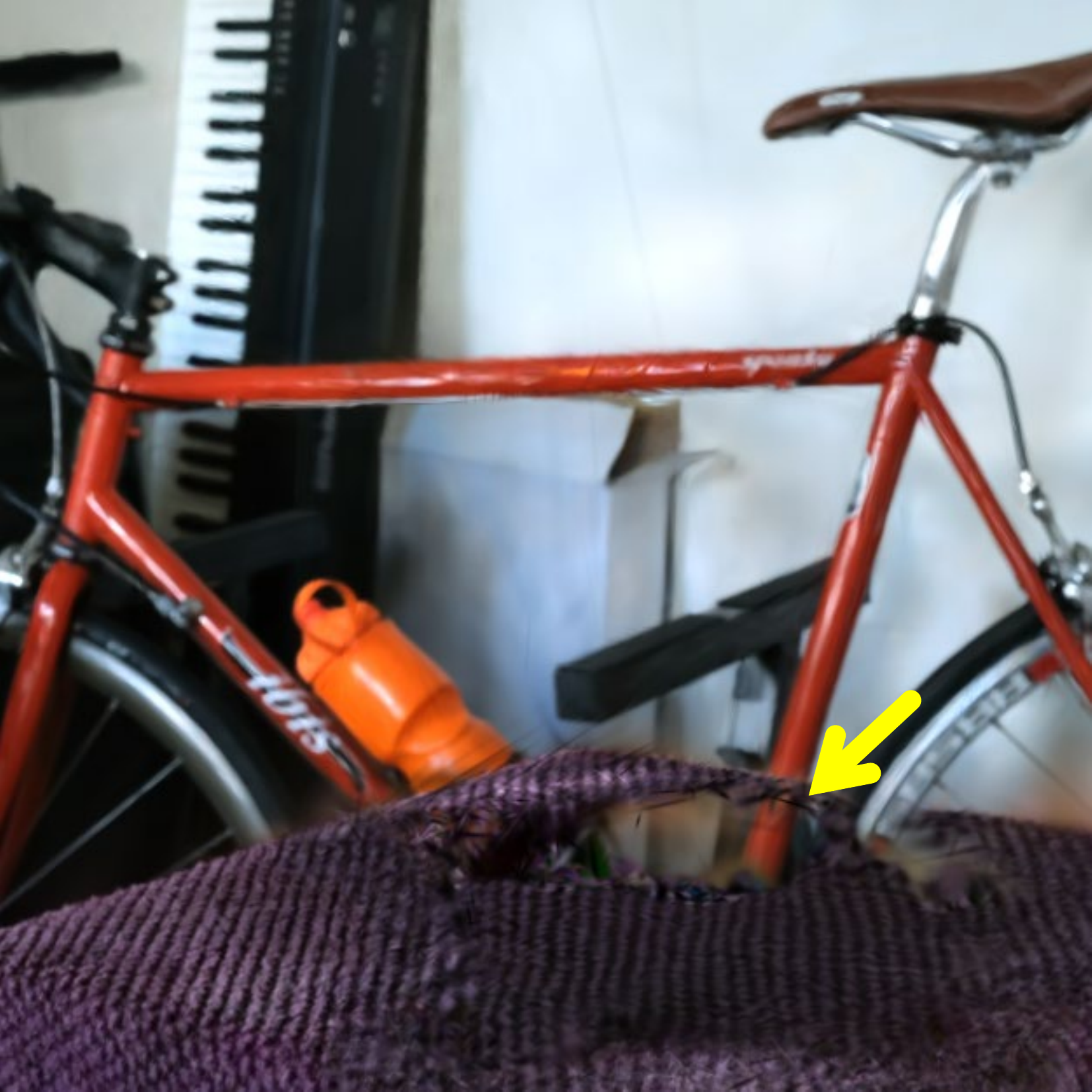} &
    \includegraphics[width=0.15\textwidth]{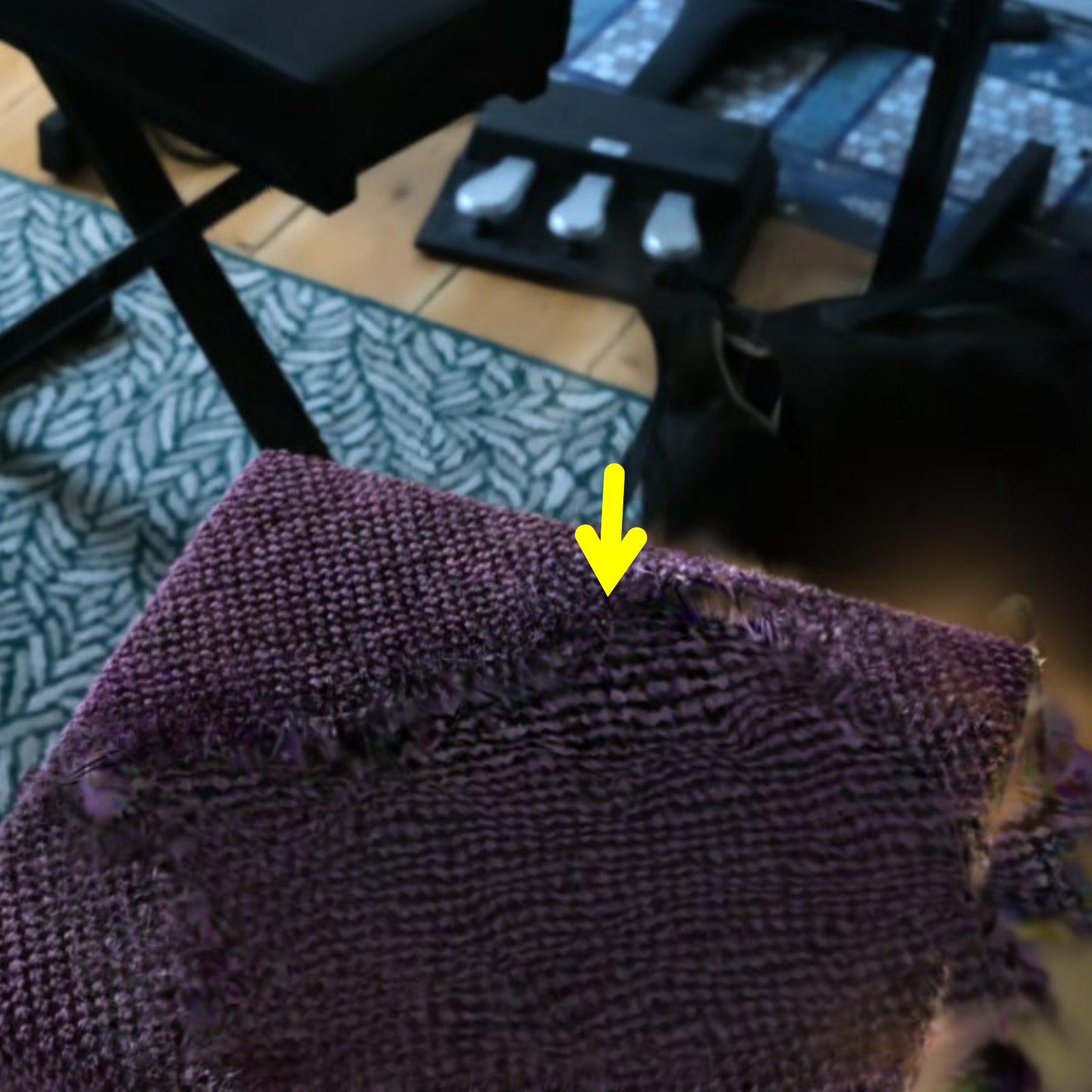} &
    \includegraphics[width=0.15\textwidth]{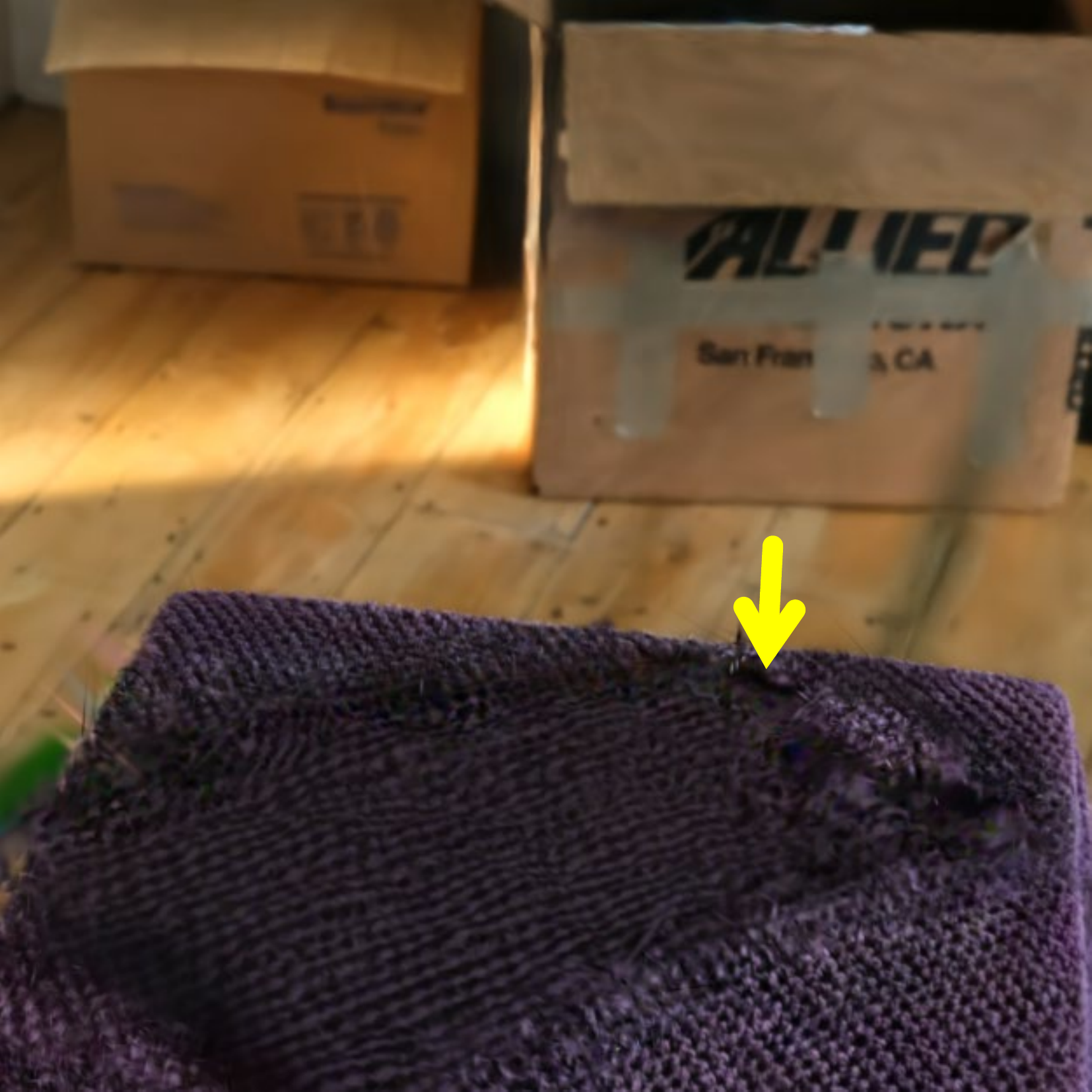} &
    \rotatebox{90}{\scriptsize Ours} &
    \includegraphics[width=0.15\textwidth]{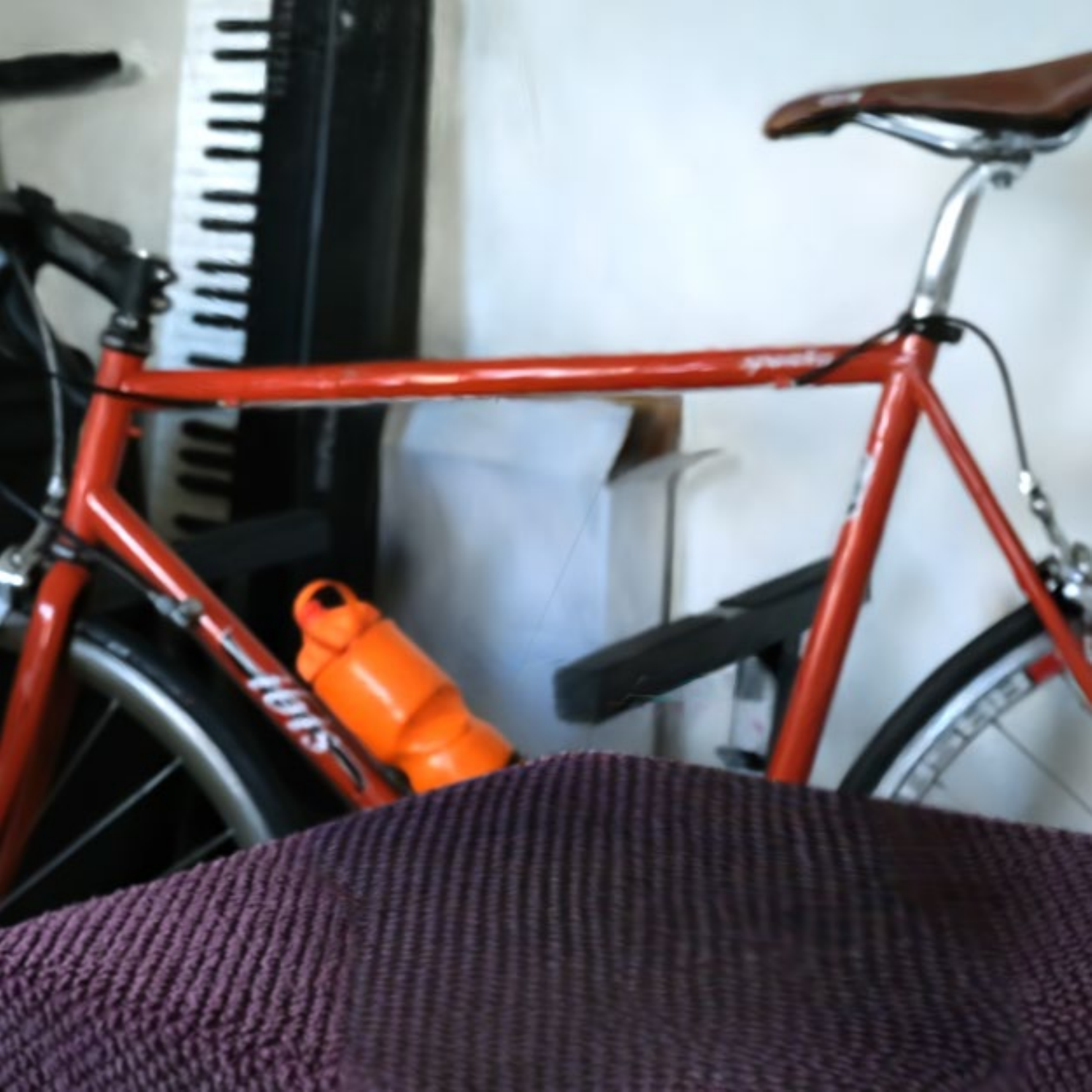} &
    \includegraphics[width=0.15\textwidth]{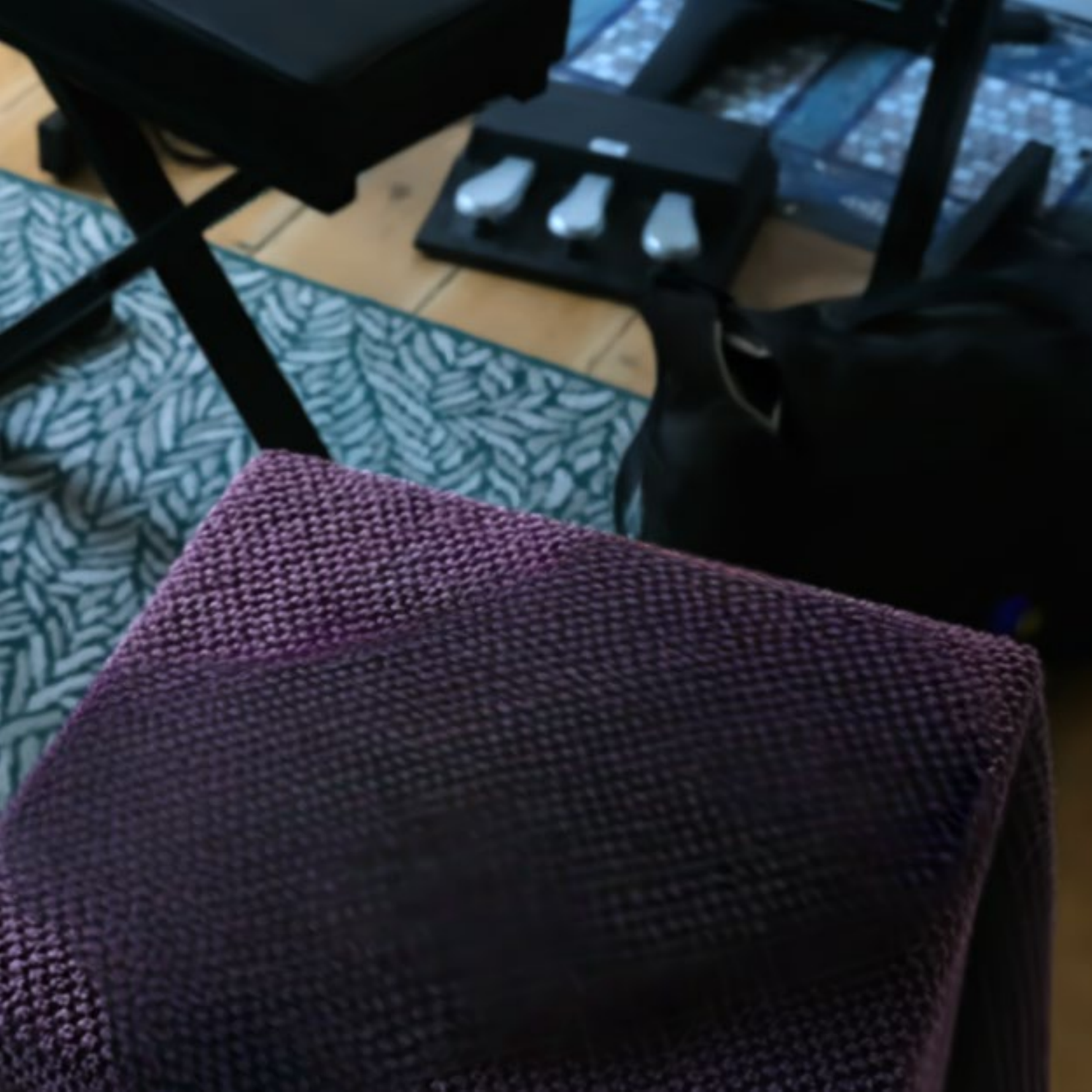} &
    \includegraphics[width=0.15\textwidth]{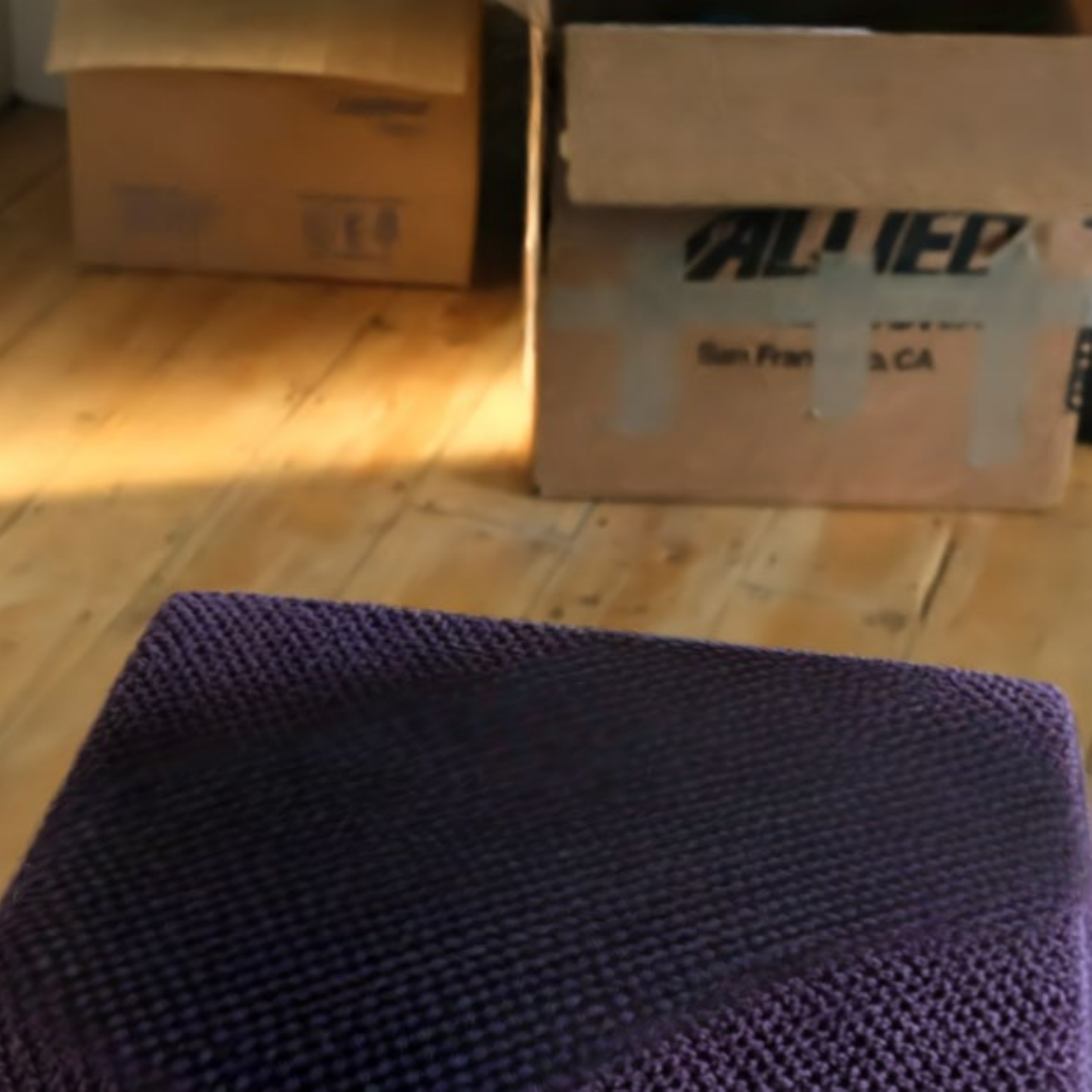} \\
\end{tabular}

\vspace{0.5mm}

\begin{tabular}{cccc @{\hspace{2mm}} cccc}
    \rotatebox{90}{\scriptsize SPIn-NeRF (reimpl.)} &
    \includegraphics[width=0.15\textwidth]{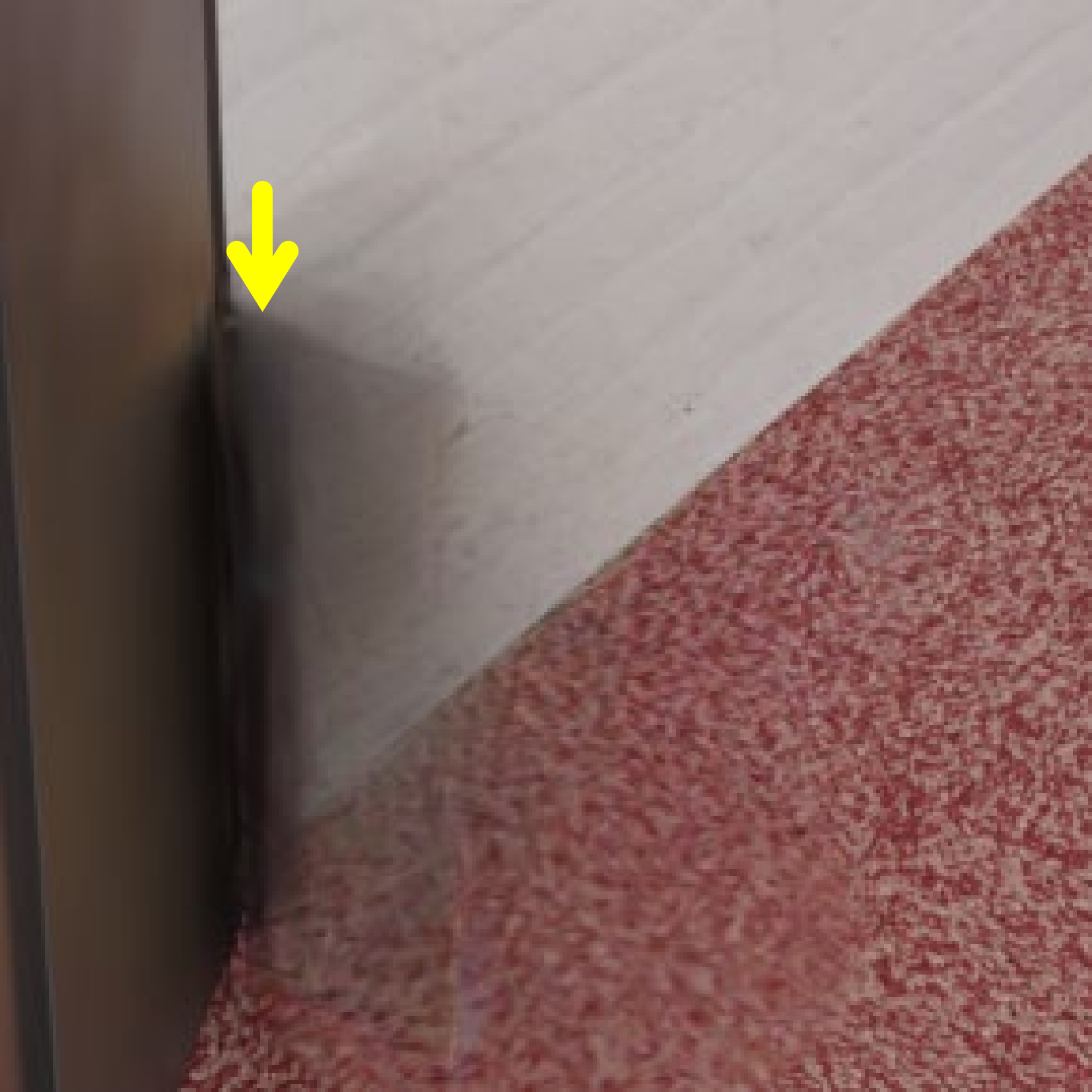} &
    \includegraphics[width=0.15\textwidth]{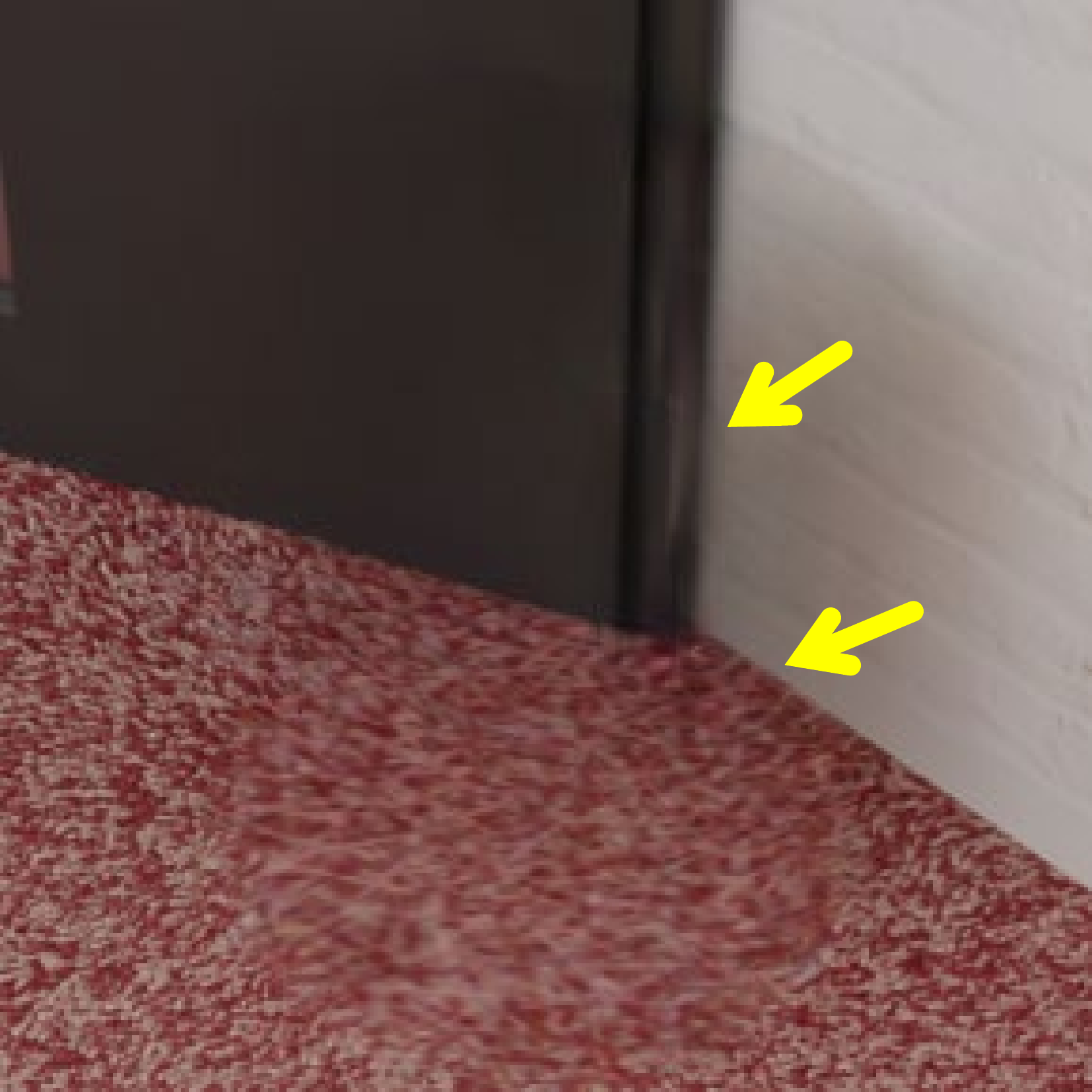} &
    \includegraphics[width=0.15\textwidth]{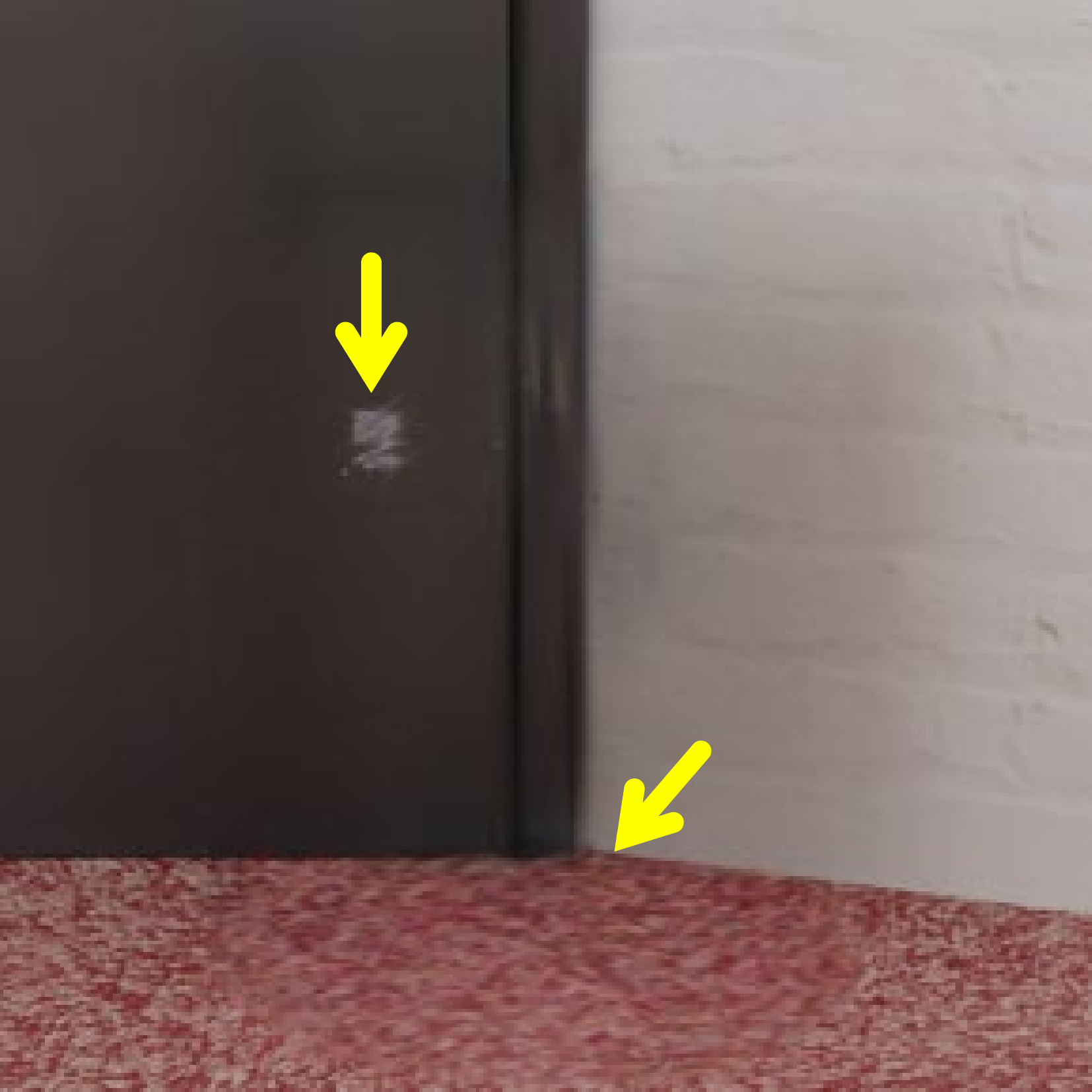} &
    \rotatebox{90}{\scriptsize Gaussian Grouping} &
    \includegraphics[width=0.15\textwidth]{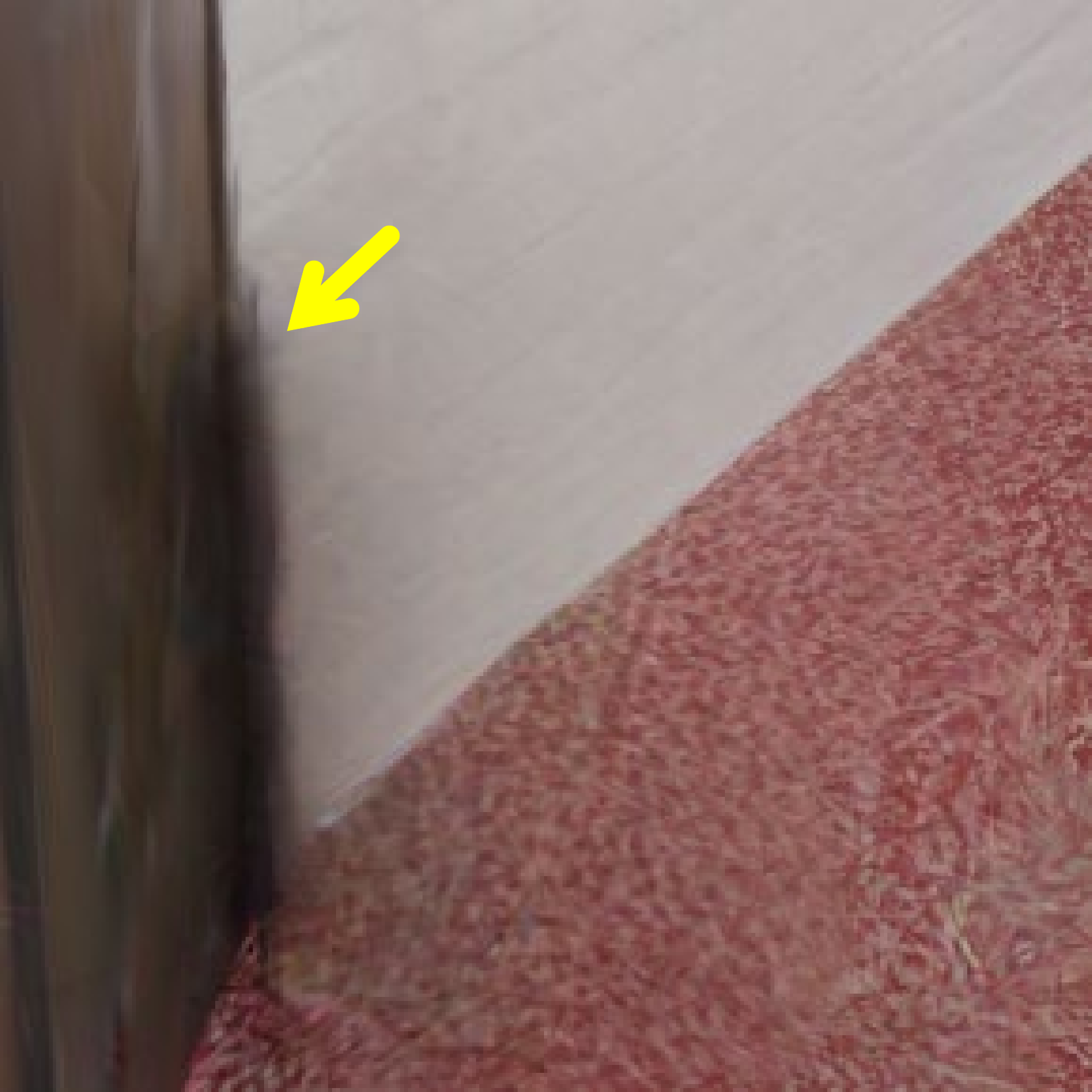} &
    \includegraphics[width=0.15\textwidth]{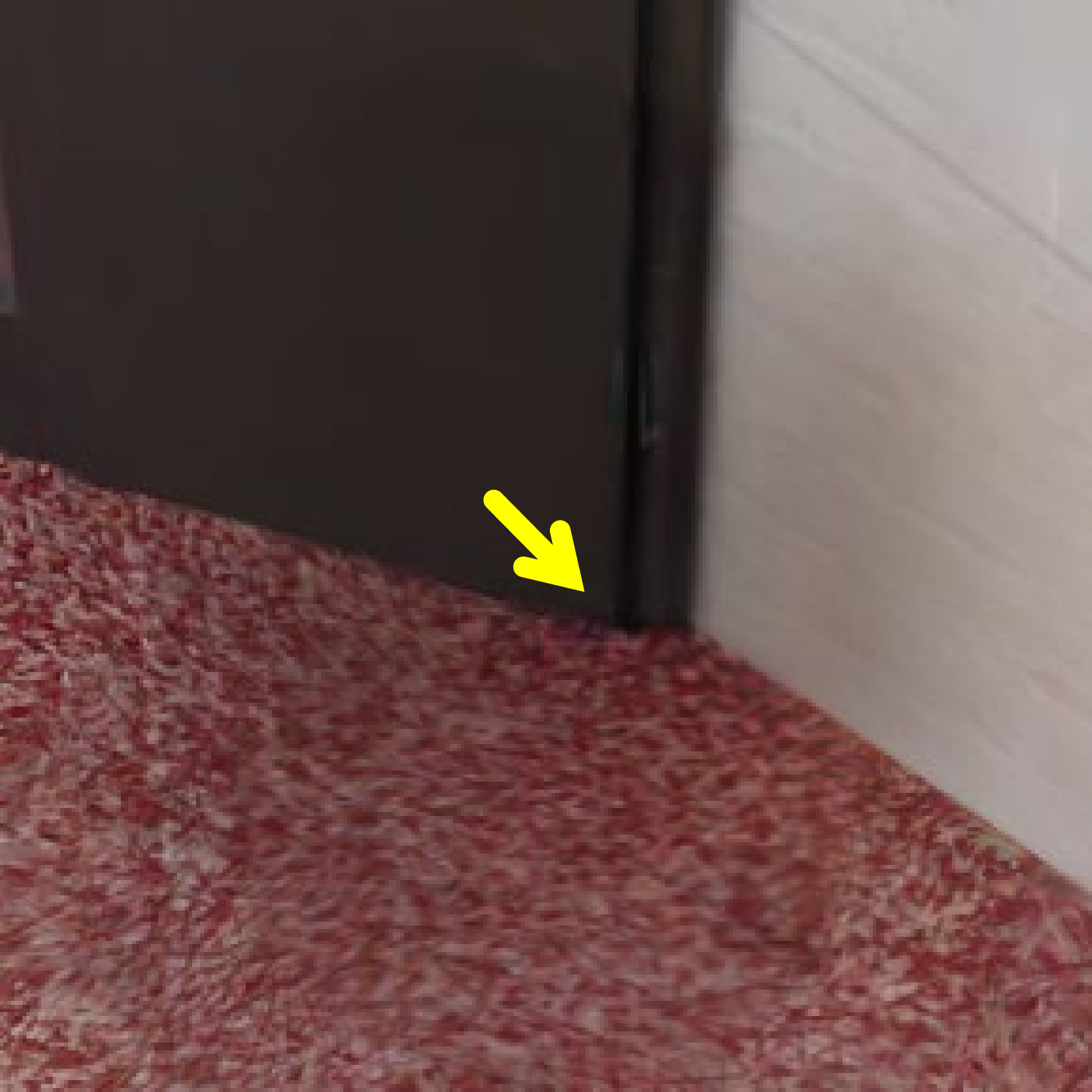} &
    \includegraphics[width=0.15\textwidth]{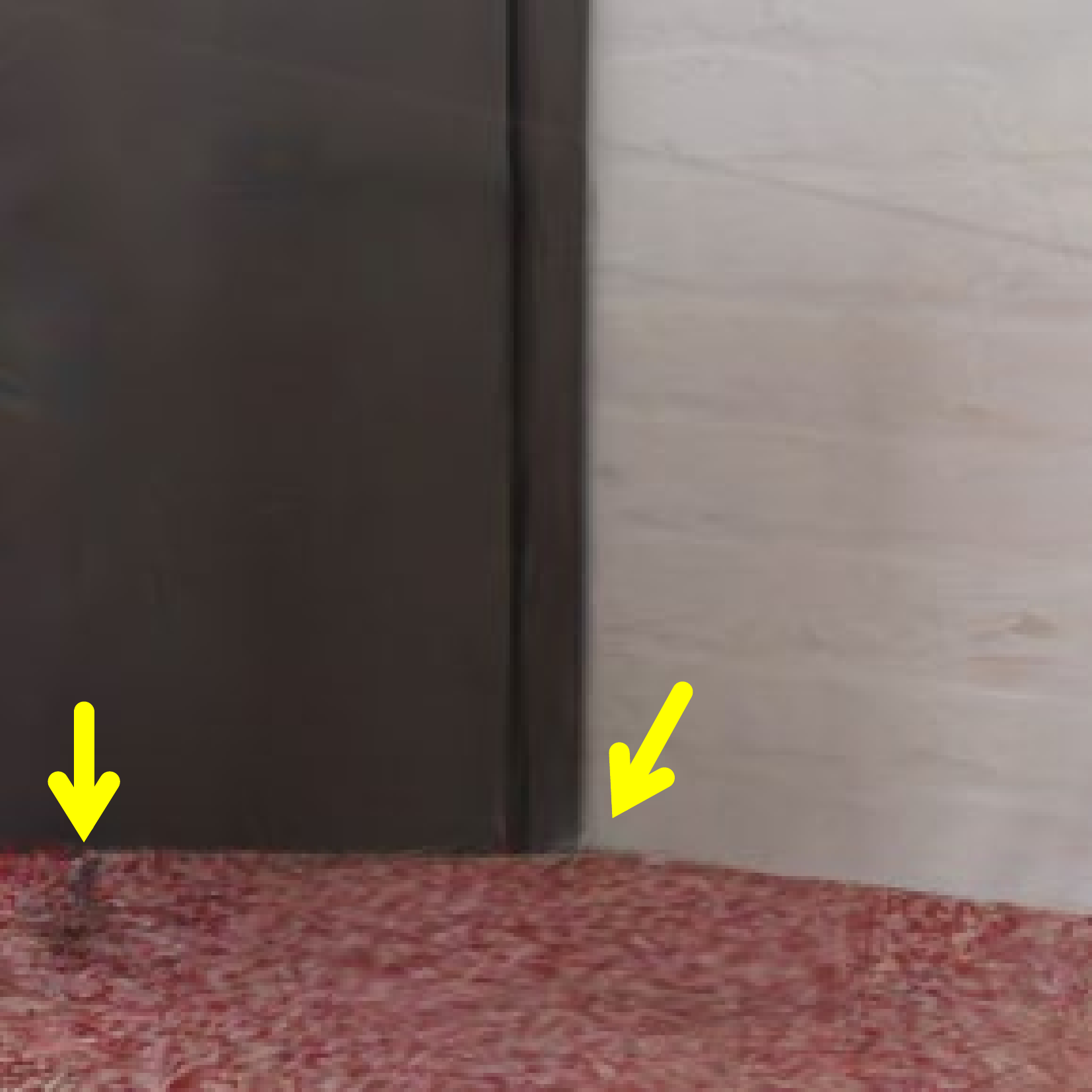} \\

    \rotatebox{90}{\scriptsize Infusion} &
    \includegraphics[width=0.15\textwidth]{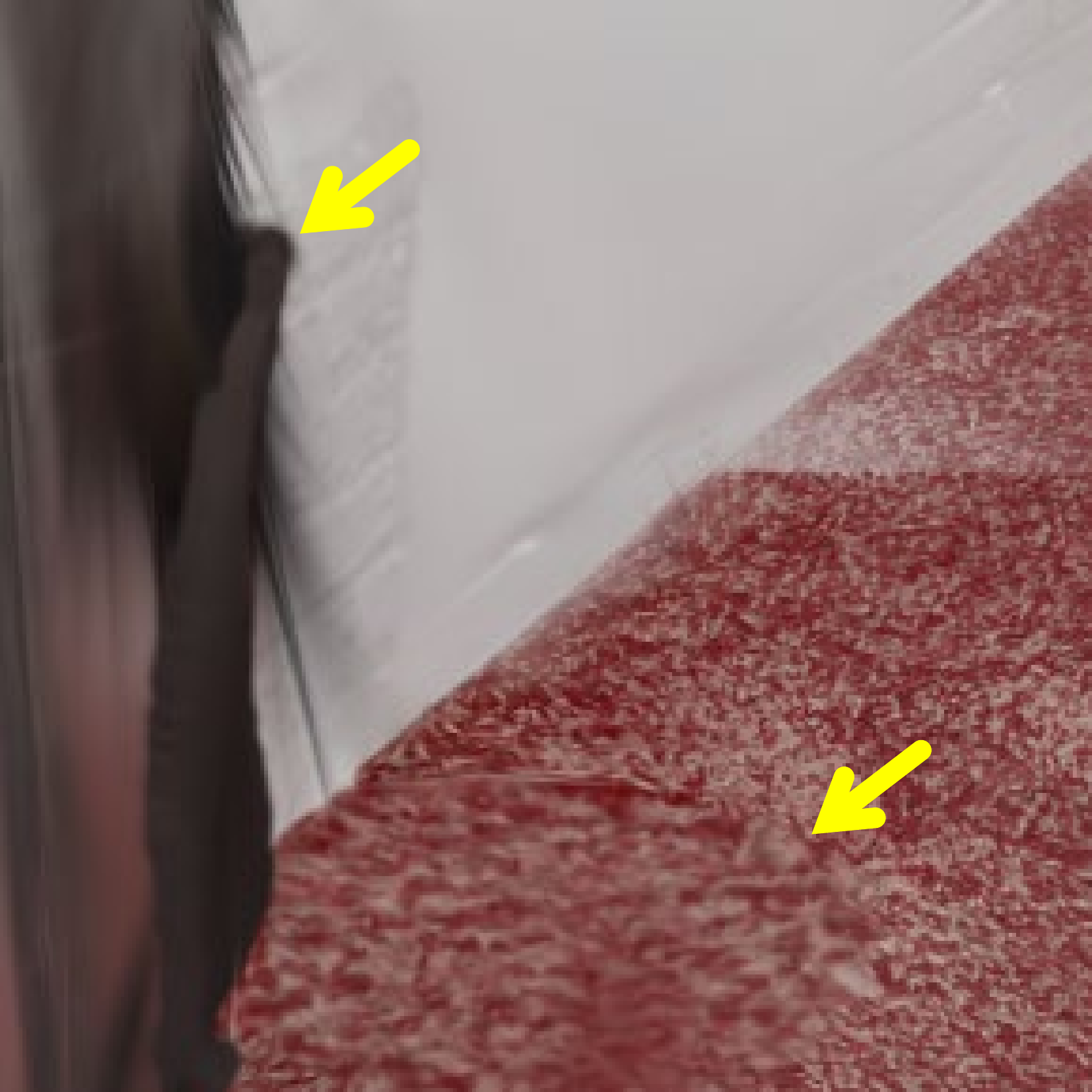} &
    \includegraphics[width=0.15\textwidth]{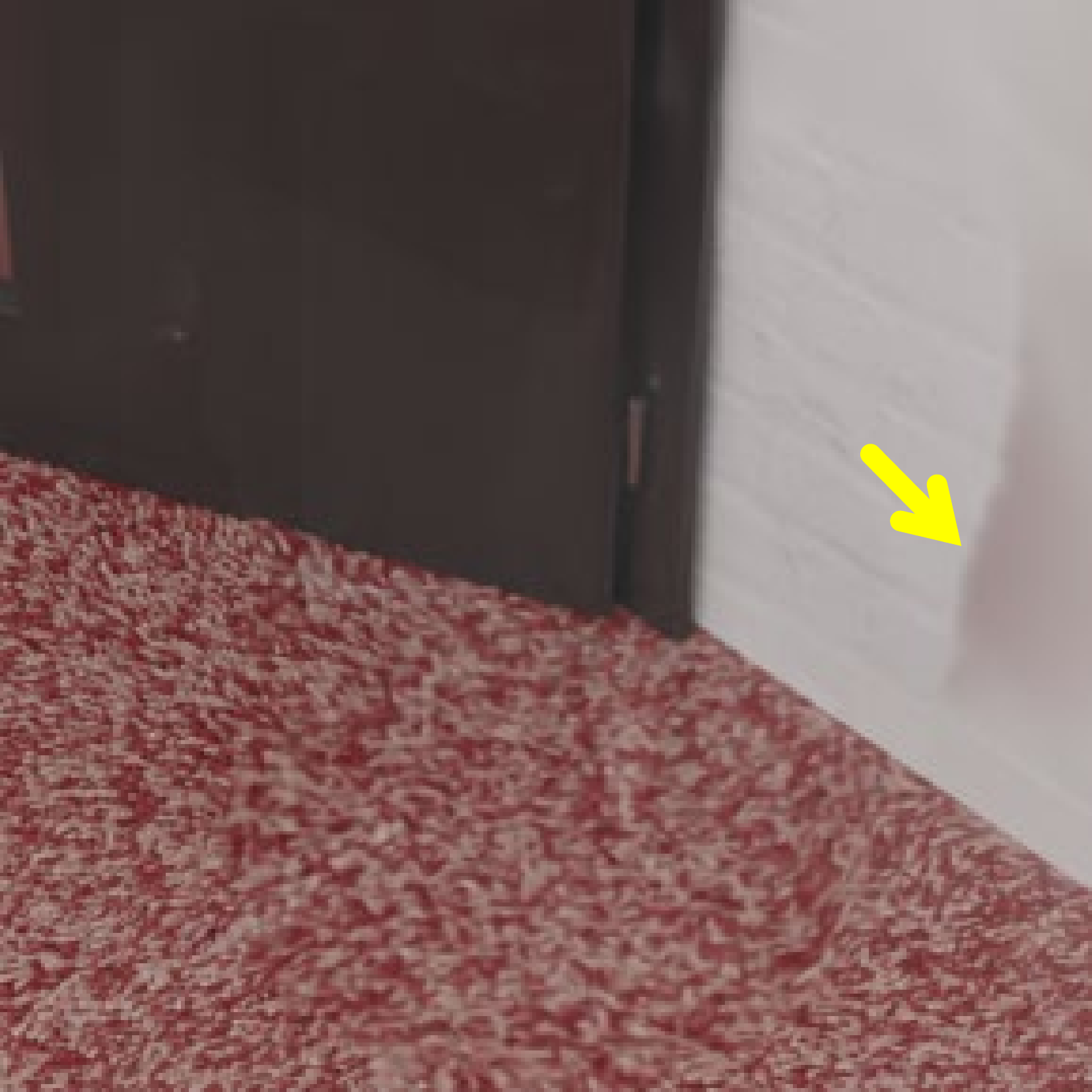} &
    \includegraphics[width=0.15\textwidth]{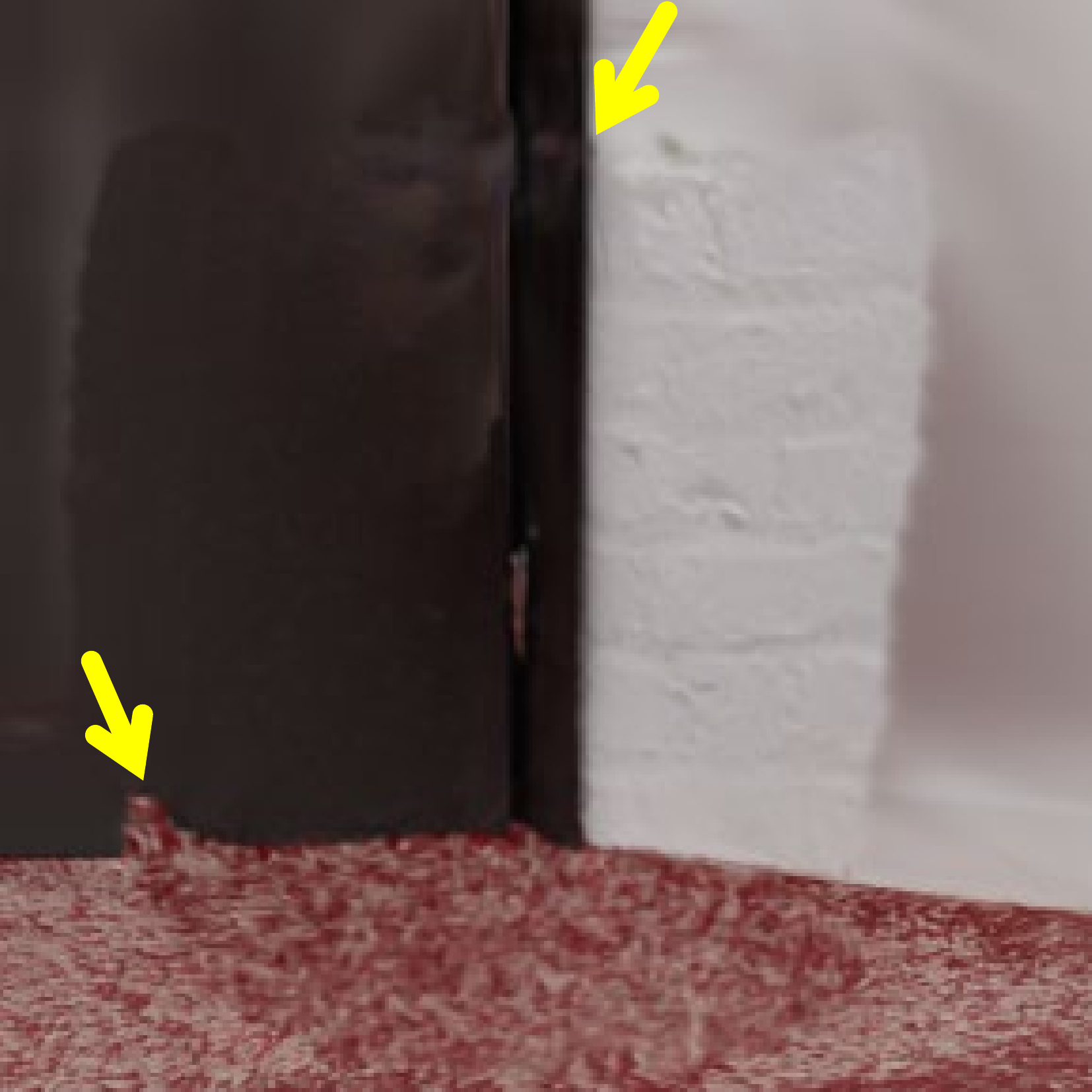} &
    \rotatebox{90}{\scriptsize Ours} &
    \includegraphics[width=0.15\textwidth]{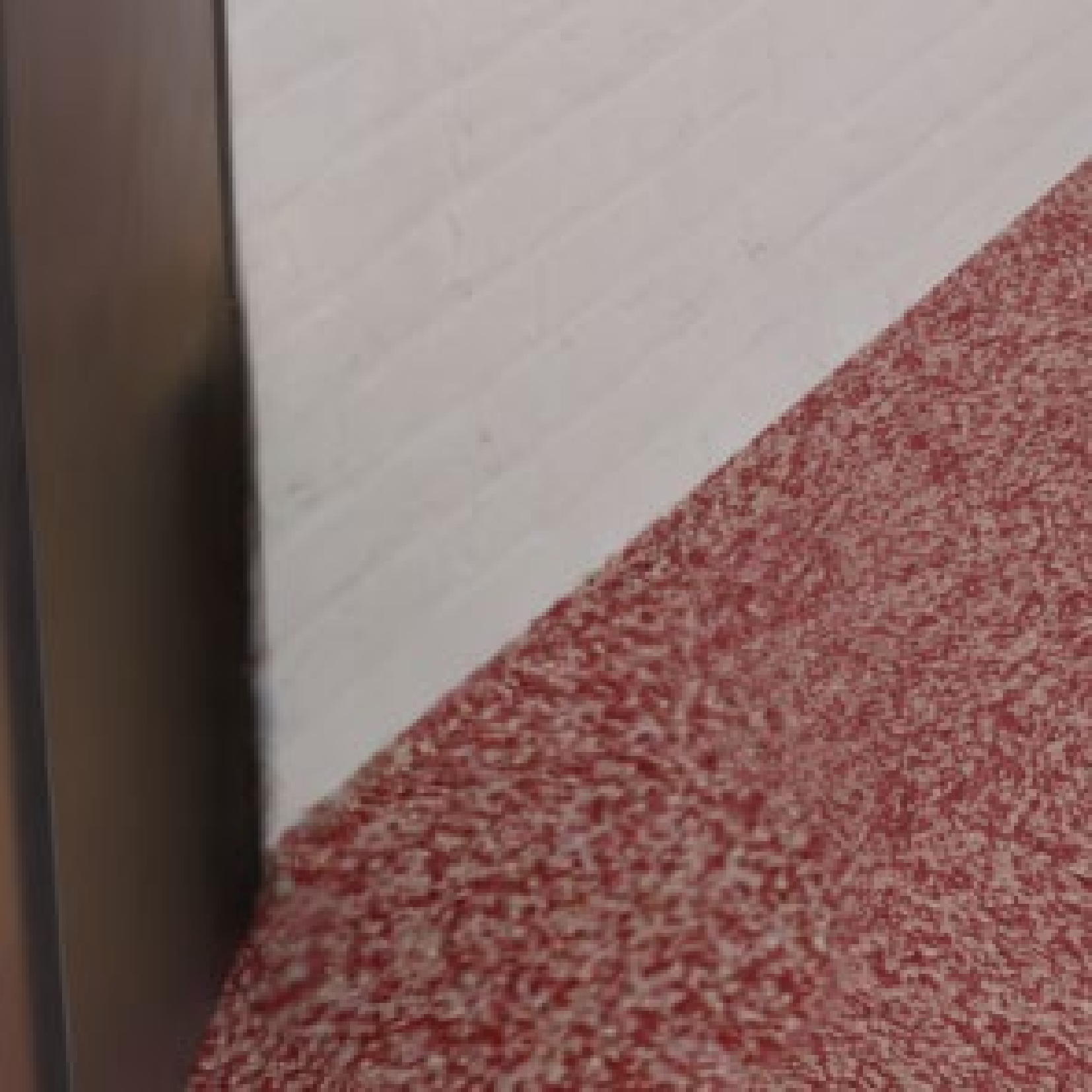} &
    \includegraphics[width=0.15\textwidth]{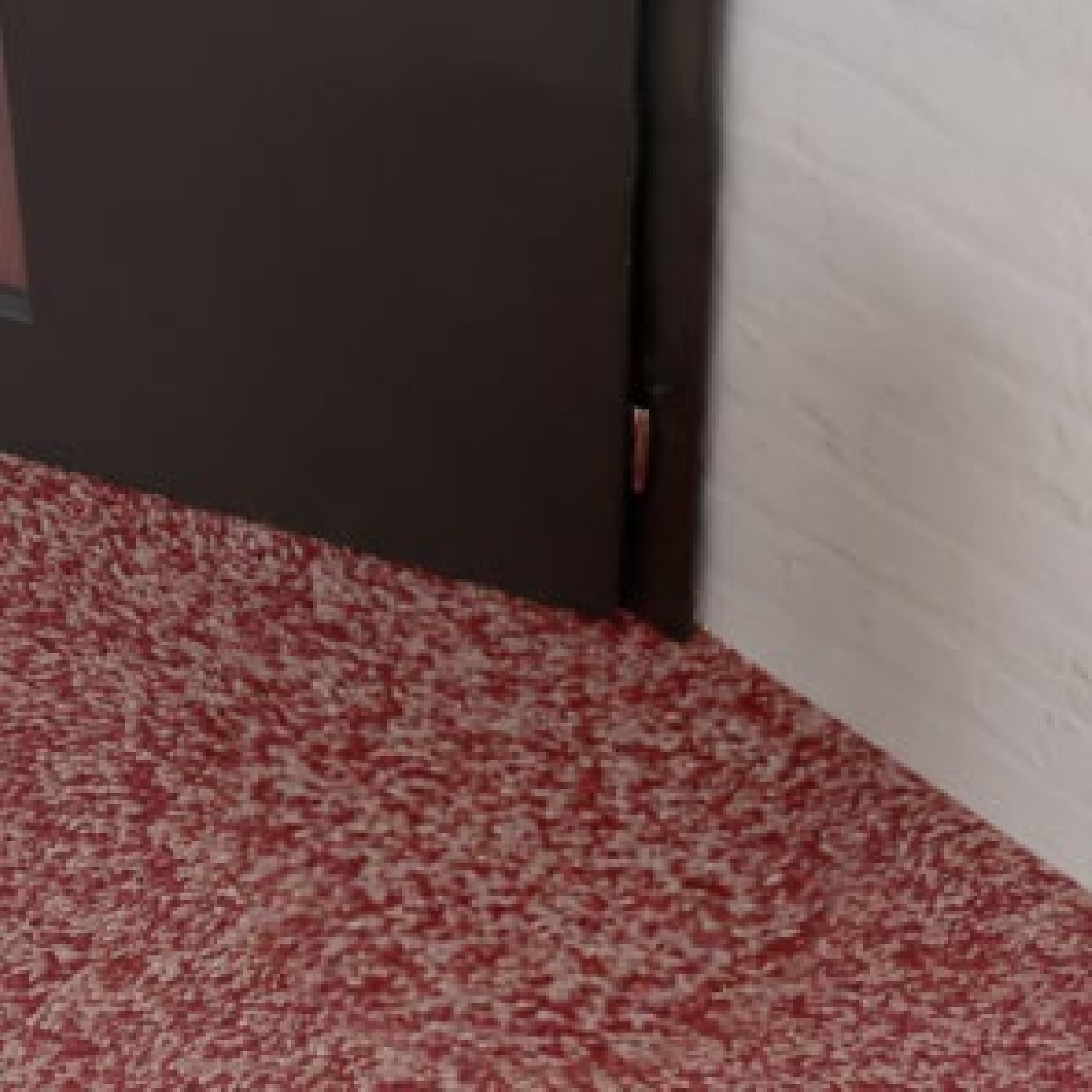} &
    \includegraphics[width=0.15\textwidth]{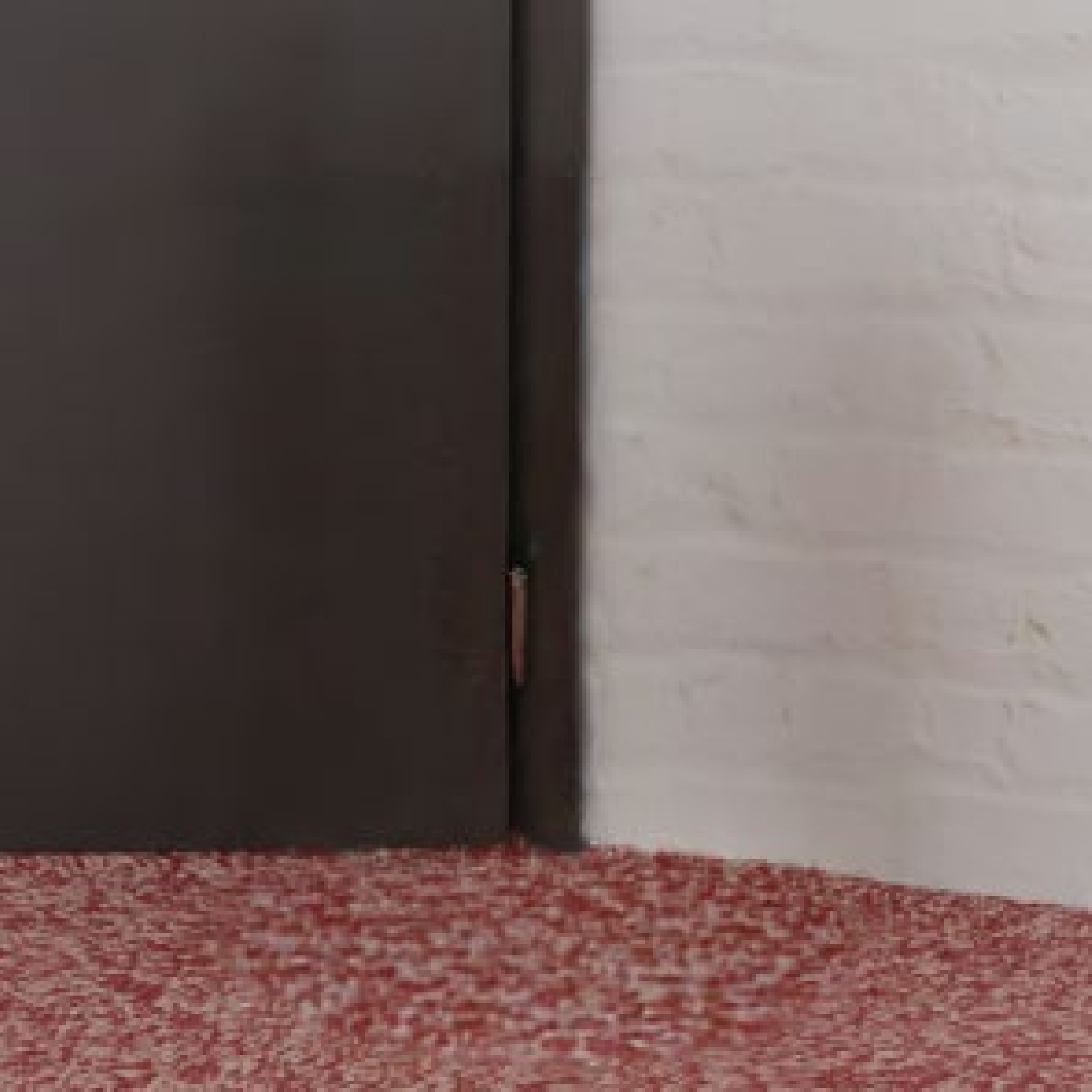} \\
\end{tabular}

\vspace{0.5mm}

\begin{tabular}{cccc @{\hspace{2mm}} cccc}
    \rotatebox{90}{\scriptsize SPIn-NeRF (reimpl.)} &
    \includegraphics[width=0.15\textwidth]{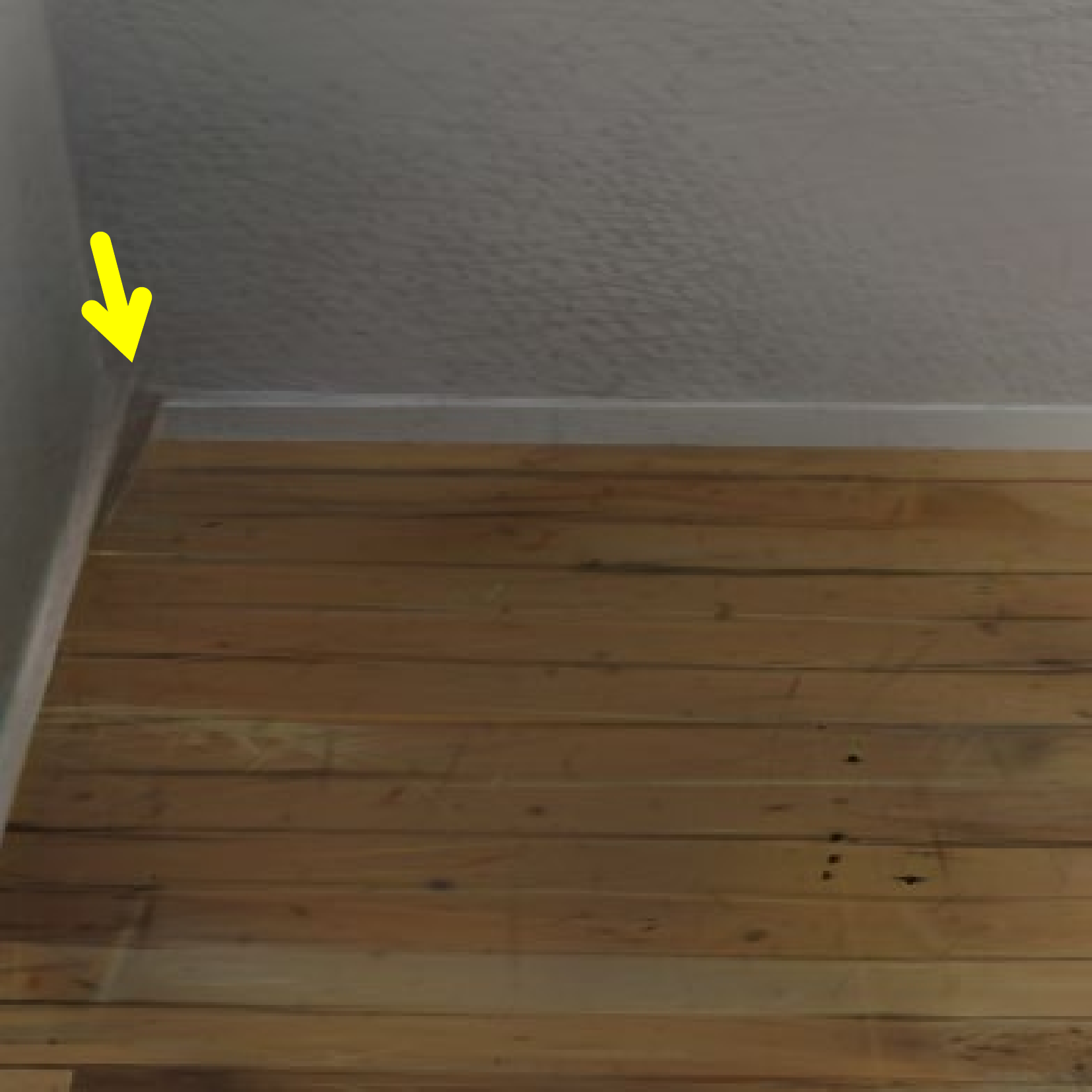} &
    \includegraphics[width=0.15\textwidth]{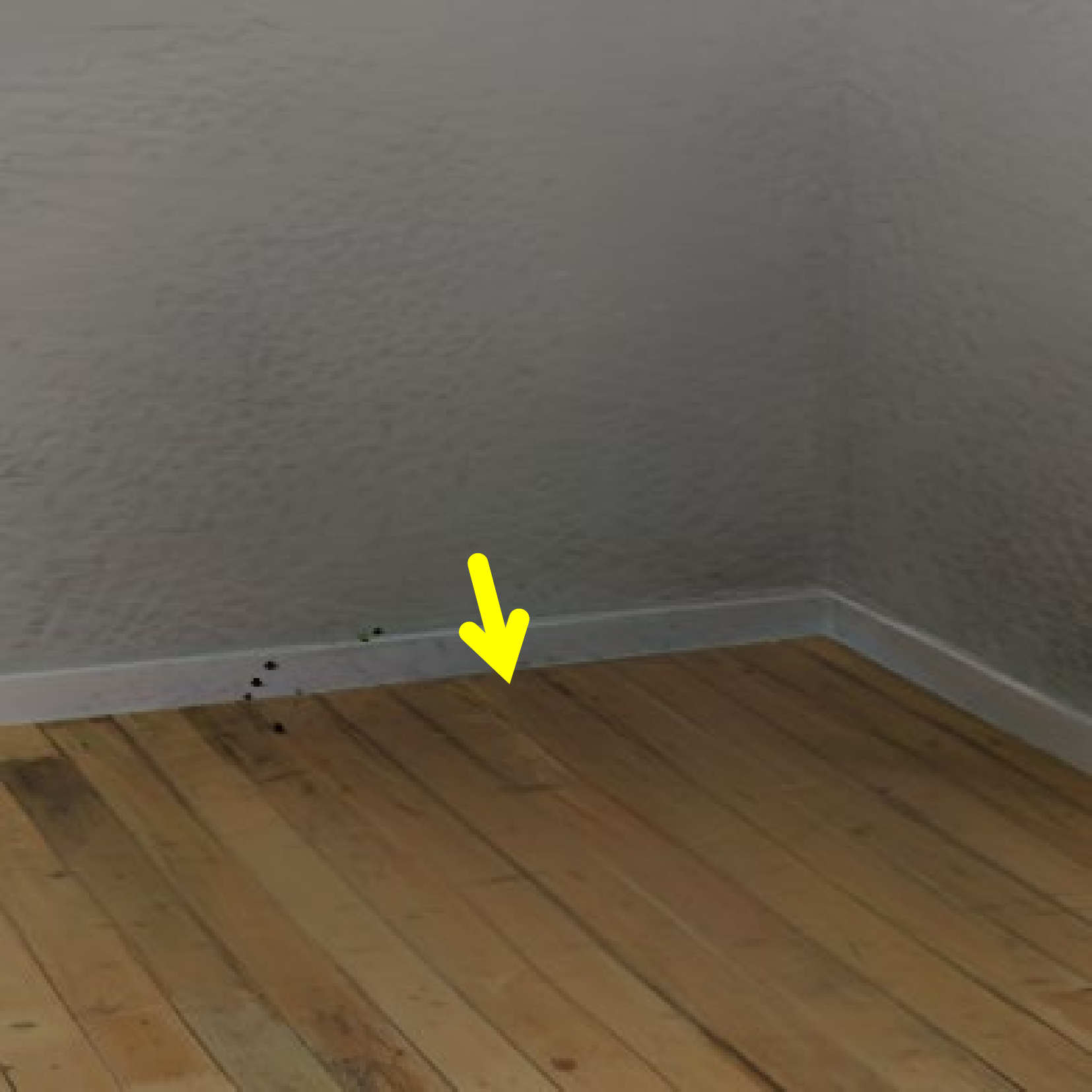} &
    \includegraphics[width=0.15\textwidth]{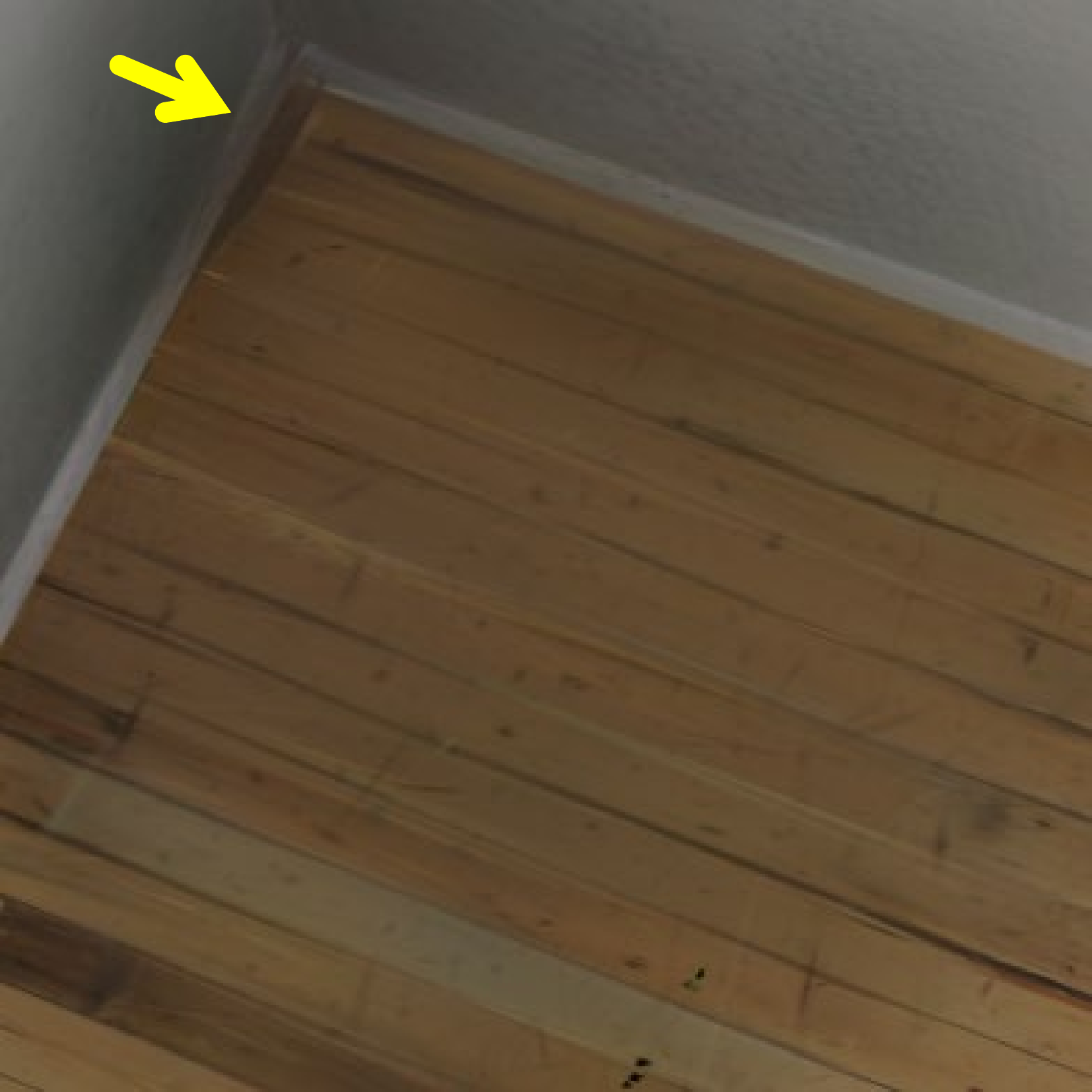} &
    \rotatebox{90}{\scriptsize Gaussian Grouping} &
    \includegraphics[width=0.15\textwidth]{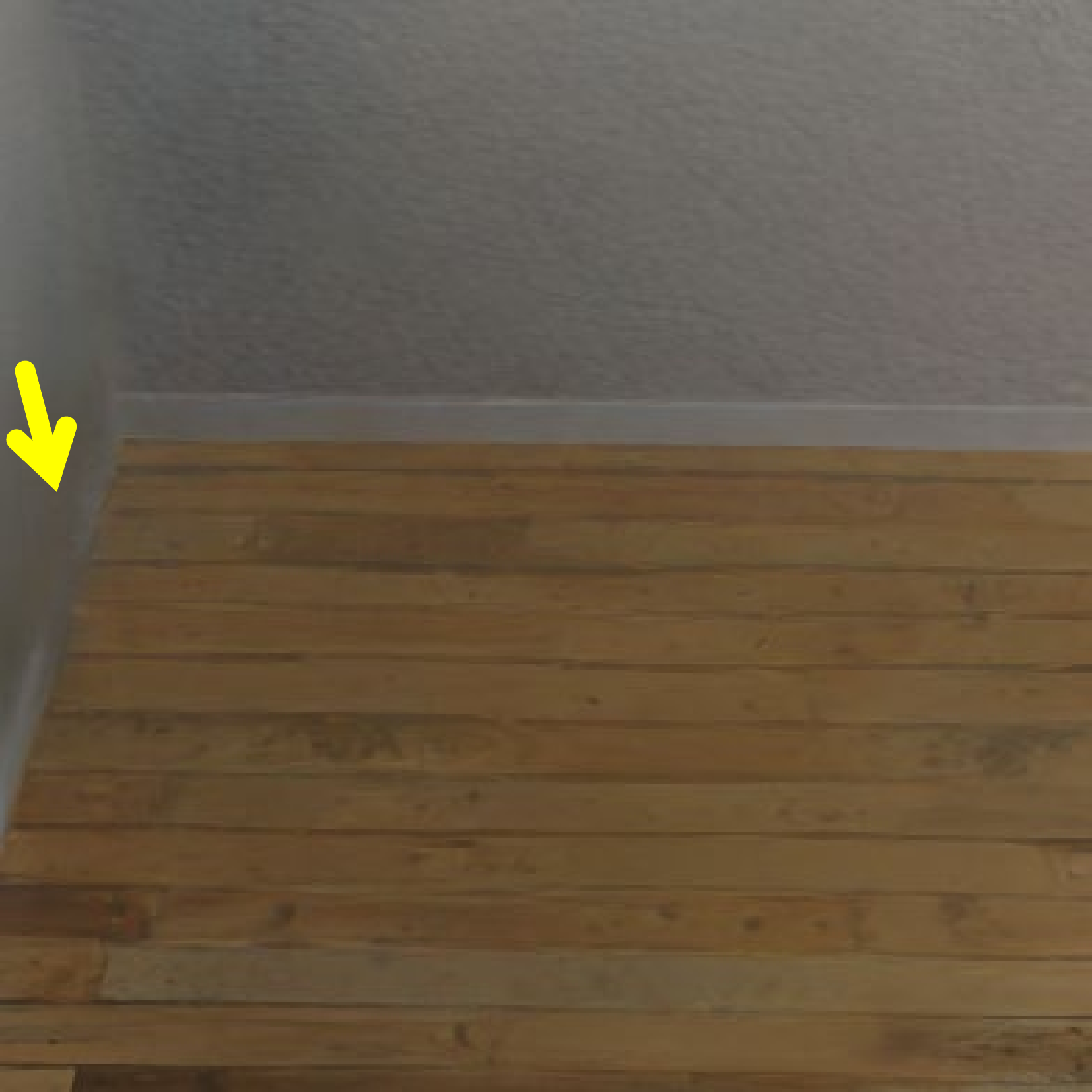} &
    \includegraphics[width=0.15\textwidth]{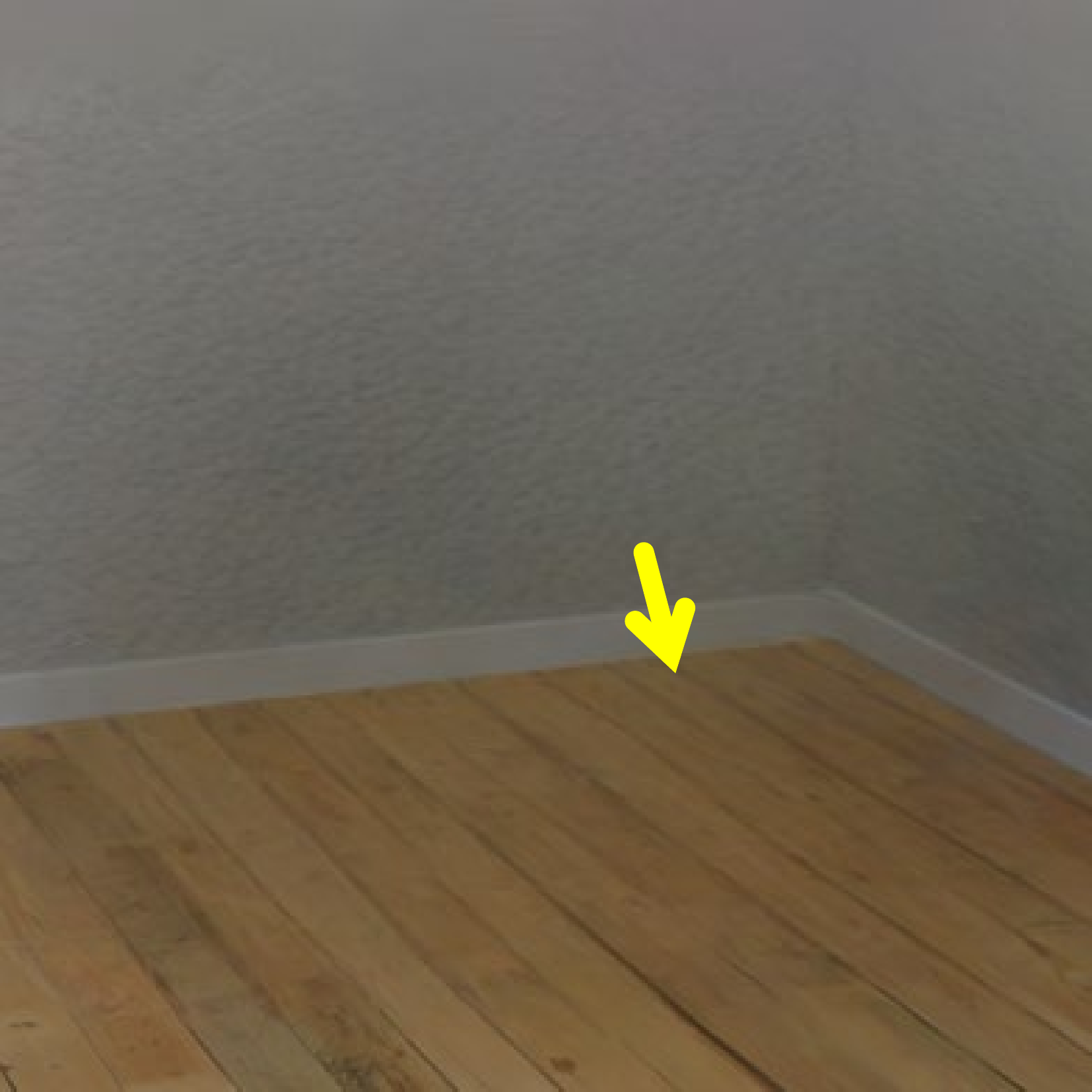} &
    \includegraphics[width=0.15\textwidth]{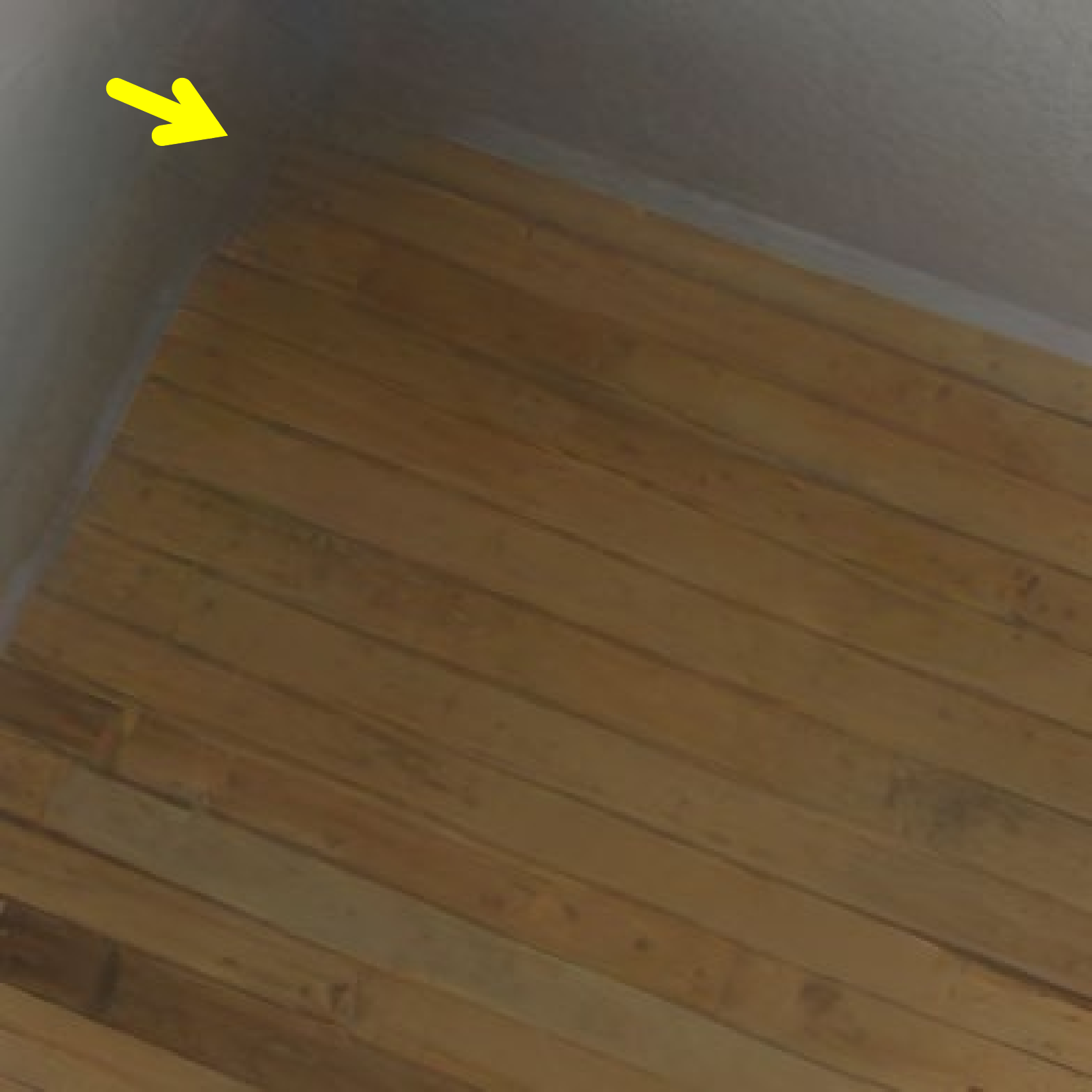} \\

    \rotatebox{90}{\scriptsize Infusion} &
    \includegraphics[width=0.15\textwidth]{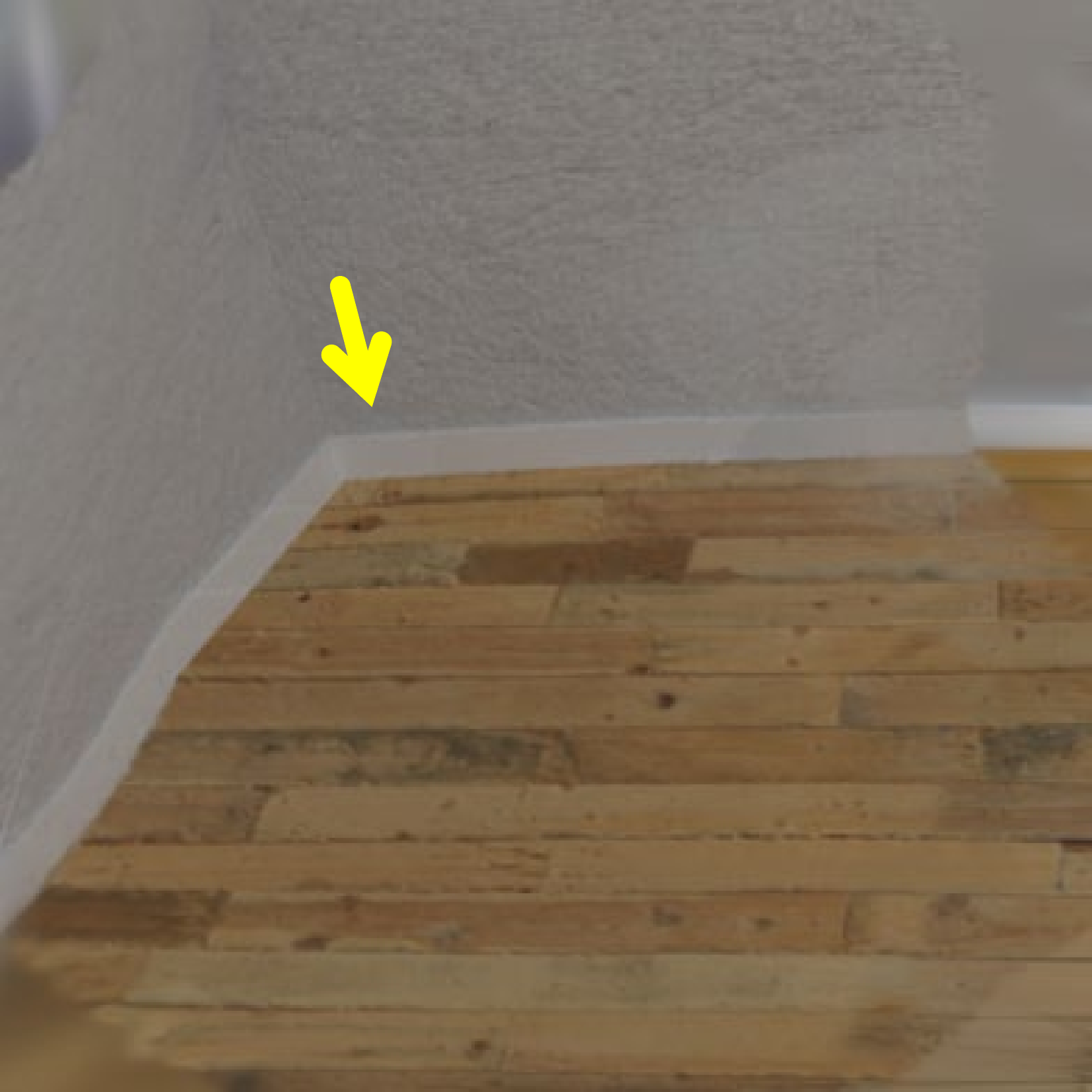} &
    \includegraphics[width=0.15\textwidth]{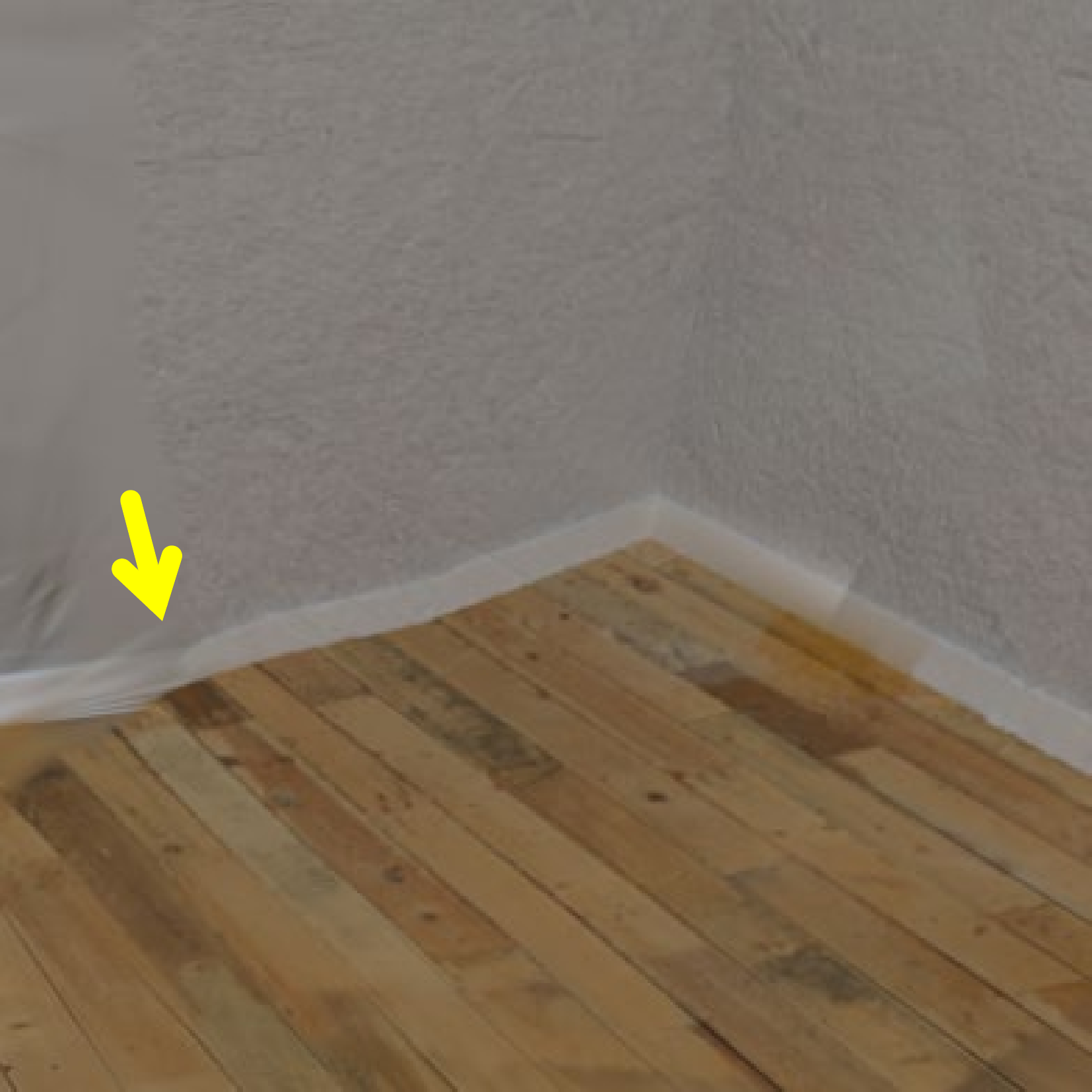} &
    \includegraphics[width=0.15\textwidth]{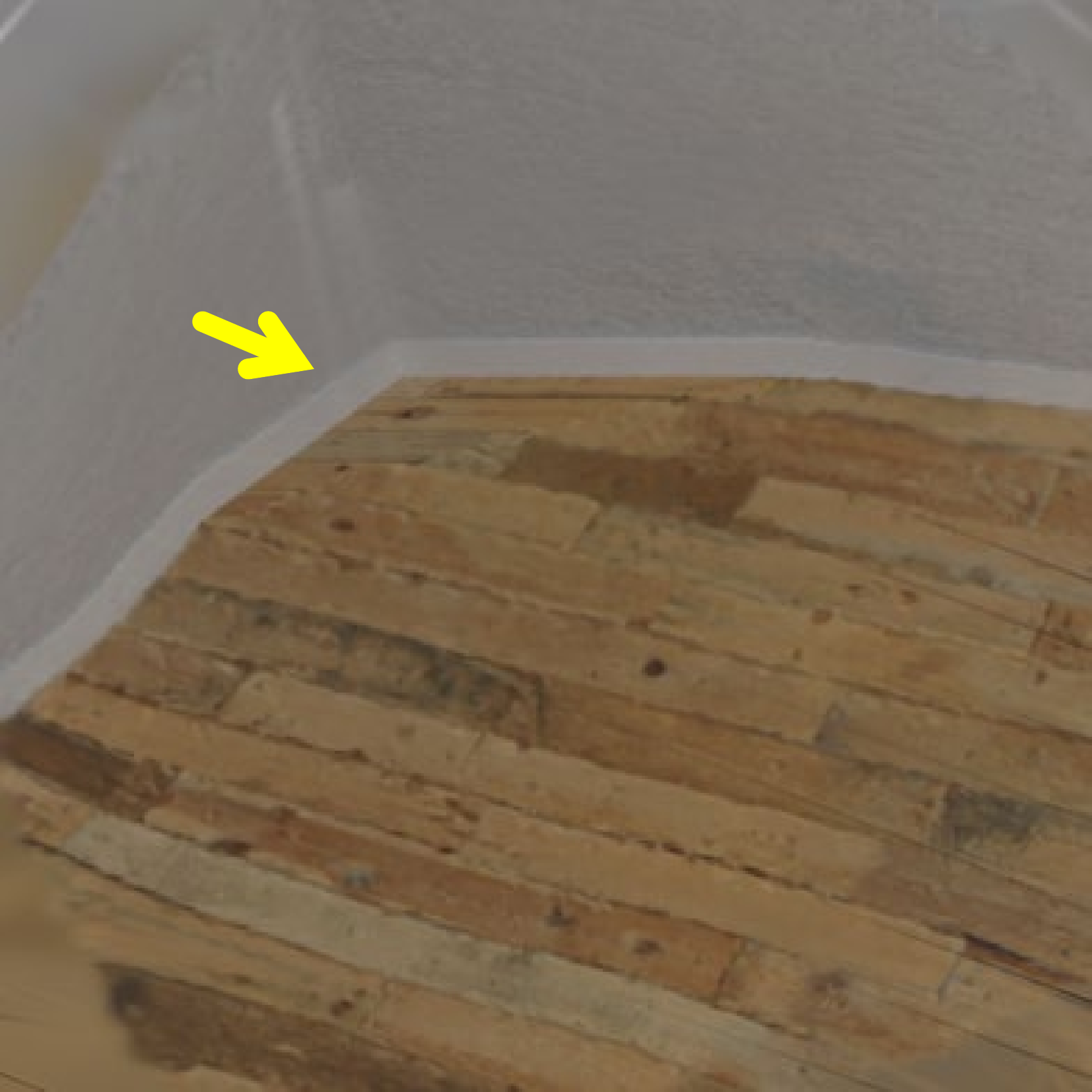} &
    \rotatebox{90}{\scriptsize Ours} &
    \includegraphics[width=0.15\textwidth]{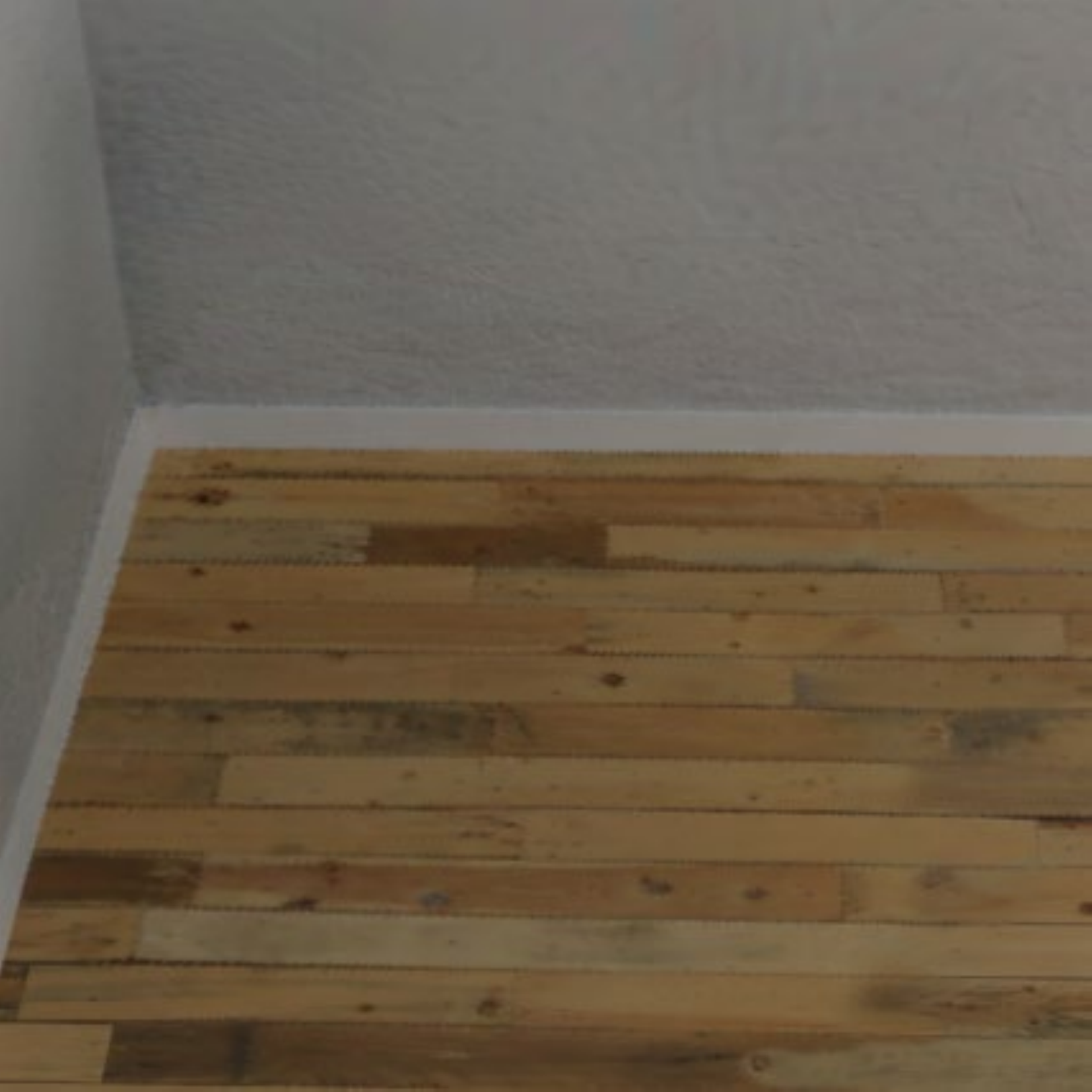} &
    \includegraphics[width=0.15\textwidth]{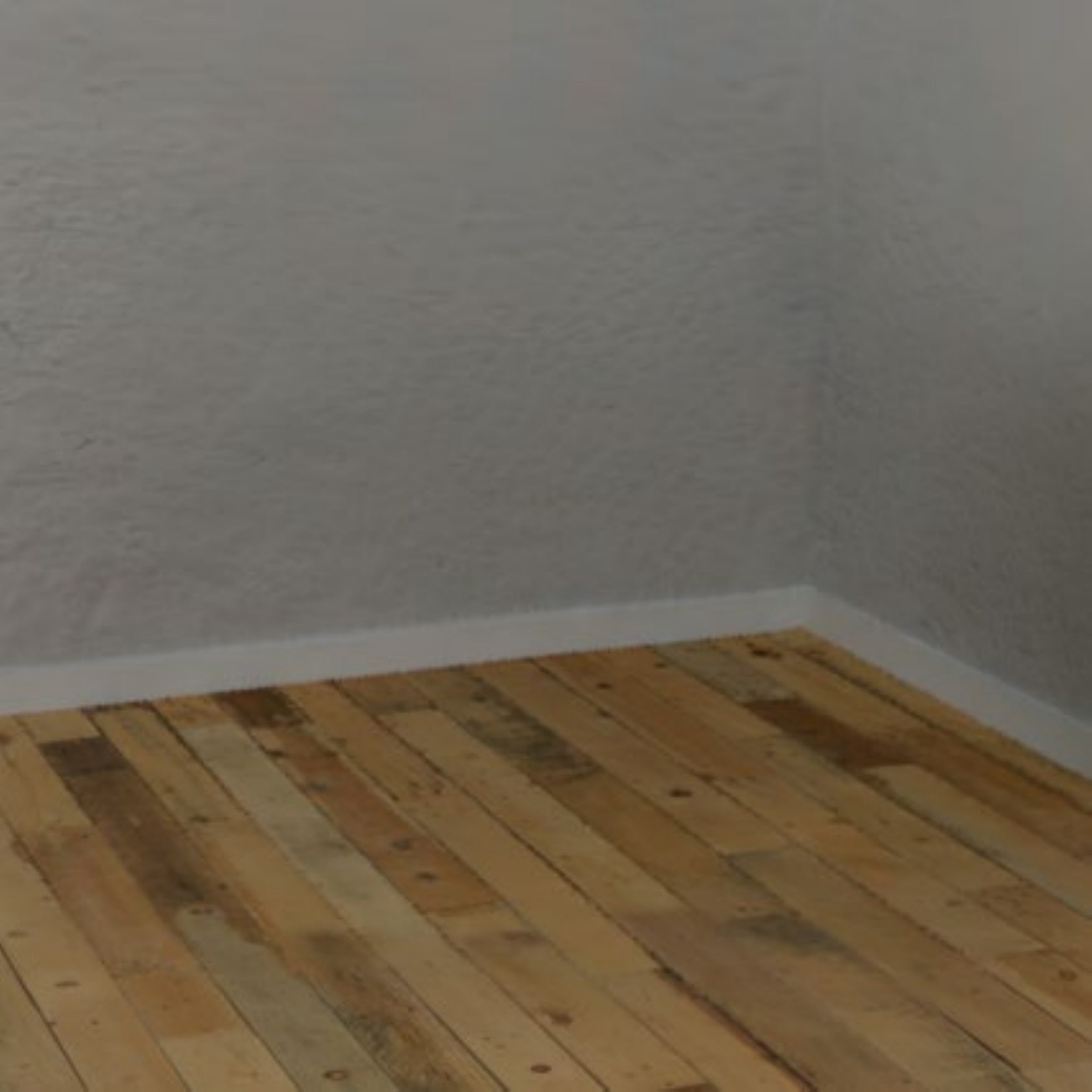} &
    \includegraphics[width=0.15\textwidth]{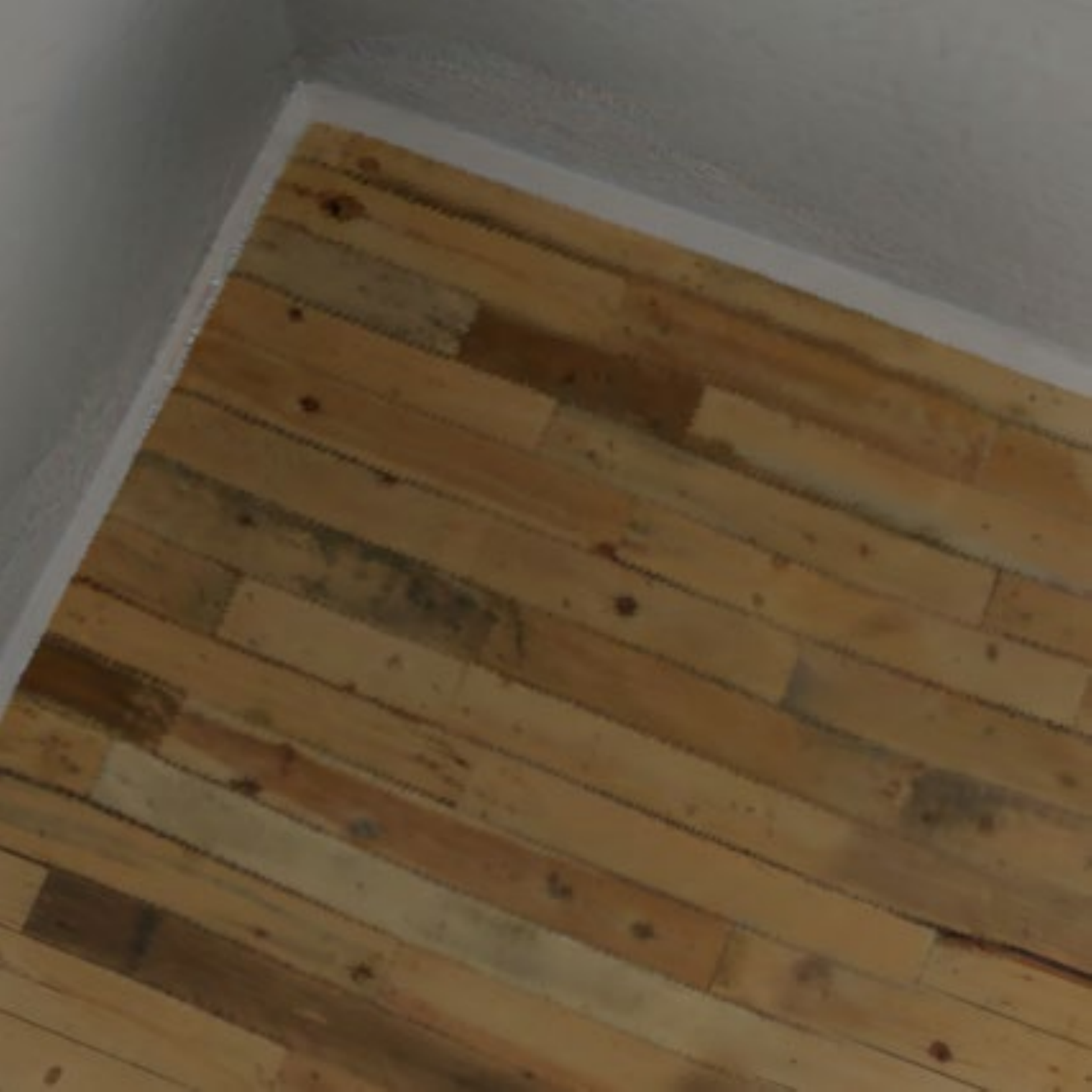} \\
\end{tabular}

\vspace{0.5mm}

\begin{tabular}{cccc @{\hspace{2mm}} cccc}
    \rotatebox{90}{\scriptsize SPIn-NeRF (reimpl.)} &
    \includegraphics[width=0.15\textwidth]{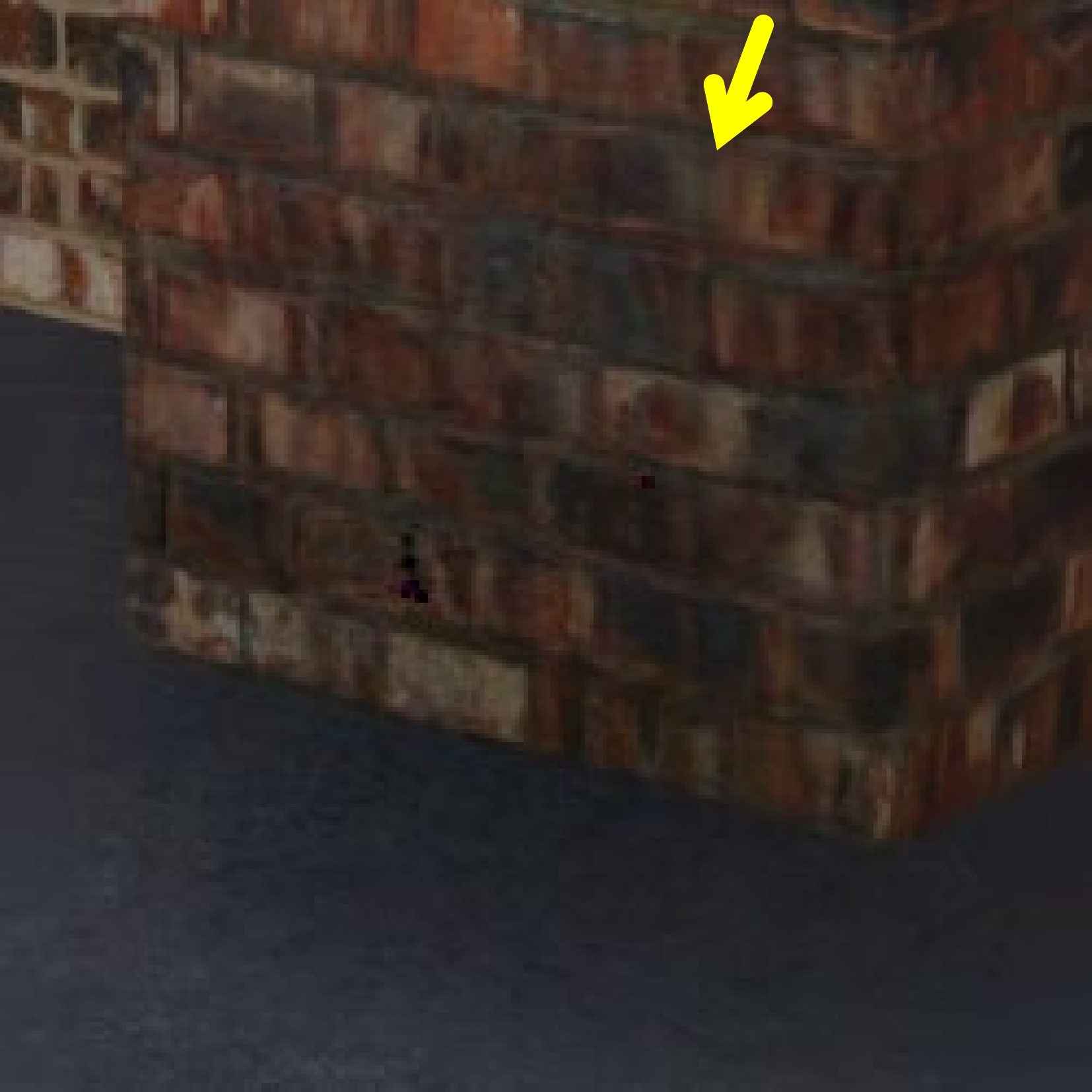} &
    \includegraphics[width=0.15\textwidth]{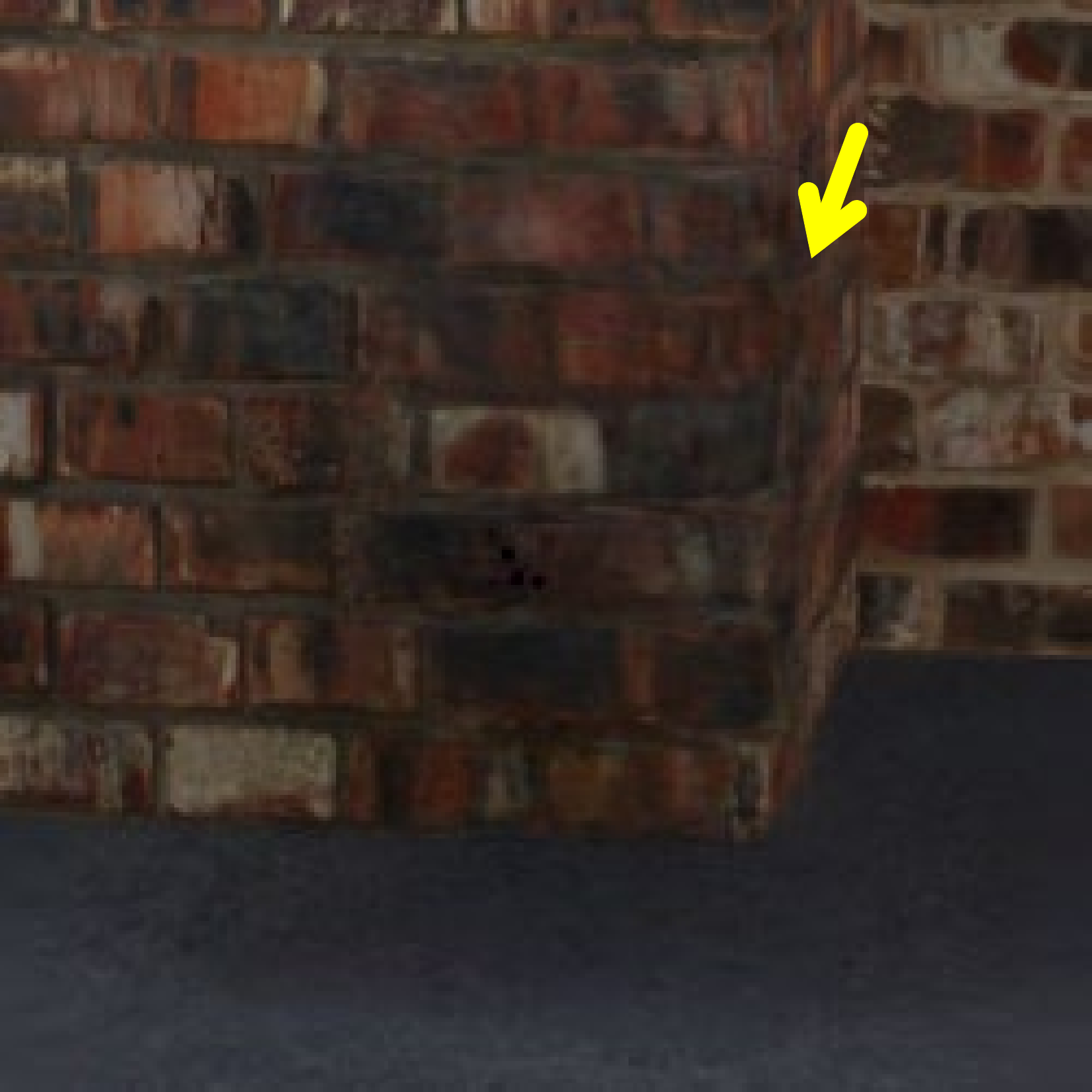} &
    \includegraphics[width=0.15\textwidth]{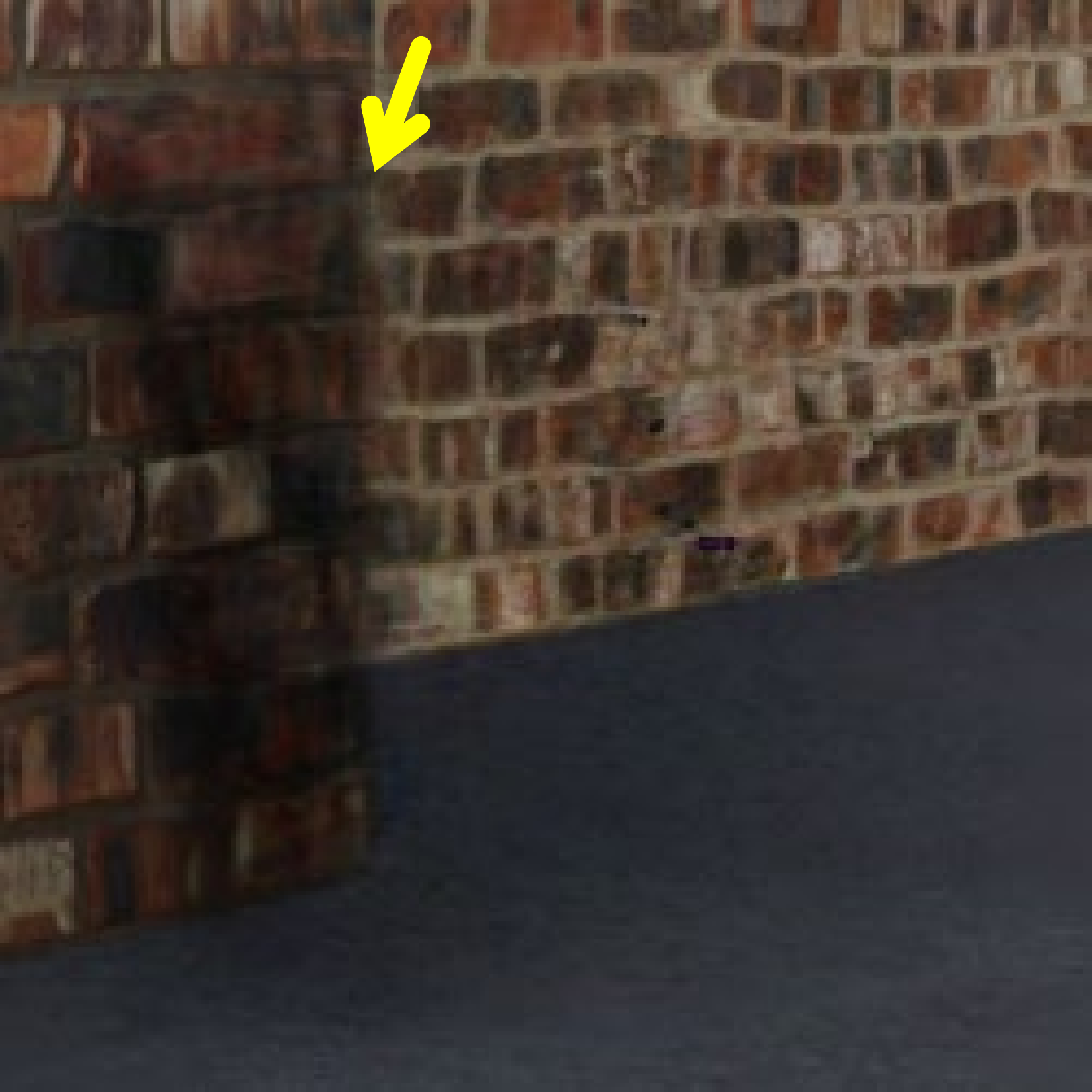} &
    \rotatebox{90}{\scriptsize Gaussian Grouping} &
    \includegraphics[width=0.15\textwidth]{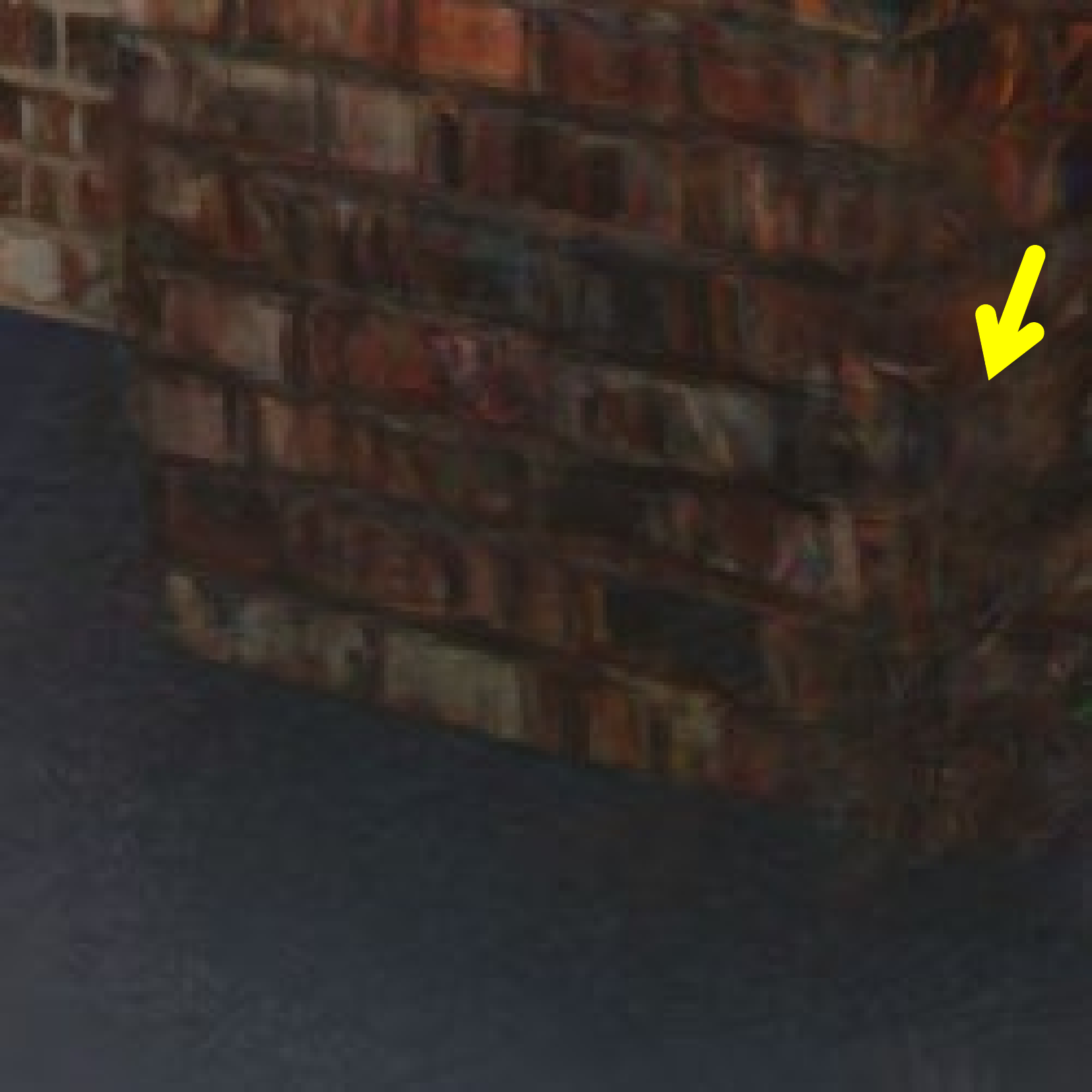} &
    \includegraphics[width=0.15\textwidth]{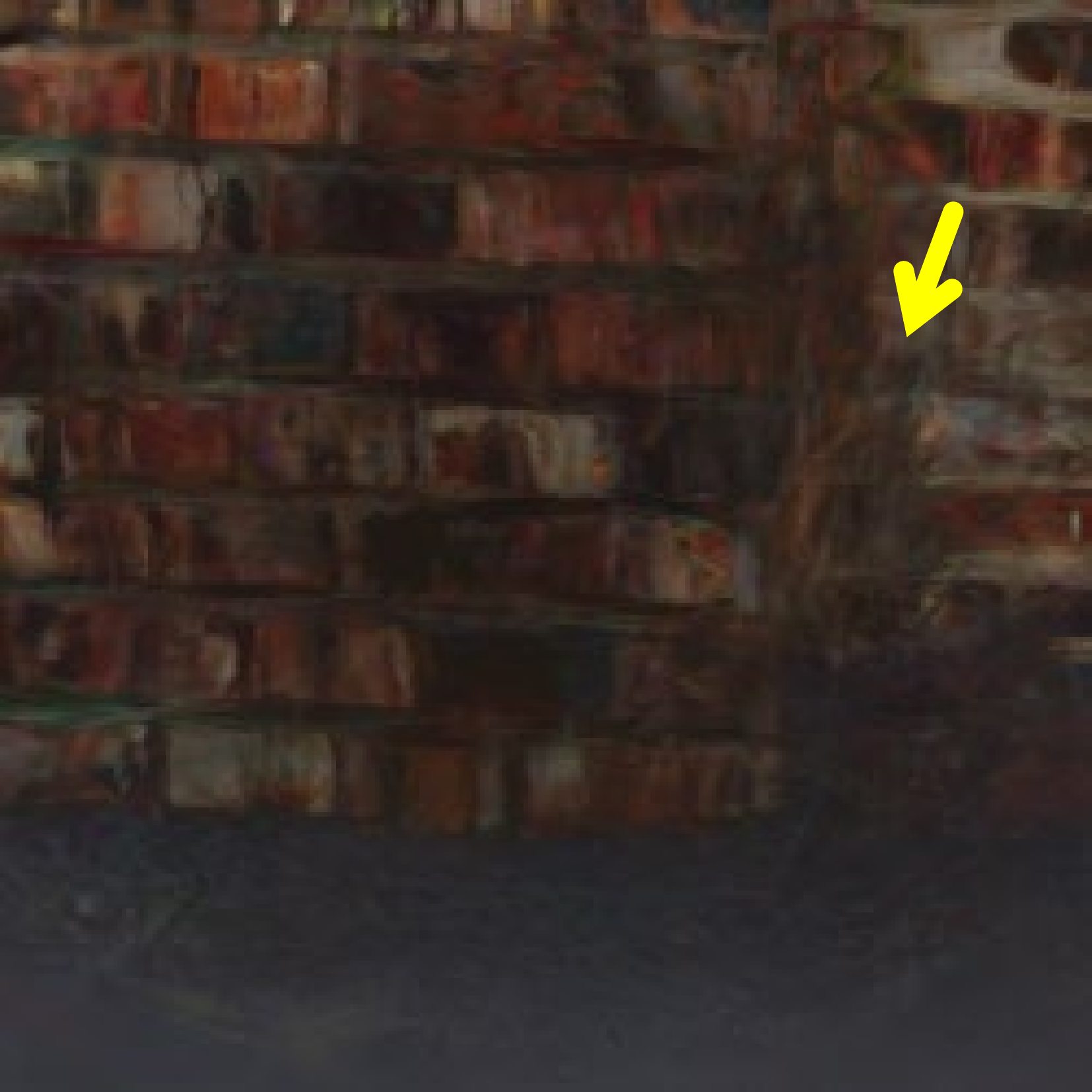} &
    \includegraphics[width=0.15\textwidth]{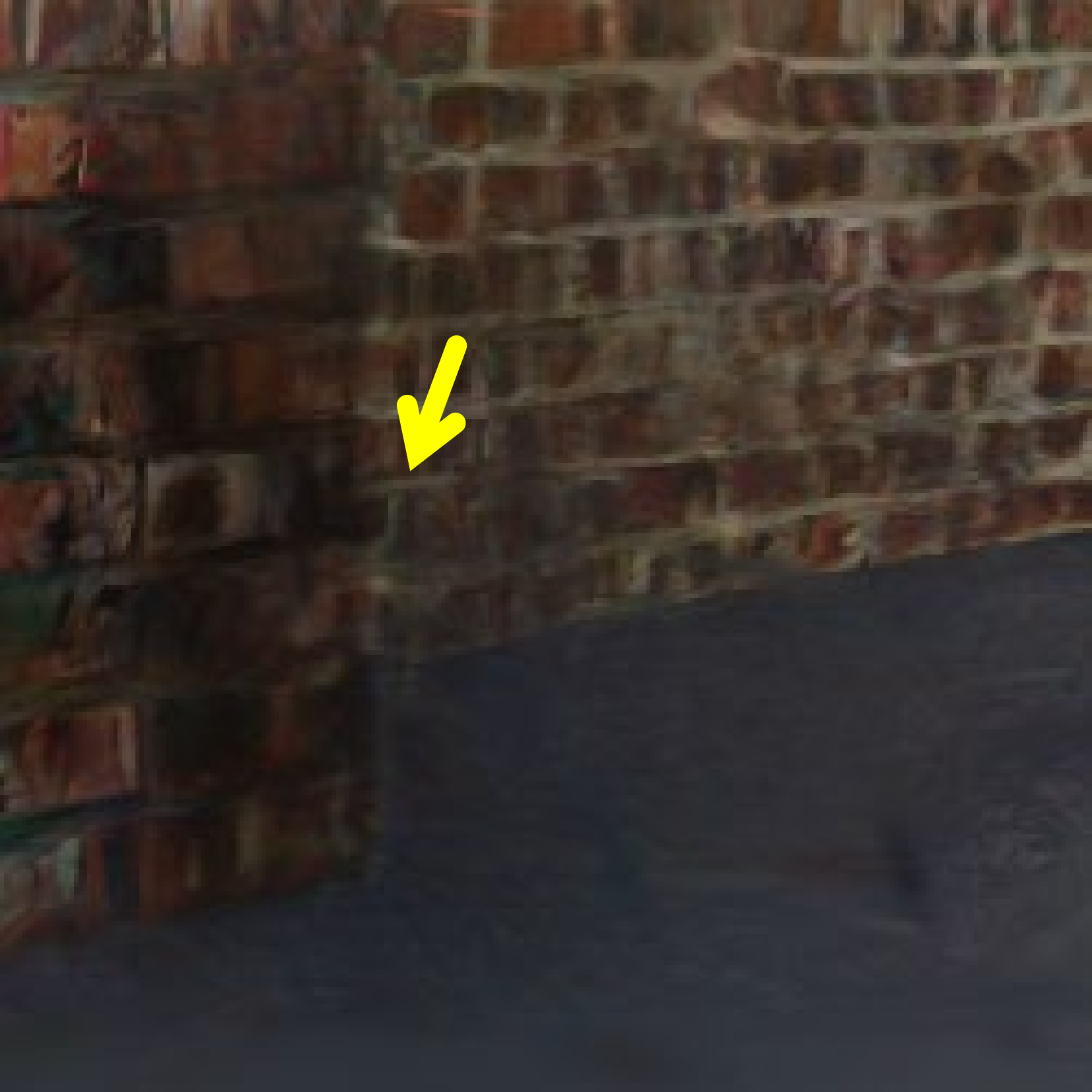} \\

    \rotatebox{90}{\scriptsize Infusion} &
    \includegraphics[width=0.15\textwidth]{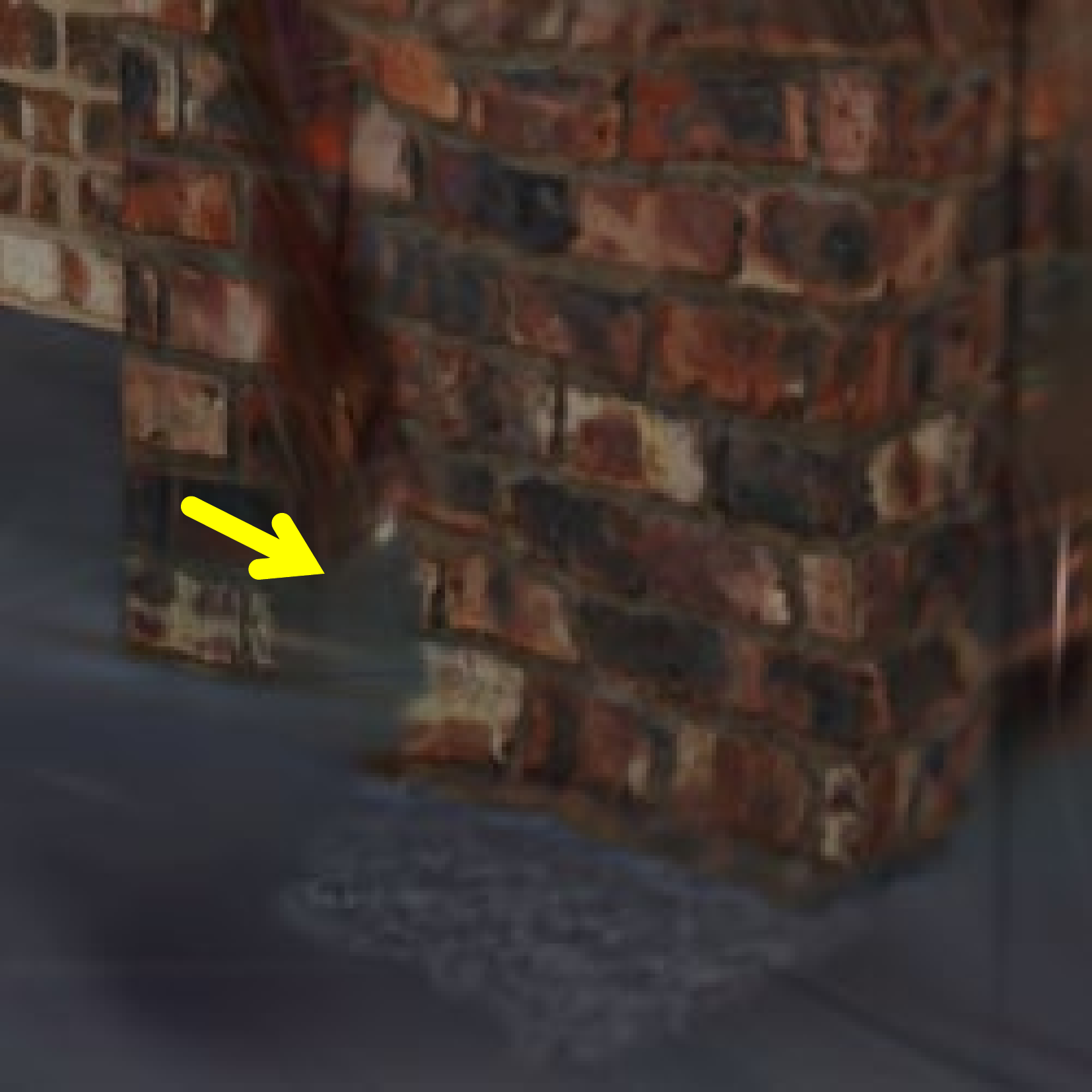} &
    \includegraphics[width=0.15\textwidth]{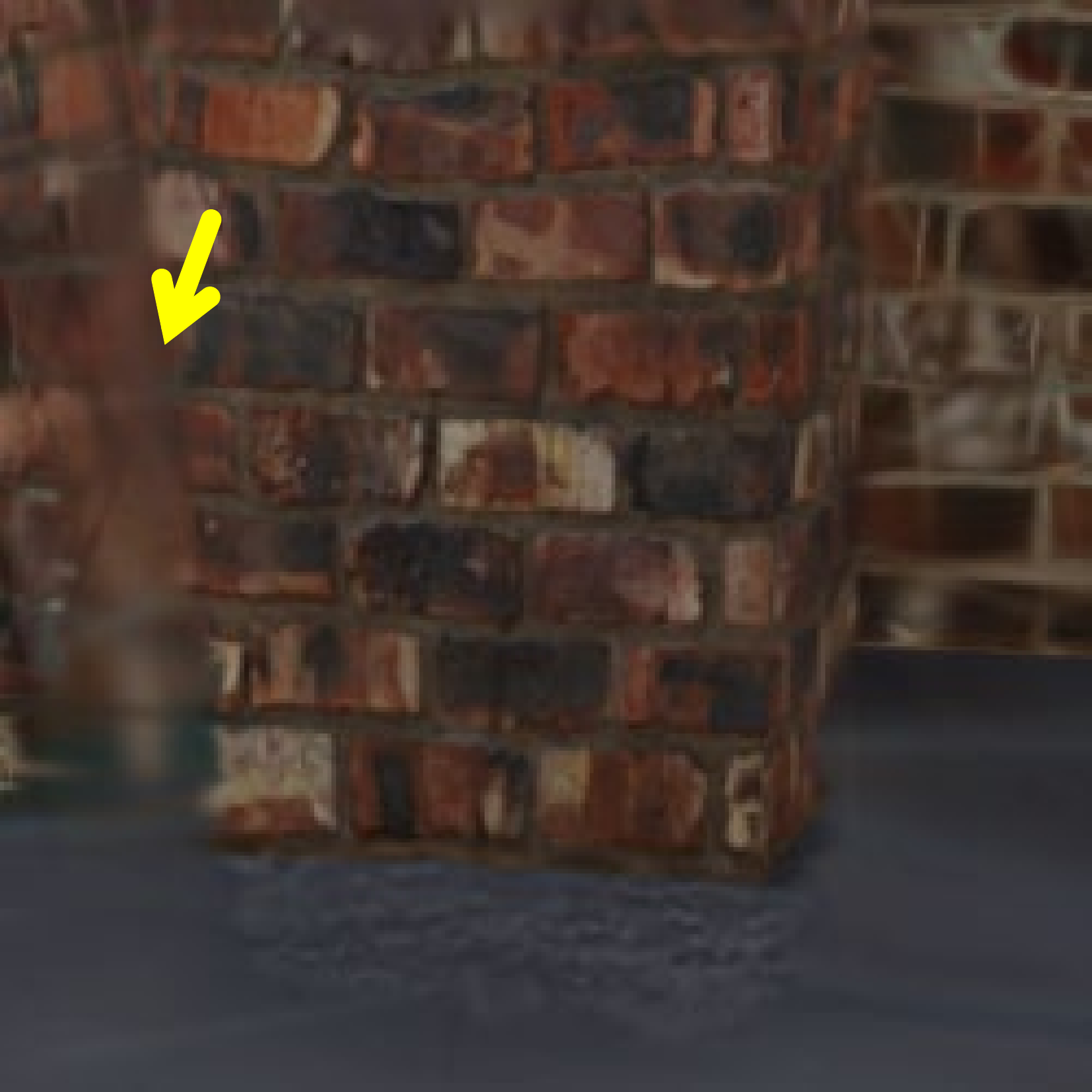} &
    \includegraphics[width=0.15\textwidth]{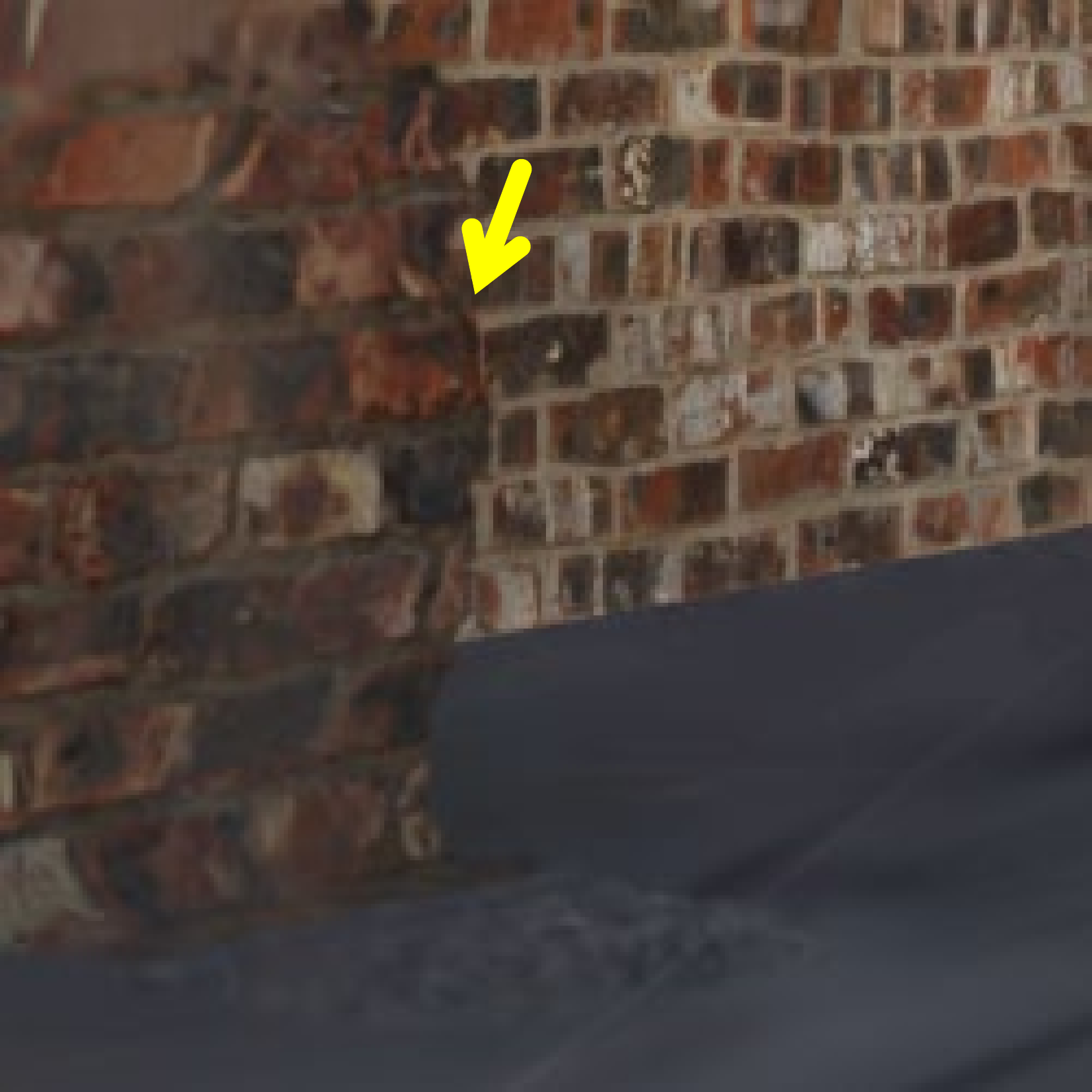} &
    \rotatebox{90}{\scriptsize Ours} &
    \includegraphics[width=0.15\textwidth]{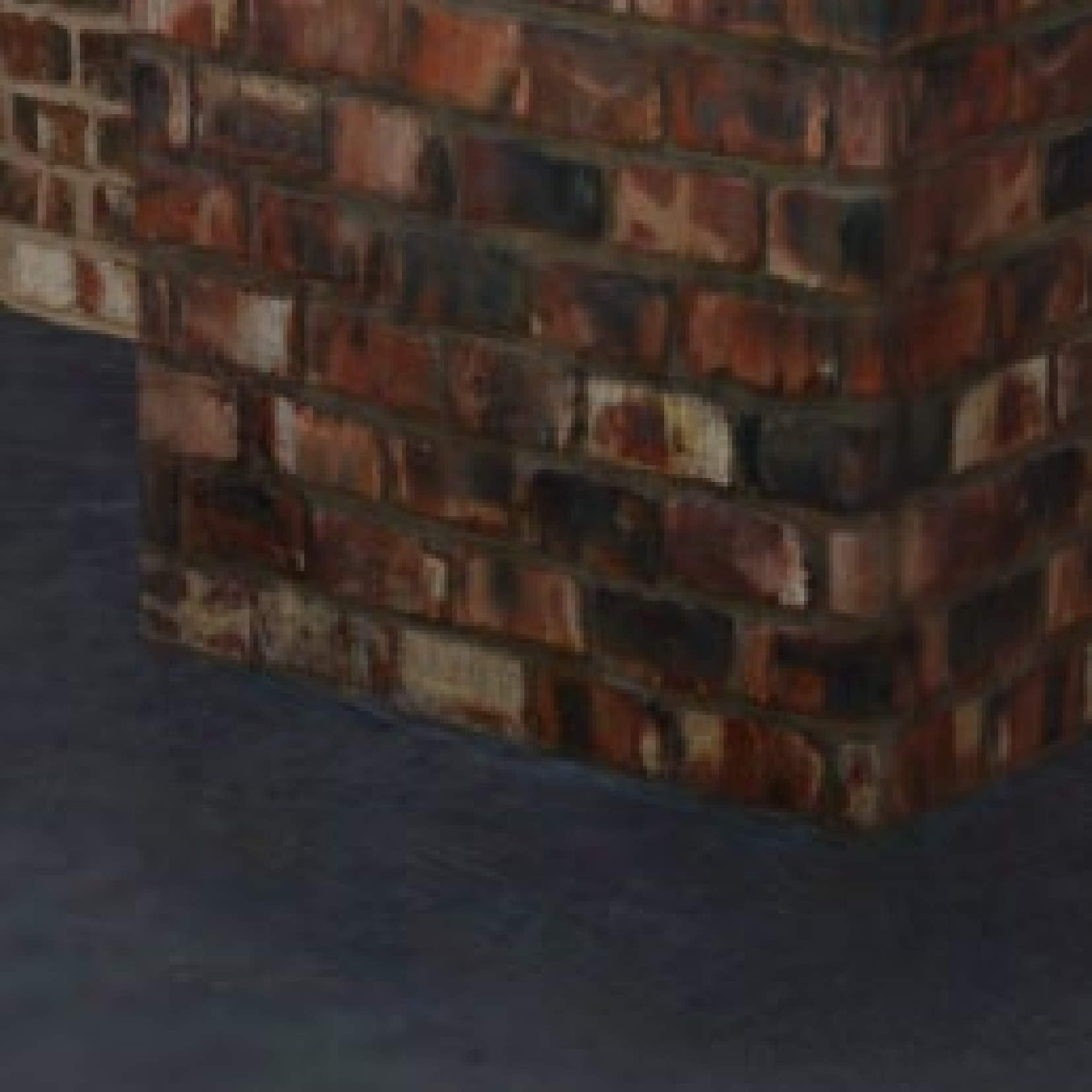} &
    \includegraphics[width=0.15\textwidth]{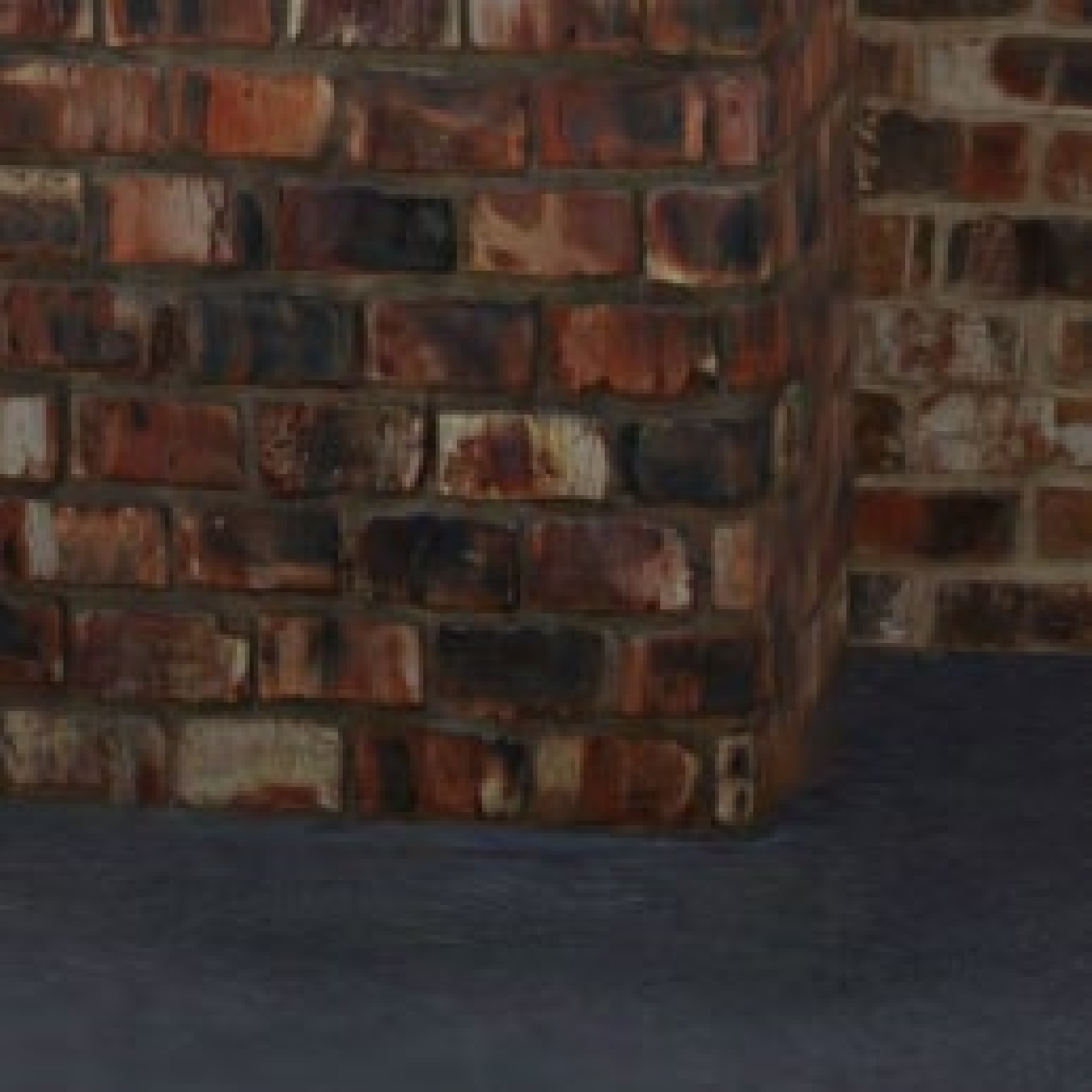} &
    \includegraphics[width=0.15\textwidth]{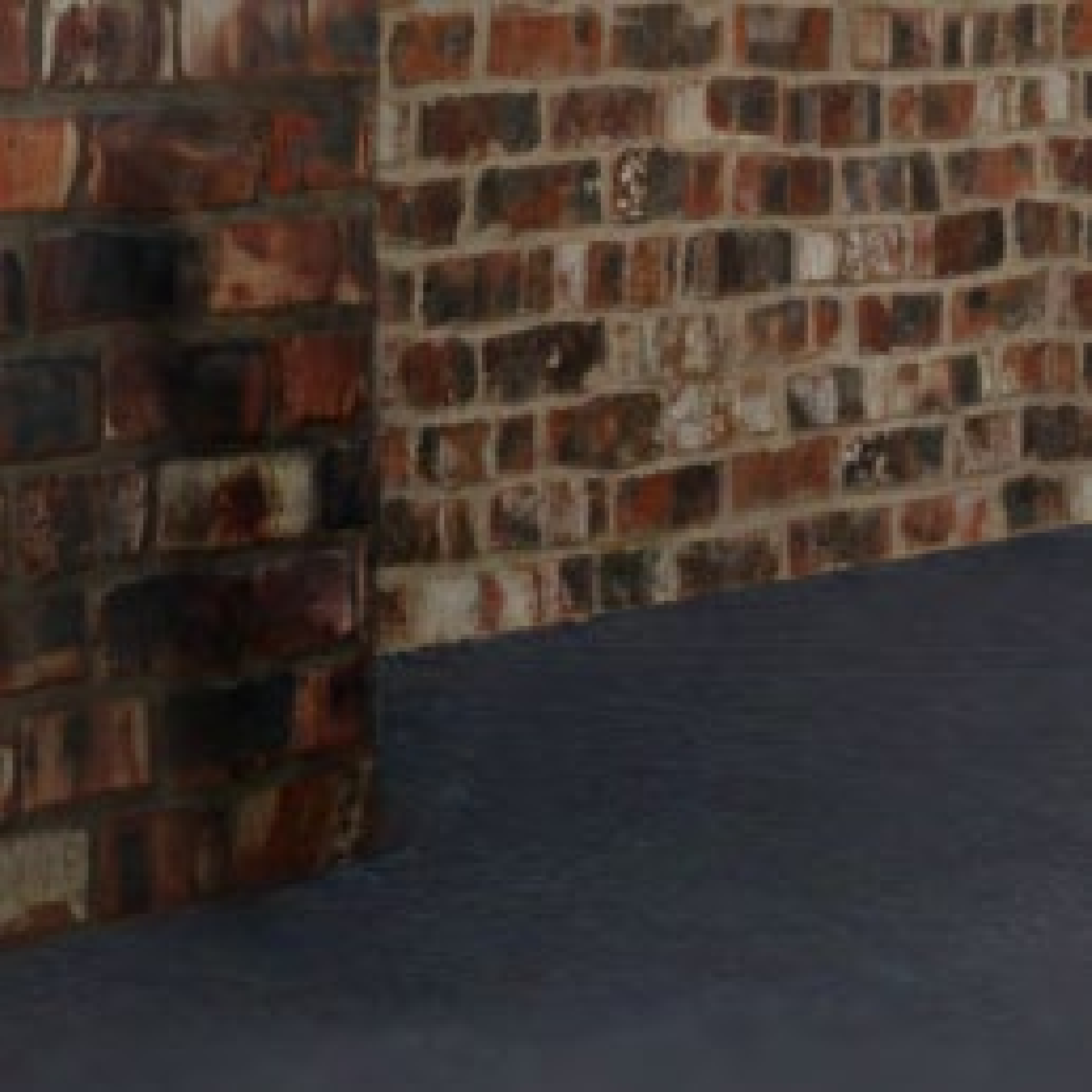} \\
\end{tabular}

\endgroup
\caption{Qualitative comparison on three novel views. This figure demonstrates the effectiveness of our method in producing coherent and photorealistic 3D scene inpainting. Arrows indicate areas where the baseline underperforms. For clearer visualization, the images are cropped to the inpainted regions, with brightness and contrast adjusted where necessary.}
\label{fig:eval2_comparison}
\end{figure*}

We conduct experiments on SPIn-NeRF, MipNeRF360, and our custom dataset, comprising a total of 10 scenes. However, it should be noted that, in an effort to ensure equal conditions, the inpainted images generated after our initial 3DGS were provided to all baseline methods. In this experiment, SPIn-NeRF can also be regarded as undergoing a 3D segmentation process. 
Fig.~\ref{fig_4_comp} shows the qualitative comparisons with baseline methods. The images are rendered from one of the test views. It can be seen that our method plausibly and finely reconstructs the background appearance, producing inpainted regions that blend naturally with the original scene. On the other hand, while SPIn-NeRF generally captures the overall semantic content well, it fails to produce fine details and performs poorly on challenging scenes with severe inpainting inconsistencies, such as the bench scene (Fig.~\ref{fig_4_comp}(a)). 
In the case of Gaussian Grouping, floating artifacts appear depending on the level of inconsistency in the input, as it does not utilize depth inpainting. Lastly, for Infusion, the view selected for depth completion was set to be the same as the reference view used in our method. Although Infusion produced visually plausible results for views close to the selected, errors in depth estimation became more apparent as the viewpoint moved further away. Such behavior is more clearly illustrated in Fig.~\ref{fig:eval2_comparison}, which compares results on three different novel views across several wide-range view datasets. Unlike baselines that lose visual details or cause geometric distortions, our method preserve realistic and consistent content across multiple viewpoints.

\begin{figure*}[htbp]
    \centering
\includegraphics[width=\textwidth]{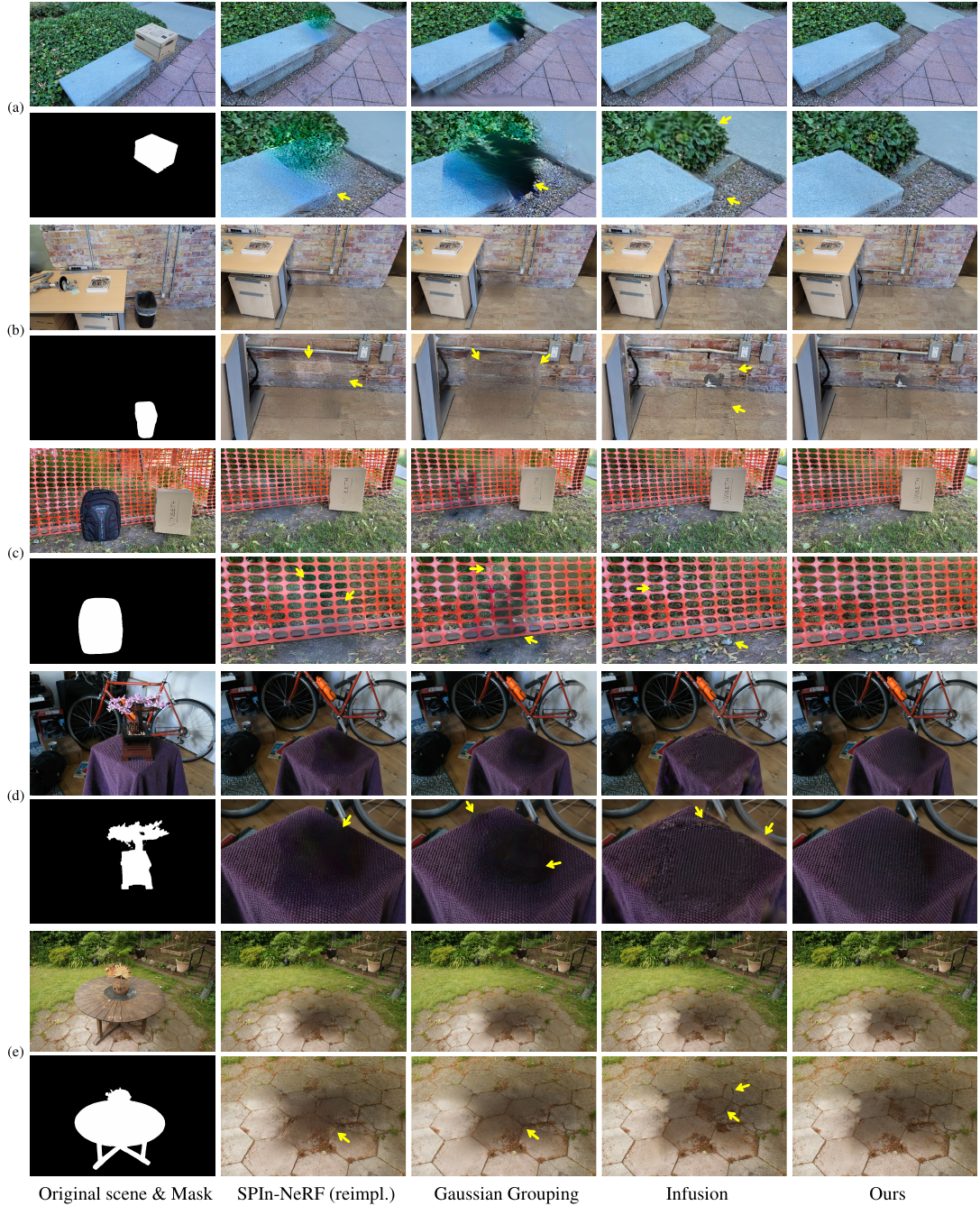}
\caption{Visual comparison of test set renderings. This figure presents sample renderings from the novel test view: (a)–(c) are from SPIn-NeRF, (d)–(e) from MipNeRF360, and (f)–(h) from our custom dataset. Each scene is shown in two rows—the top row presents the full image, while the bottom row shows a cropped region focusing on the inpainted area. Arrows indicate areas where the baseline underperforms. Our method not only preserves fine details from the inpainting input but also reconstructs perceptually consistent scenes across views, guided by a single reference image. For clearer comparison, some images have been adjusted in brightness and contrast.}
\label{fig_4_comp}
\end{figure*}

\begin{figure*}[htb]
    \ContinuedFloat
    \centering
\includegraphics[width=\textwidth]{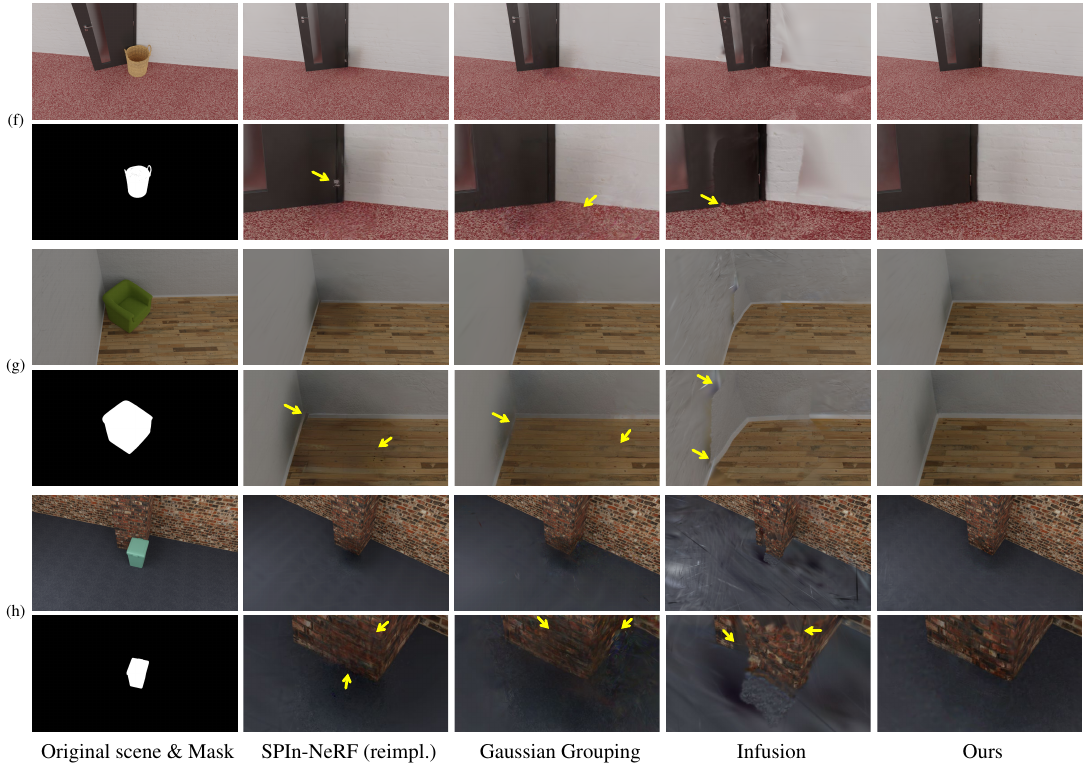}
\caption{Visual comparison of test set renderings. This figure presents sample renderings from the novel test view: (a)–(c) are from SPIn-NeRF, (d)–(e) from MipNeRF360, and (f)–(h) from our custom dataset. Each scene is shown in two rows—the top row presents the full image, while the bottom row shows a cropped region focusing on the inpainted area. Arrows indicate areas where the baseline underperforms. Our method not only preserves fine details from the inpainting input but also reconstructs perceptually consistent scenes across views, guided by a single reference image. For clearer comparison, some images have been adjusted in brightness and contrast.}
\end{figure*}

\begin{table}[h!t]
\centering
\caption{Qualitative results on the SPIn-NeRF dataset. Average values were calculated from five scenes. Measurements were computed within the bounding box of the inpainted region.}
\begin{tabular}{lccc}
\toprule
Method                      & SSIM $\uparrow$ & LPIPS $\downarrow$ & FID $\downarrow$ \\
\midrule
SPIn-NeRF (reimpl.) &0.3696    & 0.3521     & 174.62                        \\
Gaussian Grouping           &0.2795    & 0.4409     & 218.84                        \\
Infusion                    &0.4313    & 0.3902     & 154.02                        \\
Ours                        &\textbf{0.4740}    & \textbf{0.2431}     & \textbf{70.73}    \\
\bottomrule
\end{tabular}
\label{tab:result_spindata}
\end{table}

\begin{table}[h!t]
\centering
\caption{Qualitative results on our custom dataset. Average values were calculated from three scenes. Measurements were computed within the bounding box of the inpainted region.}
\begin{tabular}{lccc}
\toprule
Method                      & SSIM $\uparrow$ & LPIPS $\downarrow$ & FID $\downarrow$ \\
\midrule
SPIn-NeRF (reimpl.) & 0.7659   & 0.2776 & 153.23  \\
Gaussian Grouping           & 0.7352   & 0.2994 & 165.31  \\
Infusion                    & 0.7601   & 0.3544 & 194.50  \\
Ours                        & \textbf{0.7913} & \textbf{0.2267} & \textbf{104.00}  \\
\bottomrule
\end{tabular}
\label{tab:result_bprocdata}
\end{table}

We also evaluate the quantitative performance of our method. To assess the quaility of 3D scene inpainting, the evaluation was conducted within the bounding box of the inpainting region.
Table~\ref{tab:result_spindata} and Table~\ref{tab:result_bprocdata} present quantitative comparisons against baseline methods. These results show that our proposed method performs best in terms of SSIM, LPIPS, and FID.

\begin{figure*}[!ht] 
    \centering
\includegraphics[width=\textwidth]{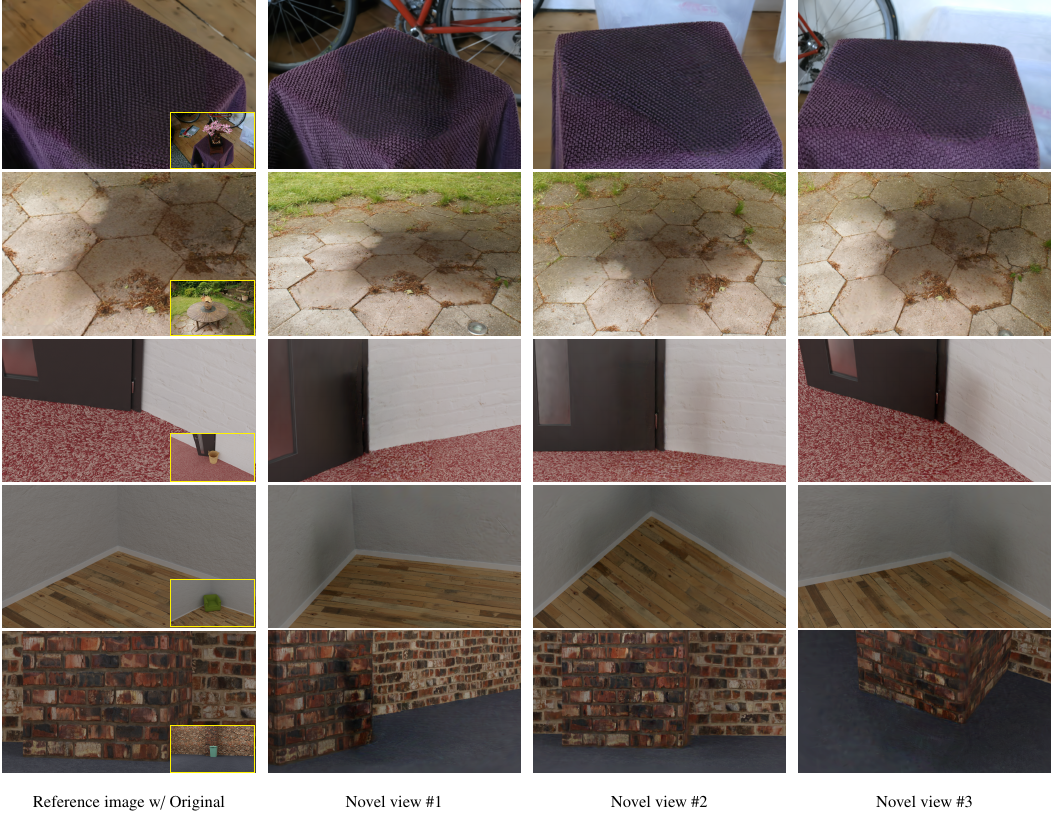}

\caption{High-fidelity appearance from a single reference view. 
We present sample novel views from wide-view datasets (MipNeRF360~\cite{Barron2022MipNeRF360} and our custom scenes), demonstrating that our method produces results that remain perceptually faithful to the reference across a wide range of viewpoints. The images show cropped regions of the inpainted areas, with brightness adjusted for better inspection.}
\label{fig_4_consistent}
\end{figure*}

Moreover, Fig.~\ref{fig_4_consistent} shows that our method produces results that remain perceptually faithful to the single reference across a wide range of viewpoints. Although the overall appearance is slightly darker due to shadows left by the removed object, a consistent appearance with the reference image is maintained. 

\subsection{Ablation studies}
To assess the contribution of each component in our proposed methods, we conducted ablation experiments focusing on two key aspects. One compares the performance of inpainting's confidence-based weighting, and the other evaluates the effect of different loss term combinations used during optimization.

Table~\ref{tab:result_abl1} shows the results of the ablation study on confidence-based weighting. In the table, Uniform-Weight refers to training inpaint-3DGS with equal weights to each view, regardless of inpainting confidence, and Confidence-Threshold denotes the setting in which inpainted regions with confidence values below a certain threshold are skipped during training. It shows that adjusting each view's weight based on their inpainting confidence shows moderate improvement.

\begin{table}[h!]
\caption{An ablation study on the effect of confidence-guided view weighting was conducted on our custom dataset. \textit{Uniform-Weight} uses equal weights for all views, \textit{Confidence-Threshold} excludes low-confidence views based on a threshold, and \textit{Confidence-Weighted} adaptively weights views by confidence.}
\centering
\begin{tabular}{lccc}
\toprule
Method                          &SSIM$\uparrow$ & LPIPS$\downarrow$ & FID$\downarrow$ \\ 
\midrule
Uniform-Weight                  & 0.7763 & 0.2312      & 107.93    \\ 
Confidence-Threshold            & 0.7889 & 0.2278      & 104.28    \\ 
Confidence-Weighted (ours)      & \textbf{0.7913} & \textbf{0.2267} & \textbf{104.00}    \\ 
\bottomrule
\end{tabular}
\label{tab:result_abl1}
\end{table}

While its impact is limited when the inpainted images are largely geometrically consistent, it offers significant improvements in the presence of severe inconsistencies.

Following, Table~\ref{tab:result_abl2} reports the impact of varying loss term combinations in our ablation study.
We ablate the losses related to the inpainted regions from the full loss formulation (Eq. (\ref{eq:total_loss})). Specifically, we sequentially exclude the normal, perceptual, and color loss on the warped reference image. The depth loss, which is commonly used in prior work, is retained as its effectiveness has already been well established. Here, \textit{w/o $\mathcal{L}_{iC}$} denotes the removal of the inpainting color loss component from $\mathcal{L}_{C}$, specifically by excluding the warped reference image from the calculation. Although $\mathcal{L}_{C}$ is not strictly decomposable, we use it for clarity in the ablation setting.

\begin{table}[h!]
\centering
\caption{Ablation study of our proposed loss formulation on the custom dataset.
We evaluate the contribution of each loss term by removing them individually. \textit{w/o $\mathcal{L}_{iC}$} refers to excluding the warped reference image from $\mathcal{L}_C$ during training.}
\label{tab:result_abl2}
\begin{tabular}{lccc}
\toprule
Method                &SSIM$\uparrow$ & LPIPS$\downarrow$ & FID$\downarrow$ \\ 
\midrule
w/o $L_{iN}$      & 0.7115  & 0.3466 & 207.33    \\ 
w/o $L_{iLPIPS}$  & \underline{0.8015}  & 0.3237 & 160.77    \\ 
w/o $L_{iC}$      & \textbf{0.8208} & \textbf{0.2209} & \underline{113.95}   \\ 
Ours       & 0.7913 & \underline{0.2267} & \textbf{104.00}    \\ 
\bottomrule
\end{tabular}

\end{table}

\begin{figure*}[!t]
    \centering
    \begin{minipage}[t]{0.19\textwidth}
        \centering
        \includegraphics[width=0.97\textwidth]{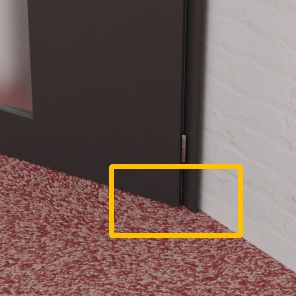}
    \end{minipage}
    \begin{minipage}[t]{0.19\textwidth}
        \centering
        \includegraphics[width=0.97\textwidth]{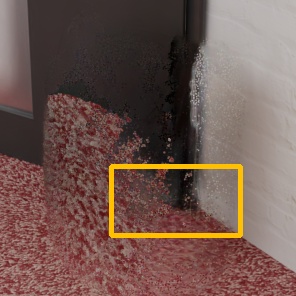}
    \end{minipage}
    \begin{minipage}[t]{0.19\textwidth}
        \centering
        \includegraphics[width=0.97\textwidth]{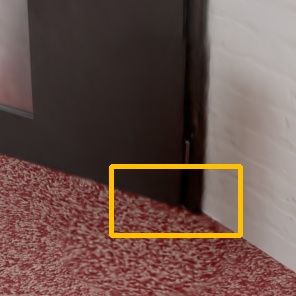}
    \end{minipage}
        \begin{minipage}[t]{0.19\textwidth}
        \centering
        \includegraphics[width=0.97\textwidth]{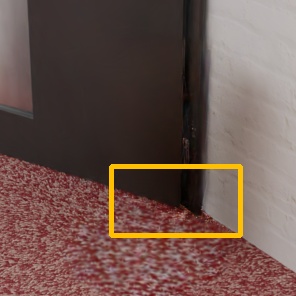}
    \end{minipage}
    \begin{minipage}[t]{0.19\textwidth}
        \centering
        \includegraphics[width=0.97\textwidth]{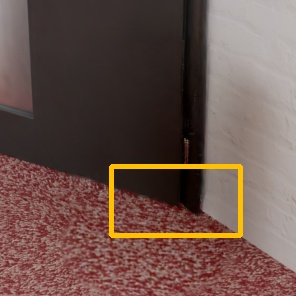}
    \end{minipage}
    \par\vspace{0.3mm}
    \begin{minipage}[t]{0.19\textwidth}
        \centering
        \includegraphics[width=0.97\textwidth]{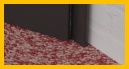}
    \end{minipage}
    \begin{minipage}[t]{0.19\textwidth}
        \centering
        \includegraphics[width=0.97\textwidth]{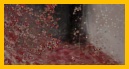}
    \end{minipage}
    \begin{minipage}[t]{0.19\textwidth}
        \centering
        \includegraphics[width=0.97\textwidth]{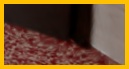}
    \end{minipage}
    \begin{minipage}[t]{0.19\textwidth}
        \centering
        \includegraphics[width=0.97\textwidth]{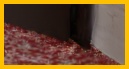}
    \end{minipage}
    \begin{minipage}[t]{0.19\textwidth}
        \centering
        \includegraphics[width=0.97\textwidth]{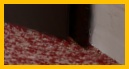}
    \end{minipage}

    \vspace{1.0mm}

    \begin{minipage}[t]{0.19\textwidth}
        \centering
        \includegraphics[width=0.97\textwidth]{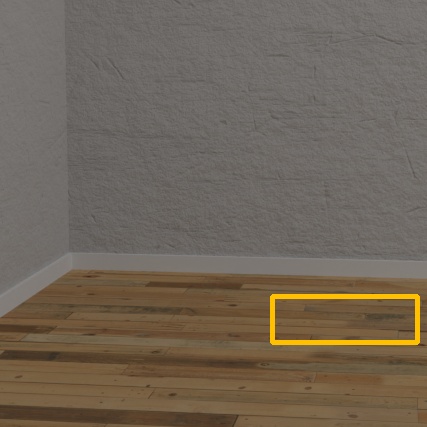}
        
    \end{minipage}
    \begin{minipage}[t]{0.19\textwidth}
        \centering
        \includegraphics[width=0.97\textwidth]{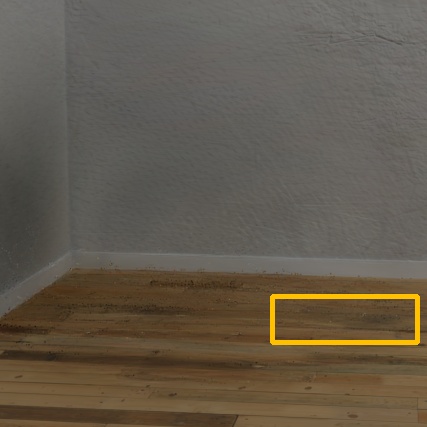}
        
    \end{minipage}
    \begin{minipage}[t]{0.19\textwidth}
        \centering
        \includegraphics[width=0.97\textwidth]{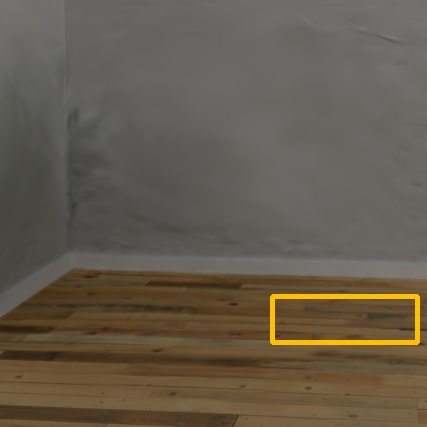}
        
    \end{minipage}
    \begin{minipage}[t]{0.19\textwidth}
        \centering
        \includegraphics[width=0.97\textwidth]{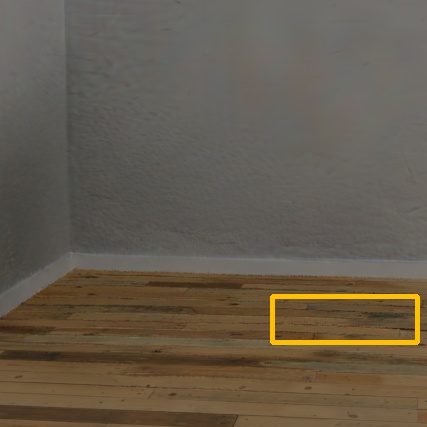}
        
    \end{minipage}
    \begin{minipage}[t]{0.19\textwidth}
        \centering
        \includegraphics[width=0.97\textwidth]{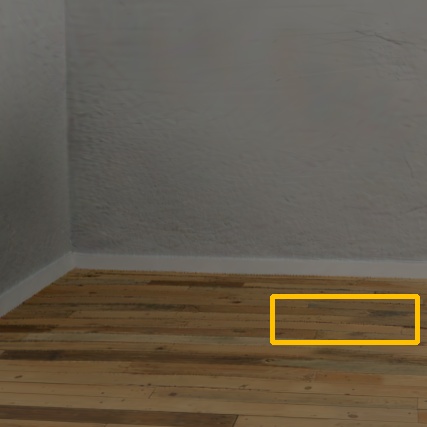}
        
    \end{minipage}
    \par\vspace{0.3mm}
    \begin{minipage}[t]{0.19\textwidth}
        \centering
        \includegraphics[width=0.97\textwidth]{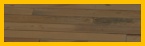}
        \caption*{Ground truth} 
    \end{minipage}
    \begin{minipage}[t]{0.19\textwidth}
        \centering
        \includegraphics[width=0.97\textwidth]{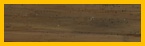}
        \caption*{w/o $L_{iN}$} 
    \end{minipage}
    \begin{minipage}[t]{0.19\textwidth}
        \centering
        \includegraphics[width=0.97\textwidth]{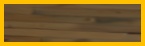}
        \caption*{w/o $L_{iLPIPS}$} 
    \end{minipage}
    \begin{minipage}[t]{0.19\textwidth}
        \centering
        \includegraphics[width=0.97\textwidth]{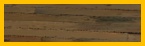}
        \caption*{w/o $L_{iC}$} 
    \end{minipage}
    \begin{minipage}[t]{0.19\textwidth}
        \centering
        \includegraphics[width=0.97\textwidth]{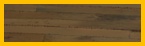}
        \caption*{Ours} 
    \end{minipage}

    \caption{Qualitative examples from the ablation study on loss term combinations. The figure shows sample results on our custom dataset. Excluding normal, LPIPS, or color loss in the inpainted region hinders geometric consistency or reduces appearance fidelity and structural detail.}
    \label{fig:exp_abl2}
\end{figure*}

Although the final proposed method seems to show slightly lower scores in LPIPS and SSIM compared to the \textit{w/o $\mathcal{L}_{iC}$} setting, Fig.~\ref{fig:exp_abl2} demonstrates that using all loss terms produces the most qualitatively satisfying results.
Without normal loss, the geometry becomes less accurate, resulting in distortions that fall within the acceptable threshold of the geometry consistency measure. Without a perceptual loss to guide the integration from inconsistent multi-view inpainting inputs, appearance tends to degrade where depth and normal information alone is insufficient to reconstruct accurate geometry.
Furthermore, relying on the appearance reconstruction solely on LPIPS loss without the support of L1 and SSIM losses leads to a noticeable degradation in fine details.

\subsection{Limitations}
\begin{figure}[htp]
    \centering
    \subfloat[Hallucination]{
        \label{fig:ours_limits_a}
        \includegraphics[width=0.60\columnwidth]{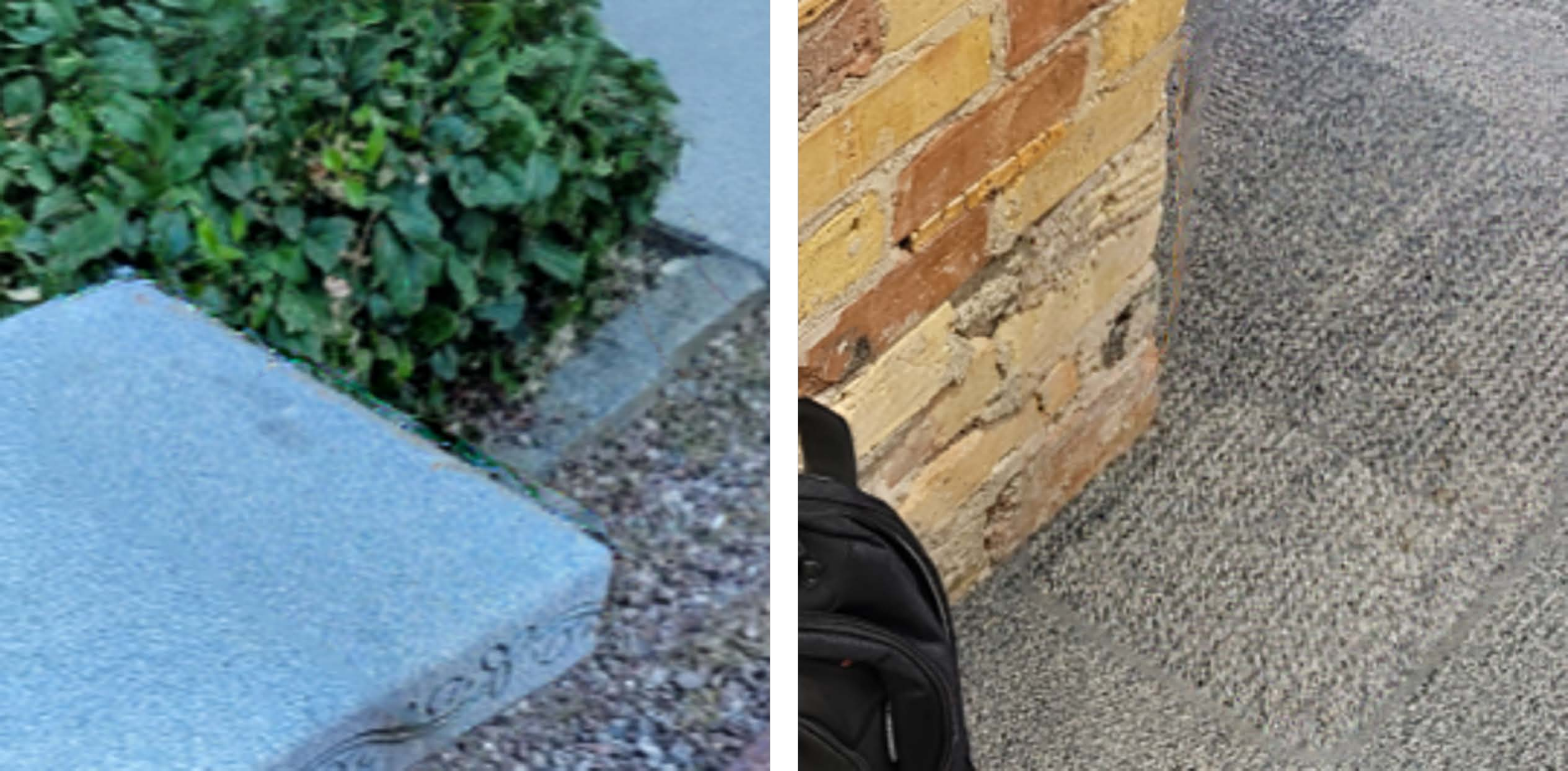}
    }
    \hfill
    \subfloat[Distortion]{
        \label{fig:ours_limits_b}
        \includegraphics[width=0.30\columnwidth]{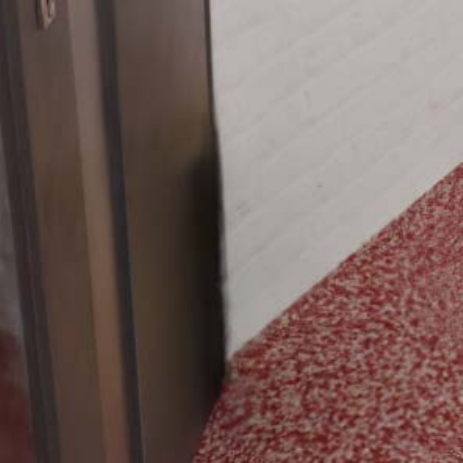}
    }
    \caption{Example of degraded performance. Due to the reliance on inpainted depth, hallucinations may appear at boundaries with abrupt depth changes in cases of severe inconsistency. For view directions with insufficient high-confidence inpainting, the reconstructed shape can be less accurate.}
    \label{fig:ours_limits}
\end{figure}

While our method enables more efficient and stable 3D scene inpainting than approaches relying on reprojection from limited views, some dependency on depth estimation remains.
It may suffer performance degradation in scenes with poor geometric consistency. As shown in Fig.~\ref{fig:ours_limits}, slight hallucinations appear around edge regions with abrupt depth changes, and distortions can be observed from view directions where high-confidence inpainting inputs were insufficient. While our method relaxes the implicit assumption in previous approaches that inpainting across views generally shares perceptual similarity, it still relies on geometric consistency. Therefore, it is vulnerable in situations where obtaining geometrically consistent inpaintings becomes difficult due to highly irregular or complex scenes, such as forests. To achieve robust 3D inpainting in more complex scenes, new methods that do not depend on inpainting consistency are required.

Additionally, although we apply Poisson blending for warping to achieve seamless integration and preserve view-dependent effects, its effectiveness diminishes when the inpainting region spans multiple objects or when residual shadows of the removed object remain.

\section{Conclusion}
In this paper, we propose a 3D scene inpainting method on 3DGS that reconstructs a coherent 3D scene. Based on a single user-specified reference view, our method estimates the reliability of inpainted results from other views by measuring their similarity to the reference. By leveraging all high-confidence information, it constructs a stable geometry and achieves a perceptually consistent appearance across multiple viewpoints through warping.
Experimental results demonstrate the effectiveness of our proposed method over existing 3D scene inpainting methods.

For future work, we aim to reduce the reliance on the geometrical consistency of inpainted images by leveraging personalized diffusion models~\cite{Ruiz2022Dreambooth, Kumari2023CustomDiffusion} to generate viewpoint-consistent inpaintings that encourage visual similarity to the reference image. 
Furthermore, we plan to investigate approaches for estimating uncertainty within the scene to enhance the identification of regions that require inpainting in 3D Gaussian scenes following object removal.

\section*{Acknowledgments}
This work was supported by the Industrial Technology Innovation Program (20012462) funded by the Ministry of Trade, Industry \& Energy (MOTIE, Korea), the National Research Foundation of Korea (NRF) grant (NRF-2021R1A2C2093065) funded by the Korea government (MSIT) and the KIST under the Institutional Program (Grant No. 2E33841).


\end{document}